\documentclass[10pt,twocolumn]{article}

\usepackage[T1]{fontenc}
\usepackage[utf8]{inputenc}
\usepackage[margin=0.75in]{geometry}

\usepackage{amsmath}
\usepackage{amssymb}
\usepackage{amsthm}
\usepackage{mathtools}

\usepackage{graphicx}
\graphicspath{{figures/}}

\usepackage{float}
\usepackage{subcaption}
\usepackage{booktabs}
\usepackage{multirow}
\usepackage{array}
\usepackage{tabularx}
\usepackage{enumitem}
\usepackage{xcolor}
\usepackage{forest}
\usepackage[most]{tcolorbox}
\usepackage{listings}
\usepackage{xurl}

\lstdefinelanguage{Python}{
    keywords={
        and,as,assert,async,await,break,class,continue,def,del,elif,
        else,except,False,finally,for,from,global,if,import,in,is,
        lambda,None,nonlocal,not,or,pass,raise,return,True,try,
        while,with,yield
    },
    keywordstyle=\color{blue}\bfseries,
    sensitive=true,
    comment=[l]\#,
    commentstyle=\color{green!50!black}\itshape,
    stringstyle=\color{red!70!black},
    morestring=[b]',
    morestring=[b]",
    morestring=[s]{'''}{'''},
    morestring=[s]{"""}{"""},
    emph={
        print,len,asyncio,PolicyEngine,FilePolicyStore,
        AuthorizationRequest,Principal,PrincipalType,Action,
        Resource,RequestContext
    },
    emphstyle=\color{violet},
}

\lstdefinestyle{pythonstyle}{
    language=Python,
    basicstyle=\ttfamily\scriptsize,
    keywordstyle=\color{blue}\bfseries,
    commentstyle=\color{green!50!black}\itshape,
    stringstyle=\color{red!70!black},
    emphstyle=\color{violet},
    numbers=left,
    numberstyle=\tiny\color{gray},
    stepnumber=1,
    numbersep=8pt,
    backgroundcolor=\color{gray!5},
    frame=single,
    rulecolor=\color{black!30},
    breaklines=true,
    breakatwhitespace=false,
    showstringspaces=false,
    tabsize=4,
    keepspaces=true,
    columns=fullflexible,
    captionpos=b
}

\lstdefinelanguage{json}{
    basicstyle=\ttfamily\scriptsize,
    numbers=none,
    breaklines=true,
    showstringspaces=false,
    frame=single,
    morestring=[b]",
    stringstyle=\color{blue},
    literate=
     *{0}{{{\color{magenta}0}}}{1}
      {1}{{{\color{magenta}1}}}{1}
      {2}{{{\color{magenta}2}}}{1}
      {3}{{{\color{magenta}3}}}{1}
      {4}{{{\color{magenta}4}}}{1}
      {5}{{{\color{magenta}5}}}{1}
      {6}{{{\color{magenta}6}}}{1}
      {7}{{{\color{magenta}7}}}{1}
      {8}{{{\color{magenta}8}}}{1}
      {9}{{{\color{magenta}9}}}{1}
      {:}{{{:}}}{1}
      {,}{{{,}}}{1}
      {\}}{{\}}}{1}
      {[}{{[}}{1}
      {]}{{]}}{1},
}

\usepackage[hidelinks]{hyperref}
\usepackage[nameinlink]{cleveref}

\usepackage{fancyhdr}
\begin{document}

\twocolumn[
  \begin{minipage}{\textwidth}
    \title{
  \Large\bfseries A Unified Policy Architecture (UPA):\\
   The Governance Kernel for Enterprise AI Operating Systems
}

\author{
  Prabhu Raghav$^{1}$ \quad
  Balamurugan Pandi$^{1}$ \quad
  Arul Vivek$^{1}$ \\[0.3em]
  Shek Mohammed$^{1}$ \quad
  Sridhar S$^{1}$ \\[0.8em]
  $^{1}$\textbf{SuperAgentX AI} \\[0.4em]
  \small\ttfamily \{prabhu, balamurugan, arul.vivek, shek.mohammed, sridhar.s\}@superagentx.ai
}

\date{}

\maketitle
    \vspace{0.5em}
    \begin{abstract}

Enterprise Artificial Intelligence is evolving from conversational
assistants toward autonomous software agents capable of planning,
reasoning, invoking external tools, accessing enterprise knowledge,
collaborating with other agents, and executing complex business
workflows with limited human intervention. As autonomy increases,
governance requirements extend beyond conventional model safety,
prompt filtering, and content moderation to include authorization,
compliance, human oversight, data protection, auditability, and
runtime control of agent actions.

Current enterprise AI governance mechanisms are fragmented across
multiple domains, including identity and access management, AI
guardrails, workflow orchestration, compliance frameworks, security
controls, audit systems, and business-specific authorization rules.
These mechanisms are frequently implemented as independent controls,
resulting in duplicated governance logic, inconsistent enforcement,
limited interoperability, and increasing operational complexity as
organizations deploy autonomous AI agents across heterogeneous
enterprise environments.

This paper proposes a \textit{Unified Policy Architecture (UPA)}, a
governance architecture that treats policy enforcement as a
first-class architectural capability of Enterprise AI Operating
Systems. UPA introduces a deterministic Policy Kernel that separates
policy specification, policy matching, policy evaluation, and runtime
enforcement from autonomous agent reasoning and application logic.
The architecture uses a standardized Principal--Action--Resource--Context
(PARC) request model and enables policies to produce both governance
decisions and executable governance obligations.

Unlike conventional AI guardrails that primarily focus on model inputs
and outputs, UPA provides a unified governance model spanning the
lifecycle of autonomous execution, including authentication,
authorization, planning, memory access, tool invocation, inter-agent
collaboration, workflow execution, human approval, compliance
validation, audit generation, and runtime governance. The architecture
further introduces declarative policy composition, standardized policy
namespaces, extensible governance plugins, and industry-specific
policy packs for heterogeneous enterprise environments.

The paper presents the architectural foundations of UPA and its
associated Declarative Governance Policy Language (DGPL), together
with a benchmark-oriented evaluation methodology for enterprise AI
governance scenarios. The proposed architecture provides a reference
model for integrating authorization, safety, compliance, and runtime
governance mechanisms under a common policy-centric control plane,
establishing a foundation for interoperable and governable Enterprise
AI Operating Systems.
    
\end{abstract}
    \vspace{0.5em}

\vspace{1cm}
    \vspace{1.5em}
  \end{minipage}
]

\thispagestyle{plain}

\section{Introduction}
Enterprise software has historically evolved through successive architectural abstractions. Operating systems abstracted hardware resources, middleware abstracted distributed communication, cloud platforms abstracted infrastructure management, and identity systems centralized authentication and authorization. Each architectural evolution emerged in response to increasing system complexity and the need for standardized control mechanisms.
\\
Artificial Intelligence is now driving a comparable architectural shift. Large Language Models (LLMs) have rapidly evolved into autonomous AI agents capable of reasoning over enterprise knowledge, generating execution plans, invoking external tools, maintaining persistent memory, collaborating with specialized agents, and executing complex workflows with minimal human supervision. Rather than acting solely as conversational interfaces, these systems increasingly perform operational tasks that directly influence enterprise processes and business outcomes.
\\
This transition fundamentally changes the governance requirements of enterprise AI systems. Traditional governance mechanisms were designed for deterministic software whose execution paths were explicitly defined by developers. Autonomous AI agents, however, dynamically determine execution strategies, select tools at runtime, adapt to changing contexts, and coordinate with other intelligent agents. As agent autonomy increases, governance can no longer be limited to controlling software access or filtering generated responses. Instead, governance must encompass the complete lifecycle of autonomous decision making.
\\
Current approaches to AI governance largely focus on model-level safety through techniques such as prompt guardrails, jailbreak detection, toxicity classification, output moderation, and responsible AI evaluation. These capabilities are essential to improve interaction safety and reduce harmful behavior of the model. However, enterprise deployments require governance over substantially broader concerns, including authorization, memory access, tool invocation, workflow orchestration, business rules, regulatory compliance, cost management, human approvals, auditability, risk management, and policy enforcement across multiple collaborating agents and humans.
\\
Consequently, \textit{\textbf{enterprise AI governance should be viewed as a system architecture problem rather than as a model safety problem only}}. Organizations require a unified governance layer capable of consistently enforcing enterprise policies across heterogeneous AI runtimes, business domains, and operational environments while remaining independent of any specific model provider, agent framework, or execution platform.
\\
This paper proposes the Unified Policy Architecture (UPA), a policy-centric governance architecture that unifies enterprise governance, AI runtime governance, and domain-specific governance within a common architectural framework. UPA separates governance logic from application logic through declarative policies, enabling consistent enforcement across the complete lifecycle of autonomous AI execution while supporting extensibility, interoperability, and organizational scalability.
\\\\
The primary contributions of this paper are:

\begin{enumerate}

\item We identify Enterprise AI Governance as a systems architecture challenge that extends beyond traditional authorization, AI safety, and guardrail mechanisms, requiring a unified runtime governance layer for Enterprise AI Operating Systems.

\item We propose the Unified Policy Architecture (UPA), introducing the \emph{UPA Policy Kernel} as the declarative runtime governance layer that unifies authorization, runtime governance, compliance, auditability, approval workflows, and multi-agent governance, business rules, within a single policy-centric architecture.

\item We introduce a framework-independent Semantic Normalization model that transforms heterogeneous runtime events into a canonical policy representation, enabling consistent governance across LLMs, ML models, AI agents, workflows, tools, APIs, and enterprise applications.

\item We introduce a Governance Provider Framework that separates Evaluation Providers (e.g., threat detection, risk scoring, contextual enrichment) from Obligation Providers (e.g., privacy protection, compliance verification, audit logging), enabling extensible governance without modifying the core Policy Kernel.

\item We define a declarative governance policy model supporting authorization, AI safety, threat detection, approval workflows, governance obligations, and industry-specific policy packs, providing a unified governance language for heterogeneous Enterprise AI Operating Systems.

\end{enumerate}

As enterprises increasingly rely on autonomous AI systems to perform operational tasks, governance must evolve from isolated runtime controls toward comprehensive architectural governance. We argue that governance should become a foundational layer of enterprise AI systems, providing organizations with the mechanisms necessary to balance autonomy, security, compliance, accountability, and operational trust.

\section{Related Work}
The rapid adoption of Large Language Models (LLMs) and autonomous AI agents has accelerated research across AI safety, governance, security, policy enforcement, and multi-agent systems. Existing work has introduced valuable mechanisms for securing AI interactions, enforcing organizational policies, managing enterprise identities, and orchestrating autonomous workflows. However, these efforts largely address individual aspects of AI governance rather than providing a unified governance architecture spanning the complete lifecycle of autonomous AI agents.

This section reviews the major research areas related to enterprise AI governance and identifies the architectural gap addressed by the proposed Unified Policy Architecture (UPA).

\subsection{Policy-Based Access Control}
Policy-Based Access Control (PBAC) has long been adopted within enterprise systems to externalize authorization decisions from application logic. Standards such as {eXtensible Access Control Markup Language (XACML) (OASIS, 2005)} \cite{xacml2013} introduced declarative policy languages capable of expressing complex authorization rules, while Open Policy Agent (OPA)  \cite{opa} \cite{opa2020} demonstrated policy-as-code for cloud-native infrastructure using the Rego language.

Similarly, Role-Based Access Control (RBAC) \cite{sandhu1996rbac}, Attribute-Based Access Control (ABAC) \cite{hu2015abac}, and Relationship-Based Access Control (ReBAC) \cite{fong2011rebac} provide mature mechanisms for authorization within enterprise applications.

Although these approaches effectively govern identities, resources, and permissions, they assume deterministic software execution and were not designed to govern autonomous AI agents capable of dynamic planning, reasoning, memory access, tool selection, and inter-agent collaboration.

\subsection{Cedar Policy Language and Fine-Grained Authorization}

Cedar, introduced by Cutler et al.~\cite{cedar2024}, provides a
declarative, analyzable, and high-performance policy language for
fine-grained authorization. Its formal policy semantics, validation
mechanisms, and separation of authorization logic from application
code provide an important foundation for modern policy-based access
control.

UPA is substantially influenced by the design principles demonstrated
by Cedar, particularly the use of declarative policies, explicit
principal--action--resource relationships, contextual evaluation, and
a policy evaluation model that separates governance decisions from
application logic. UPA does not attempt to replace or generalize
Cedar's authorization semantics. Instead, it builds upon the broader
policy-centric abstraction to address governance requirements that
extend beyond authorization.

In particular, UPA introduces governance constructs for autonomous AI
execution, including semantic normalization, runtime policy
obligations, plugin-based enforcement, compliance verification, human
approval workflows, governance evidence, industry-specific policy
packs, and multi-agent governance. These mechanisms are intended to
govern not only whether an action is authorized, but also how an
authorized
AI operation may be executed within an enterprise governance
environment.

Accordingly, Cedar represents an important authorization foundation,
while UPA investigates a broader architectural layer for lifecycle-wide
governance of Enterprise AI Operating Systems.

\begin{figure*}[!t]
\centering
\includegraphics[width=0.75\textwidth]{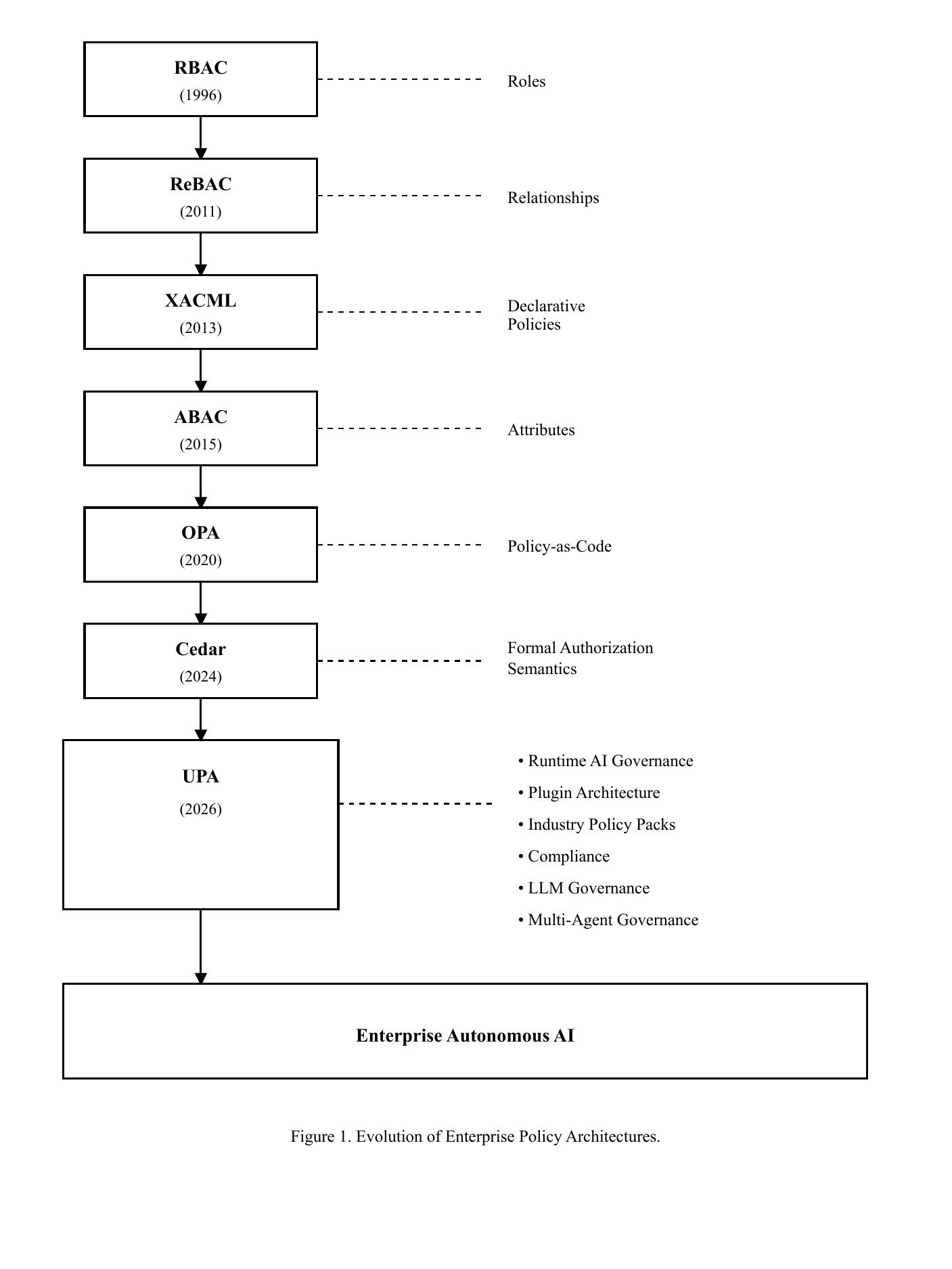}
\caption{
Evolution of enterprise policy architectures from RBAC to UPA, illustrating the progression from authorization-centric models toward policy-centric runtime governance for Enterprise AI Operating Systems.
}
\label{fig:evolution}
\end{figure*}

The Unified Policy Architecture (UPA) draws inspiration from Cedar's declarative policy model and operational semantics. However, UPA extends authorization beyond permission evaluation by introducing plugin-based runtime transformations, governance evidence generation, compliance verification, risk assessment, policy obligations, approval workflows, audit generation, and industry-specific policy packs as first-class architectural concepts.

\subsection{AI Safety and Guardrails}
Recent advances in Generative AI have led to extensive research on model safety, prompt injection defense, jailbreak detection, toxicity filtering, hallucination mitigation, and output moderation. Significant contributions include Constitutional AI Bai et al. (2022) \cite{bai2022constitutional} from Anthropic, OpenAI's alignment and safety research, Google DeepMind's Gemini safety framework, NVIDIA NeMo Guardrails Kotha et al. (2023) \cite{rebedea2023nemoguardrails}.

The AI security community has also developed standardized threat models for LLM applications. The OWASP Top 10 for Large Language Model Applications identifies the most critical security risks affecting production LLM systems, including prompt injection, insecure output handling, training data poisoning, sensitive information disclosure, insecure plugin design, excessive agency, overreliance, and model theft \cite{owasp2025llm}. These risks provide an industry-recognized baseline for securing LLM-enabled applications and emphasize that security challenges extend beyond traditional software vulnerabilities to encompass model-mediated and agentic behaviors.

Commercial guardrail frameworks provide mechanisms for validating prompts, detecting harmful content, applying safety filters, and preventing unsafe model outputs. However, guardrails primarily focus on validating model inputs and generated responses rather than governing enterprise concerns such as authorization, workflow execution, memory access, business approvals, regulatory compliance, auditability, policy inheritance, or multi-agent coordination.

\subsection{Enterprise Identity and Zero Trust}

Enterprise security architectures have progressively evolved from perimeter-based security models toward identity-centric security paradigms. Zero Trust Architecture (ZTA), formalized by Rose et al. (2020) \cite{rose2020zerotrust} in NIST Special Publication 800-207, establishes the principle that no user, device, application, or workload should be implicitly trusted based solely on its network location or ownership. Instead, authentication and authorization decisions are continuously evaluated using contextual information such as user identity, device posture, workload characteristics, resource sensitivity, and environmental conditions before granting access to enterprise resources \cite{rose2020zerotrust}. Zero Trust therefore shifts enterprise security from protecting network boundaries toward protecting enterprise resources through continuous verification and least-privilege access.

Modern enterprise security platforms implement Zero Trust through identity providers, multi-factor authentication, privileged access management, policy decision points, policy enforcement points, continuous risk assessment, and adaptive authorization. These mechanisms significantly improve enterprise security by reducing lateral movement, minimizing implicit trust, and enforcing identity-aware access control across distributed cloud environments.

However, autonomous AI systems introduce execution contexts that extend beyond traditional human identities. Enterprise AI agents autonomously retrieve enterprise knowledge, invoke external tools, execute workflows, coordinate with other agents, maintain persistent memory, and dynamically generate execution plans without continuous human supervision. Consequently, identity verification alone cannot govern autonomous AI execution because governance decisions increasingly depend not only on who performs an action, but also on which agent is executing, which tools are invoked, which enterprise policies apply, which compliance obligations must be satisfied, and whether runtime behavior remains consistent with organizational governance policies.

The Unified Policy Architecture (UPA) complements Zero Trust by extending identity-centric security into policy-centric runtime governance. While Zero Trust determines whether an authenticated principal should access an enterprise resource, UPA continuously governs the complete lifecycle of autonomous AI execution through declarative policy evaluation, authorization, plugin-based runtime transformations, compliance verification, approval workflows, governance evidence generation, audit logging, and multi-agent policy enforcement. In this sense, Zero Trust establishes the identity foundation, whereas UPA provides the runtime governance layer required for enterprise autonomous AI systems.

\subsection{Governance-First AI Architectures}

Recent work has begun to recognize governance as a first-class
architectural concern for agentic AI systems. Wen et al. (2026)
proposed Arbiter-K, a governance-first execution architecture that
introduces a deterministic neuro-symbolic kernel positioned between
probabilistic language models and environment-impacting operations.
Arbiter-K employs a Semantic Instruction Set Architecture (ISA), an
instruction dependency graph, and a taint-aware execution model to
enforce security policies during agent execution, demonstrating that
governance can be integrated into the execution substrate rather than
implemented solely through application-level guardrails
\cite{wen2026arbiterk}.

Arbiter-K primarily focuses on execution security and deterministic
control of agent behaviors. Enterprise deployments, however, require a
broader governance architecture encompassing authorization,
compliance, policy management, approval workflows, auditability,
industry-specific policies, and heterogeneous enterprise
integrations. UPA addresses these broader requirements through a
declarative Policy Kernel designed to serve as a governance control
layer for Enterprise AI Operating Systems.
\subsection{Multi-Agent Systems}
Research in multi-agent systems (MAS) has established the theoretical foundations for autonomous reasoning, distributed decision making, cooperation, coordination, negotiation, and communication among intelligent agents. Foundational work by Wooldridge (2009) and Russell and Norvig (2021) formalized many of the principles underlying autonomous agents and distributed artificial intelligence. More recently, the emergence of Large Language Models has accelerated the development of LLM-based autonomous agent frameworks, including ReAct \cite{yao2023react}, Toolformer \cite{schick2023toolformer}, AutoGen \cite{wu2023autogen}, CAMEL \cite{li2023camel}, and Voyager \cite{wang2023voyager}. These systems demonstrate significant advances in autonomous planning, reasoning, tool invocation, memory management, task decomposition, reflection, and collaborative problem solving.

These frameworks primarily focus on improving agent capabilities, including reasoning strategies, planning algorithms, memory architectures, tool integration, and multi-agent collaboration. While several frameworks provide callback mechanisms, execution hooks, orchestration logic, or framework-specific control mechanisms, governance policies are generally embedded within application code or tightly coupled to the execution framework. Consequently, policy enforcement is often implementation-specific and difficult to standardize across heterogeneous agent runtimes, enterprise applications, and organizational domains.

As enterprise deployments increasingly integrate multiple autonomous agents operating across diverse business systems, organizations require governance mechanisms that are independent of any individual agent framework or model provider. Such governance must consistently regulate authorization, tool usage, memory access, workflow execution, compliance obligations, approval workflows, audit generation, and inter-agent interactions throughout the execution lifecycle. The Unified Policy Architecture (UPA) addresses this requirement by externalizing governance into a declarative policy layer that operates independently of the underlying LLM, agent framework, orchestration engine, or execution environment.

\textit{"Unlike existing multi-agent frameworks that primarily focus on agent intelligence and orchestration, UPA focuses on governance as a first-class architectural layer, enabling consistent policy enforcement across heterogeneous autonomous AI systems regardless of the underlying agent implementation."}












\subsection{Positioning of DGPL and UPA}

The reviewed approaches establish strong foundations for access
control, policy evaluation, zero-trust security, AI safety, and
autonomous-agent execution. However, these capabilities are commonly
addressed as separate architectural concerns.

DGPL and UPA combine these concerns into a unified policy-centric
governance model for Enterprise AI systems. The objective is not to
replace existing authorization engines, guardrails, or agent
frameworks, but to provide a common declarative policy layer through
which they can be governed consistently.

The architectural distinction can be summarized as follows:

\begin{table*}[!t]
\centering
\caption{Positioning of DGPL and UPA Relative to Existing Approaches}
\label{tab:relatedwork}
\small
\begin{tabular}{|p{2.7cm}|p{4.2cm}|p{4.0cm}|p{4.0cm}|}
\hline
\textbf{Approach}
&
\textbf{Primary Focus}
&
\textbf{Primary Governance Scope}
&
\textbf{Relationship to UPA}
\\
\hline

RBAC
&
Role-based authorization
&
Permissions associated with roles
&
Authorization foundation
\\
\hline

ABAC
&
Attribute-based authorization
&
Principal, resource, action, and environmental attributes
&
Context-aware policy foundation
\\
\hline

ReBAC
&
Relationship-based authorization
&
Relationships among principals and resources
&
Relationship-aware authorization mechanism
\\
\hline

XACML
&
Declarative access-control policy
&
Policy evaluation, combining algorithms,
obligations, and advice
&
Important foundation for declarative governance
\\
\hline

OPA / Rego
&
General-purpose Policy-as-Code
&
Structured policy evaluation across systems
&
External policy decision mechanism that can
serve governance use cases
\\
\hline

Cedar
&
Formal authorization
&
Principal-action-resource-context authorization
&
Authorization foundation and PARC model
\\
\hline

Zero Trust
&
Continuous security
&
Identity, resource, device, and contextual access
&
Security foundation complemented by runtime governance
\\
\hline

AI Guardrails
&
AI interaction safety
&
Input/output validation, filtering, and safety controls
&
Governance providers/plugins within UPA
\\
\hline

Agent Frameworks
&
Agent reasoning and orchestration
&
Planning, memory, tools, and multi-agent execution
&
Execution clients governed by external policies
\\
\hline

Arbiter-K
&
Governance-first agent execution
&
Deterministic execution security and
instruction-level enforcement
&
Complementary execution substrate
\\
\hline

\textbf{DGPL / UPA}
&
\textbf{Declarative enterprise AI governance}
&
\textbf{Authorization, approval, transformation,
compliance, audit, and runtime governance}
&
\textbf{Unified policy-centric governance layer}
\\
\hline

\end{tabular}
\end{table*}

The key distinction is that UPA treats authorization as one component
of a larger governance lifecycle. A policy evaluation may therefore
produce not only an authorization decision but also a set of
governance obligations:

\[
\operatorname{Eval}(q,\Pi)
=
(d,\Gamma)
\]

where $d$ represents the governance decision and $\Gamma$ represents
the obligations generated by the applicable policies.

These obligations may include plugin execution, human approval,
audit generation, notification, compliance validation, or other
policy-defined governance actions.

Consequently, UPA separates three concerns that are frequently
coupled in existing systems:

\[
\begin{aligned}
  &\text{Policy Semantics} \\
  &\quad\downarrow \\
  &\text{Governance Decision} \\
  &\quad\downarrow \\
  &\text{Runtime Enforcement}
\end{aligned}
\]

This separation enables existing authorization technologies,
AI-safety mechanisms, and agent execution frameworks to participate
in a common enterprise governance architecture without requiring
their internal implementations to become part of the Policy Kernel.

\section{Problem Statement \& Research Gap}

Enterprise autonomous AI systems fundamentally differ from traditional enterprise software. Unlike deterministic applications, autonomous AI agents dynamically reason, plan, invoke external tools, retrieve enterprise knowledge, maintain persistent memory, collaborate with other agents, and adapt their execution strategies at runtime. Consequently, governance decisions must be continuously enforced throughout the execution lifecycle rather than evaluated at a single authorization or safety checkpoint.

Existing research addresses individual aspects of enterprise AI governance but lacks a unified architectural approach. Authorization frameworks determine access permissions, guardrails improve interaction safety, Zero Trust continuously verifies identities, and multi-agent frameworks enhance autonomous reasoning and collaboration. However, these capabilities operate independently and do not provide a common governance abstraction capable of consistently enforcing enterprise policies across heterogeneous AI systems.

This fragmentation creates several challenges for enterprise deployments:

\begin{itemize}
    \item \textbf{Fragmented Governance}: Authorization, AI safety, compliance, approvals, and audit mechanisms are implemented independently, resulting in duplicated logic and inconsistent policy enforcement.
    \item \textbf{Framework Dependency}: Governance mechanisms are tightly coupled to individual LLM providers, agent frameworks, or orchestration platforms, limiting interoperability and portability.
    \item \textbf{Limited Runtime Governance}: Existing approaches primarily evaluate access permissions or model interactions but do not continuously govern memory access, tool invocation, workflow execution, inter-agent communication, and dynamic execution plans.
    \item \textbf{Compliance Integration}: Regulatory requirements and industry-specific governance policies are typically embedded within application code rather than expressed as reusable declarative policies.
    \item \textbf{Lack of Standardization}: Organizations lack a common policy architecture capable of governing heterogeneous autonomous AI systems using a consistent governance model.
\end{itemize}

As enterprises increasingly deploy autonomous AI agents that interact with critical business systems, sensitive enterprise knowledge, regulated data, and operational workflows, these limitations become significantly more pronounced. The absence of a unified governance layer increases implementation complexity, reduces policy consistency, complicates auditing, and limits the portability of governance policies across heterogeneous enterprise environments. More fundamentally, the emergence of autonomous AI systems suggests that enterprises are transitioning toward Enterprise AI Operating Systems, where multiple agents, models, tools, memories, workflows, and enterprise applications execute within a common runtime environment. Similar to traditional operating systems, which rely on a kernel to manage processes, memory, resources, security, and system calls, an Enterprise AI Operating System requires a policy kernel that continuously governs autonomous execution. Rather than treating governance as an application-specific concern, the policy kernel becomes the central runtime component responsible for authorization, policy enforcement, plugin execution, compliance verification, approval workflows, audit generation, and coordination across heterogeneous AI agents and enterprise services. 


We argue that this architectural abstraction is essential for building scalable, trustworthy, and interoperable enterprise autonomous AI systems.
\section{Unified Policy Architecture (UPA)}
\subsection{Relationship to Existing Governance Architectures}
Recent research has increasingly recognized that autonomous AI systems require governance mechanisms beyond traditional authorization and model safety. Arbiter-K introduces a governance-first execution architecture that places a deterministic symbolic kernel between probabilistic reasoning and environment-impacting operations. Its primary contribution is a Semantic Instruction Set Architecture (ISA), instruction dependency graph, and taint-aware execution model that enables deterministic enforcement of security policies during agent execution.

Similarly, A Five-Plane Reference Architecture for Runtime Governance of Production AI Agents argues that production AI agents require stateful runtime governance beyond traditional request-time authorization. The proposed architecture introduces composable governance planes, composite principals, capability attenuation, interruption primitives, and structured audit evidence for governing delegated agent actions.

The Unified Policy Architecture (UPA) shares the objective of treating governance as a first-class architectural concern. However, UPA addresses a complementary problem by positioning governance as the Policy Kernel of an Enterprise AI Operating System. Rather than focusing primarily on execution security or runtime mediation, UPA introduces a declarative policy architecture that unifies authorization, semantic normalization, runtime governance, plugin orchestration, compliance verification, approval workflows, governance evidence generation, industry-specific policy packs, and multi-agent governance within a common policy framework. The architecture is intentionally platform-independent, enabling consistent governance across heterogeneous LLMs, agent frameworks, enterprise applications, workflow engines, and organizational domains.

\subsection{Design Principles}

The Unified Policy Architecture (UPA) is a reference architecture for policy-centric runtime governance within Enterprise AI Operating Systems. Unlike existing approaches that treat authorization, AI safety, compliance, workflow governance, and auditability as independent capabilities, UPA externalizes governance into a unified \textbf{Policy Kernel} responsible for continuously governing autonomous AI execution throughout its lifecycle.

The design of UPA is motivated by the observation that enterprise autonomous AI systems execute across heterogeneous environments comprising multiple Large Language Models (LLMs), agent frameworks, workflow engines, enterprise applications, external tools, and organizational policies. Embedding governance logic directly within these execution environments results in fragmented implementations, inconsistent policy enforcement, limited portability, and increased operational complexity. UPA therefore adopts a \textbf{policy-centric architecture} in which governance is separated from application logic and implemented as an independent runtime layer.

\subsection{Architectural Overview}
\begin{figure*}[!t]
\centering
\includegraphics[width=0.95\textwidth]{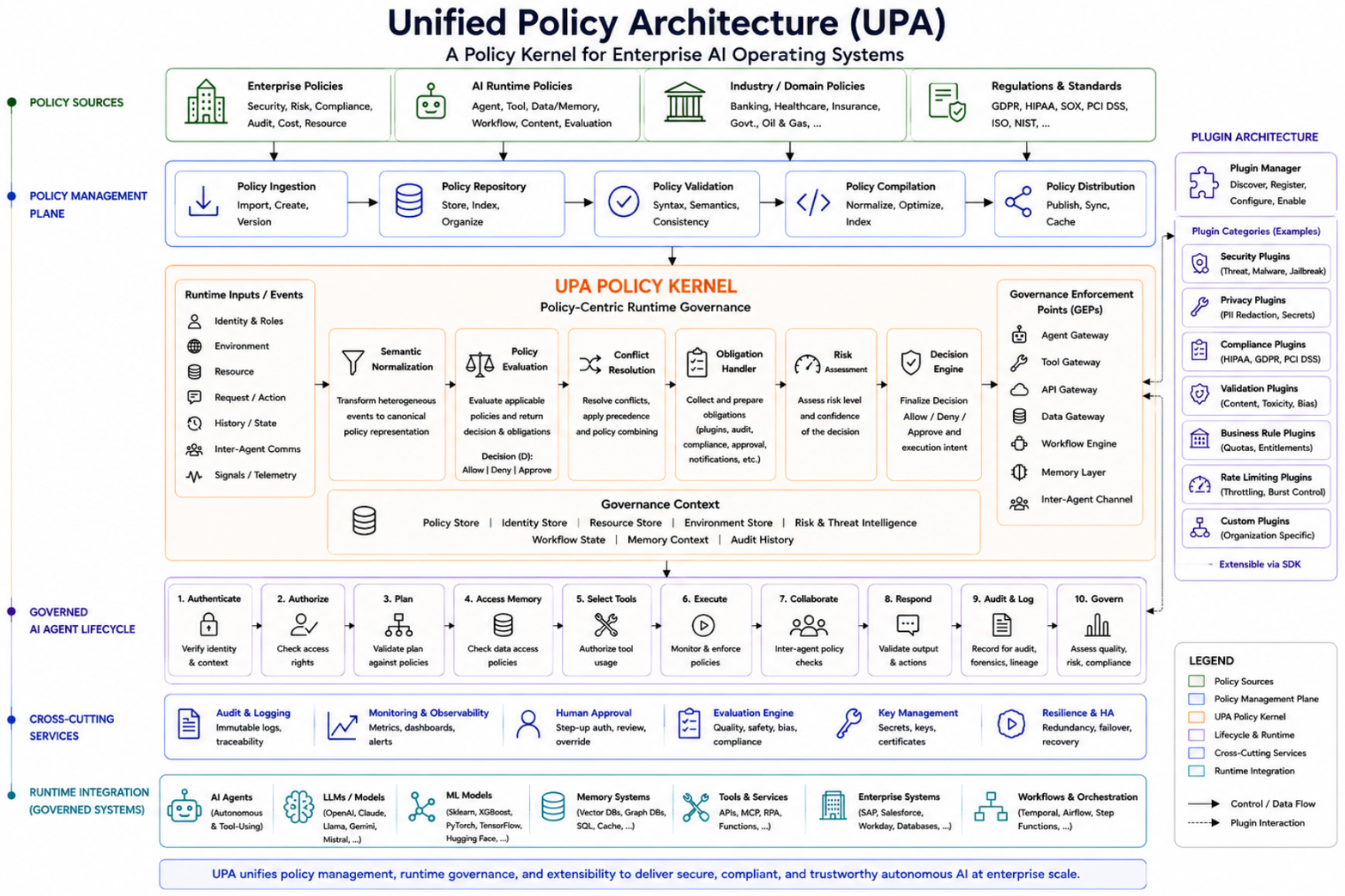}
\caption{
Overall architecture of the Unified Policy Architecture (UPA). The UPA Policy Kernel serves as the declarative runtime governance layer of an Enterprise AI Operating System, providing semantic normalization, policy evaluation, runtime governance, plugin orchestration, compliance verification, auditability, approval workflows, and industry-specific governance across heterogeneous AI runtimes.
}
\label{fig:upa_architecture}
\end{figure*}

Figure~\ref{fig:upa_architecture} presents the overall architecture of the Unified Policy Architecture (UPA). UPA is designed as a Policy Kernel within an Enterprise AI Operating System, providing a unified governance layer that continuously regulates autonomous AI execution independently of the underlying Large Language Models (LLMs), agent frameworks, workflow engines, enterprise applications, and cloud platforms. By externalizing governance from application logic, UPA enables organizations to consistently enforce enterprise policies across heterogeneous execution environments while maintaining portability, extensibility, and interoperability.

The architecture is organized into three logical layers: the Policy Kernel, Governance Services, and the AI Runtime Integration Layer. The Policy Kernel forms the core of the architecture and is responsible for transforming heterogeneous runtime events into canonical policy representations, evaluating declarative policies, coordinating runtime governance decisions, and orchestrating extensible governance plugins. Surrounding the kernel, Governance Services provide enterprise capabilities including compliance verification, approval workflows, audit and governance evidence generation, and industry-specific policy packs. Finally, the AI Runtime Integration Layer connects the Policy Kernel with autonomous agents, foundation models, enterprise memory systems, external tools, workflows, and enterprise applications, enabling governance policies to be enforced consistently throughout the execution lifecycle.

Unlike traditional authorization architectures that evaluate permissions only before resource access, UPA performs continuous runtime governance. Every significant execution event—including natural language requests, agent planning decisions, tool invocations, workflow transitions, memory operations, and inter-agent communications—is intercepted by the Policy Kernel before execution proceeds. The kernel evaluates applicable governance policies, invokes policy plugins where required, verifies compliance obligations, determines approval requirements, and records governance evidence before permitting, denying, modifying, or deferring execution.

A key architectural principle of UPA is the separation of governance from execution. AI agents, orchestration engines, workflow platforms, and enterprise applications remain responsible for performing computational tasks, while governance decisions are delegated to the Policy Kernel through declarative policies. This separation enables organizations to evolve governance independently of application implementations, promotes policy reuse across heterogeneous systems, and provides a common governance abstraction spanning multiple AI frameworks and enterprise domains.

The modular design of UPA further enables organizations to extend governance through a plugin-based architecture. Specialized governance capabilities—including personally identifiable information (PII) redaction, secret detection, content filtering, regulatory compliance verification, threat detection, and industry-specific validation—can be integrated as reusable plugins without modifying the core Policy Kernel. This extensibility allows enterprises to tailor governance behavior to organizational policies, regulatory requirements, and domain-specific operational constraints while preserving a consistent policy evaluation model.

Collectively, these architectural layers establish UPA as the governance foundation of an Enterprise AI Operating System. Rather than functioning as a standalone authorization engine or application-level middleware, the Policy Kernel provides a unified execution abstraction that governs autonomous AI behavior across heterogeneous enterprise environments.

\subsection{UPA Policy Kernel}
The UPA Policy Kernel is the core governance component of the Unified Policy Architecture. It provides a unified runtime abstraction for governing autonomous AI execution within an Enterprise AI Operating System. Inspired by the role of kernels in traditional operating systems, the Policy Kernel continuously mediates interactions between AI agents, enterprise resources, external tools, memory systems, workflow engines, and organizational policies. Rather than embedding governance logic within applications or agent frameworks, the Policy Kernel externalizes governance into a declarative runtime layer that evaluates enterprise policies before, during, and after autonomous execution.

The Policy Kernel is organized into four primary subsystems: \textbf{Semantic Normalization}, \textbf{Policy Evaluation}, \textbf{Runtime Governance}, and the \textbf{Plugin Framework}. Together, these subsystems transform heterogeneous runtime events into canonical policy requests, evaluate applicable governance policies, coordinate enforcement decisions, and extend governance through modular plugins. Figure~\ref{fig:upa_architecture} illustrates the interaction between these subsystems within the overall UPA architecture.

\subsubsection{Semantic Normalization}

Let

\[
E
\]

denote the set of heterogeneous runtime events generated by an Enterprise AI Operating System, including natural language requests, agent actions, workflow events, API invocations, memory operations, tool executions, and inter-agent communications.

Let

\[
P
\]

denote the set of canonical policy requests.

The Semantic Normalization subsystem is defined as the transformation

\[
N : E \rightarrow P
\]

where each runtime event

\[
e \in E
\]

is transformed into a canonical policy request

\[
p = N(e).
\]

The canonical policy request is represented as

\[
p=
\langle
pr,\,
ac,\,
re,\,
ct,\,
md
\rangle
\]

where

\[
\begin{aligned}
pr &:& \text{Principal},\\
ac &:& \text{Action},\\
re &:& \text{Resource},\\
ct &:& \text{Execution Context},\\
md &:& \text{Runtime Metadata}.
\end{aligned}
\]

Therefore,

\[
N(e)=
\langle
pr,\,
ac,\,
re,\,
ct,\,
md
\rangle .
\]

The output of Semantic Normalization serves as the canonical input to the Policy Evaluation subsystem, enabling framework-independent governance across heterogeneous AI runtimes.

\subsubsection{Policy Evaluation}

Following semantic normalization, the canonical policy request is submitted to the \textit{Policy Evaluation} subsystem. The Policy Evaluation subsystem determines whether the request satisfies the applicable declarative governance policies defined by the organization. Unlike conventional authorization engines that evaluate only access permissions, UPA evaluates governance policies over the complete execution context of autonomous AI systems, including identities, resources, execution context, organizational policies, regulatory constraints, and runtime state.

Let

\[
P
\]

denote the set of canonical policy requests,

\[
\Pi
\]

the policy repository, and

\[
D =
\{
\texttt{Allow},
\texttt{Deny},
\texttt{Approve}
\}
\]

the set of governance decisions.

Policy evaluation is formally defined as

\[
Eval : P \times \Pi \rightarrow D \times \Gamma
\]

where

\[
Eval(p,\Pi)=(d,\Gamma)
\]

for a canonical policy request

\[
p \in P,
\]

where

\[
d \in D
\]

represents the governance decision and

\[
\Gamma =
\{
\gamma_1,\gamma_2,\ldots,\gamma_n
\}
\]

denotes the ordered set of governance obligations generated during policy evaluation.

Each governance obligation

\[
\gamma_i \in \Gamma
\]

represents a runtime action that must subsequently be enforced by the Runtime Governance subsystem. Governance obligations may include plugin execution, audit logging, compliance verification, governance evidence generation, notification, policy obligations, and approval metadata.

The resulting decision is defined as

\[
d =
\begin{cases}
\texttt{Allow}, & \parbox[t]{2.1in}{\raggedright if all applicable policy conditions are satisfied,}\\[12pt]

\texttt{Deny},  & \parbox[t]{2.1in}{\raggedright if one or more mandatory policy conditions are violated,}\\[12pt]

\texttt{Approve}, & \parbox[t]{2.1in}{\raggedright if execution requires explicit human authorization prior to continuation.}
\end{cases}
\]

Unlike traditional policy engines that primarily return binary authorization decisions, UPA introduces \textbf{\texttt{Approve}} as a first-class declarative policy outcome. The \texttt{Approve} decision temporarily suspends autonomous execution and transfers control to an organizational approval workflow before execution may continue. This capability enables \textbf{Human-in-the-Loop (HITL) governance} for high-risk autonomous operations, including financial transactions, production system modifications, regulated data access, infrastructure changes, and other business-critical activities.

Formally,

\[
\texttt{Approve}
\Longrightarrow
AwaitApproval(principal)
\]

where execution remains suspended until the required approval workflow has been successfully completed.

The governance obligations generated during policy evaluation are accumulated as

\[
\Gamma =
\bigcup_{s \in \Pi}
Obligations(s)
\]

where

\[
Obligations(s)=
\{
\gamma_1,\gamma_2,\ldots,\gamma_k
\}
\]

represents the obligations associated with the matched policy statement.

The Policy Evaluation subsystem therefore returns both the governance decision and the associated governance obligations. While the decision determines whether execution is permitted, denied, or requires approval, the obligations define additional governance actions that must be enforced during runtime. This separation between policy decisions and governance obligations enables UPA to support extensible runtime governance while maintaining a declarative policy model independent of specific execution platforms or enterprise applications.

\[
\begin{aligned}
  \text{Plugin}(s) &= \{ plugin_1, plugin_2, \ldots, plugin_m \} \\
                   &\subseteq \text{Obligations}(s)
\end{aligned}
\]

\subsubsection{Governance Provider Framework}

Enterprise governance requirements evolve continuously across organizations, industries, regulatory frameworks, and deployment environments. Embedding every governance capability within the core Policy Kernel would significantly increase implementation complexity and reduce architectural extensibility. UPA therefore introduces a \textit{Governance Provider Framework}, a provider-based execution model that enables governance capabilities to be integrated independently of the Policy Kernel while preserving a uniform declarative policy model.

The Governance Provider Framework separates governance providers into two categories: \textit{Evaluation Providers} and \textit{Obligation Providers}. Evaluation Providers participate directly in policy evaluation by producing governance attributes required to determine policy decisions. Obligation Providers, in contrast, are executed after policy evaluation as part of the governance obligations generated by matched policy statements.

Let

\[
\Phi
=
\{
\phi_1,\phi_2,\ldots,\phi_n
\}
\]

denote the set of Evaluation Providers, and

\[
\Gamma
=
\{
\gamma_1,\gamma_2,\ldots,\gamma_m
\}
\]

denote the ordered set of governance obligations returned by the Policy Evaluation subsystem.

The Policy Evaluation function is therefore defined as

\[
Eval :
P
\times
\Pi
\times
\Phi
\rightarrow
D
\times
\Gamma
\]

where Evaluation Providers contribute dynamic governance attributes prior to determining the final governance decision.

Evaluation Providers may include threat detection, risk scoring, semantic classification, context enrichment, data sensitivity analysis, or organization-specific evaluation services. These providers execute during policy evaluation and may influence whether a request is allowed, denied, or requires explicit approval.

Obligation Providers execute after policy evaluation under the coordination of the Runtime Governance subsystem. Typical obligations include privacy protection, secret redaction, compliance verification, governance evidence generation, audit logging, notification services, business rule enforcement, and other organization-specific governance capabilities. Since obligation providers are executed declaratively through governance obligations rather than application logic, governance functionality remains externalized from enterprise applications and AI agent implementations.

One important class of Evaluation Providers is the \textit{Threat Detection Provider}. Threat Detection evaluates runtime requests before policy decisions are produced and enables policies to declaratively enforce enterprise security requirements against prompt injection, jailbreak attempts, malicious commands, data exfiltration, and organization-specific threats.

UPA supports multiple threat detection strategies through a common provider abstraction.

\paragraph{Signature-Based Threat Detection}

The Signature-Based Threat Detector performs deterministic threat detection by matching runtime requests against locally maintained YAML signature repositories. Signature repositories contain known malicious patterns including prompt injection templates, jailbreak attempts, destructive operating system commands, dangerous SQL statements, sensitive regular expressions, and organization-specific attack signatures.

For example, a signature repository may include patterns such as

\begin{verbatim}
version: 1

patterns:
  - "rm -rf /"
  - "rm -rf *"
  - "find / -delete"
  - "wipefs -a"
\end{verbatim}

The Signature-Based Threat Detector performs low-latency pattern matching before policy evaluation and returns one or more threat classifications that may be referenced by declarative policies. Because signature matching is deterministic and locally executed, it provides predictable performance suitable for enterprise deployments requiring minimal runtime overhead.

\paragraph{LLM-Based Threat Detection}

While signature matching effectively detects known attack patterns, previously unseen attacks frequently require semantic reasoning beyond deterministic pattern matching. UPA therefore supports an LLM-Based Threat Detector that evaluates the intent and semantic meaning of runtime requests using a foundation model.

The LLM-Based Threat Detector produces a confidence score together with one or more threat categories, such as Prompt Injection, Jailbreak, Data Exfiltration, or Social Engineering. Declarative policies may reference these scores to determine whether execution should be denied or require additional governance actions.

Unlike Signature-Based Threat Detection, semantic detection identifies attacks whose wording differs from previously known signatures while preserving a common provider interface within the Governance Provider Framework.

Formally, the Threat Detection Provider is represented as

\[
\phi_{Threat}
=
\{
\phi_{Signature},
\phi_{LLM}
\}
\]

where

\[
\phi_{Signature}
\]

performs deterministic signature matching and

\[
\phi_{LLM}
\]

performs semantic threat analysis.

The Policy Evaluation subsystem may invoke either provider independently or execute both providers as part of a hybrid evaluation strategy depending on organizational governance policies.

By separating Evaluation Providers from Obligation Providers, the Governance Provider Framework enables enterprises to introduce new governance capabilities without modifying the core Policy Kernel or application logic. This provider-based architecture promotes extensibility, interoperability, and long-term maintainability while preserving deterministic governance semantics across heterogeneous Enterprise AI Operating Systems.

\subsubsection{Runtime Governance}

Policy evaluation determines whether an autonomous operation satisfies the applicable governance policies. Runtime Governance extends this capability by continuously coordinating policy enforcement throughout the complete execution lifecycle of autonomous AI systems. Rather than treating governance as a single authorization decision, UPA views governance as a continuous process that supervises execution before, during, and after runtime operations.

The Runtime Governance subsystem is responsible for orchestrating the governance decision produced by the Policy Evaluation subsystem together with the associated governance obligations. This includes executing policy plugins, initiating approval workflows, generating governance evidence, recording audit events, verifying compliance requirements, enforcing organizational constraints, and coordinating interactions with enterprise enforcement points.

Let

\[
D
\]

denote the governance decision returned by the Policy Evaluation subsystem,

\[
\Gamma
\]

the set of governance obligations, and

\[
R
\]

the set of runtime outcomes.

Runtime Governance is formally defined as

\[
Gov : D \times \Gamma \rightarrow R
\]

where

\[
Gov(d,\Gamma)=r
\]

for

\[
d\in D,
\qquad
\Gamma=\{\gamma_1,\gamma_2,\ldots,\gamma_n\},
\qquad
r\in R.
\]

The set of runtime outcomes is defined as

\[
R = \left\{
\begin{aligned}
  &\texttt{Executed}, \texttt{Blocked}, \\
  &\texttt{PendingApproval}, \texttt{Failed}
\end{aligned}
\right\}.
\]

The runtime outcome is determined as:

\[
r = \begin{cases}
\texttt{Executed}, & \parbox[t]{2.1in}{\raggedright if $d = \texttt{Allow}$ and all governance obligations succeed,}\\[12pt]

\texttt{Blocked},  & \parbox[t]{2.1in}{\raggedright if $d = \texttt{Deny}$,}\\[12pt]

\texttt{PendingApproval}, & \parbox[t]{2.1in}{\raggedright if $d = \texttt{Approve}$,}\\[12pt]

\texttt{Failed},   & \parbox[t]{2.1in}{\raggedright if mandatory governance obligations cannot be completed.}
\end{cases}
\]

Unlike conventional authorization systems, Runtime Governance executes governance obligations continuously throughout autonomous execution. Governance obligations may be triggered before execution, during intermediate workflow stages, or after execution completes depending on the applicable policies and runtime context.

Runtime Governance is therefore responsible for coordinating the complete governance lifecycle, including plugin execution, audit logging, compliance verification, approval orchestration, governance evidence generation, and policy enforcement across heterogeneous enterprise execution environments. This separation between policy evaluation and runtime governance enables the Policy Kernel to remain declarative while allowing enterprises to implement complex governance workflows independently of application logic.

Collectively, the Policy Evaluation and Runtime Governance subsystems transform static authorization decisions into continuous governance processes suitable for Enterprise AI Operating Systems, where autonomous agents dynamically interact with enterprise applications, external tools, memory systems, workflow engines, and organizational resources.

\subsection{Plugin Framework}

Enterprise governance requirements continuously evolve across
industries, regulatory frameworks, organizational policies, and
deployment environments. Embedding every governance capability within
the core Policy Kernel would increase implementation complexity and
limit extensibility. UPA therefore adopts a modular
\textit{Plugin Framework} that enables governance capabilities to be
integrated independently of the core runtime while preserving a
consistent declarative policy model.

Unlike conventional middleware extensions that are invoked directly by
application logic, UPA plugins are resolved and executed declaratively
through governance obligations generated during policy evaluation. A
matched policy may specify one or more plugins that must be executed as
part of Runtime Governance. Consequently, governance behavior remains
externalized from application implementations and is controlled through
declarative policies.

The Plugin Framework is intentionally designed as an open-ended
governance extension mechanism rather than a fixed collection of
governance capabilities. New governance functions can be introduced
without modifying the core Policy Kernel, the DGPL policy model, or the
fundamental policy evaluation semantics.

\subsubsection{Governance Plugin Contract}

Each governance plugin implements a common asynchronous execution
contract. The contract provides the plugin with the canonical policy
request, the current runtime context, and the configuration associated
with the applicable governance obligation.

The abstract plugin interface is represented as:

\begin{lstlisting}[language=Python,
    caption={Standard UPA Governance Plugin Interface},
    label={lst:plugin-interface}
]
class Plugin(ABC):
    async def execute(
        self,
        request,
        context,
        config,
    ):
        ...
\end{lstlisting}

The corresponding execution function is defined conceptually as

\[
\begin{aligned}
\text{Execute}_{\pi} : \; &( \text{Request}, \text{Context}, \text{Configuration} ) \\
&\rightarrow \text{PluginResult}
\end{aligned}
\]

where:

\begin{itemize}

\item \textbf{Request} represents the canonical normalized policy
request.

\item \textbf{Context} represents the current runtime execution
context associated with the request.

\item \textbf{Configuration} represents provider-specific parameters
declared by the applicable policy.

\item \textbf{PluginResult} represents the result produced by the
governance plugin.

\end{itemize}

The asynchronous interface allows a plugin to perform local processing,
invoke external services, access machine learning models, communicate
with enterprise systems, or perform other potentially asynchronous
governance operations without coupling those capabilities to the core
Policy Kernel.

The plugin interface therefore establishes a stable architectural
boundary between the declarative governance layer and the concrete
implementation of governance capabilities.

\subsubsection{Plugin Obligations}

Let

\[
\Gamma =
\left\{
\gamma_1,\gamma_2,\ldots,\gamma_n
\right\}
\]

denote the ordered set of governance obligations generated by the
Policy Evaluation subsystem.

The subset of obligations corresponding to plugin execution is defined
as

\[
\Gamma_P \subseteq \Gamma.
\]

An individual plugin obligation can be represented conceptually as

\[
\gamma_i^P =
\left\langle
provider_i,\ config_i
\right\rangle
\]

where $provider_i$ identifies the governance provider and $config_i$
contains the provider-specific configuration.

For an ordered plugin obligation sequence,

\[
\Gamma_P =
\left\langle
\gamma_1^P,
\gamma_2^P,
\ldots,
\gamma_m^P
\right\rangle,
\]

the Runtime Governance subsystem resolves and executes each obligation
according to the ordering established by the applicable governance
policies.

This design allows a single policy decision to produce multiple
governance obligations without requiring the Policy Kernel to contain
implementation-specific governance logic.

\subsubsection{Plugin Registry}

UPA uses a registry-based provider resolution mechanism to associate
declarative provider identifiers with concrete plugin implementations.

A plugin can be registered using a stable provider identifier:

\begin{lstlisting}[language=Python,
    caption={UPA Plugin Registry Registration},
    label={lst:plugin-registry}
]
@PluginRegistry.decorator("presidio")
class PresidioPlugin(Plugin):
    ...
\end{lstlisting}

The corresponding policy can reference the provider declaratively:

\begin{lstlisting}[language=json,
    caption={Declarative Plugin Provider Configuration},
    label={lst:plugin-policy}
]
"plugins": [
    {
        "provider": "presidio",
        "mode": "redact"
    }
]
\end{lstlisting}

The registry establishes a separation between policy declaration and
implementation. The policy specifies \emph{which governance capability
is required}, while the registry determines \emph{which implementation
provides that capability}.

This separation enables different implementations of the same
governance capability to be introduced without changing the policy
representation.

\subsubsection{Plugin Execution Model}

When a policy containing plugin obligations becomes applicable, the
Runtime Governance subsystem resolves the corresponding providers and
invokes them using the standardized plugin contract.

For an ordered sequence

\[
\Gamma_P =
\left\langle
\gamma_1^P,
\gamma_2^P,
\ldots,
\gamma_m^P
\right\rangle,
\]

the execution of the plugin pipeline can be represented as

\[
C_0
\xrightarrow{\gamma_1^P}
C_1
\xrightarrow{\gamma_2^P}
C_2
\rightarrow
\cdots
\xrightarrow{\gamma_m^P}
C_m,
\]

where $C_0$ represents the runtime context before plugin execution and
$C_m$ represents the resulting context after all applicable plugin
obligations have been processed.

A plugin may operate in an observational mode, in which the context
remains unchanged,

\[
C_{i+1}=C_i,
\]

or in a transformational mode, in which the plugin produces a modified
context,

\[
C_{i+1}=Transform_{\pi_i}(C_i).
\]

This distinction allows governance plugins to perform either
inspection and evidence generation or active transformation of runtime
data.

The overall plugin execution can therefore be represented as

\[
C_m =
Execute_{\gamma_m^P}
\circ
\cdots
\circ
Execute_{\gamma_2^P}
\circ
Execute_{\gamma_1^P}
(C_0).
\]

The resulting context is then supplied to the subsequent governance or
runtime execution stage according to the applicable policy.

\subsubsection{Plugin Result}

A governance plugin returns a structured result describing the outcome
of its execution. Conceptually, the result can be represented as

\[
\begin{aligned}
\text{PluginResult} = \big\langle &\text{Success}, \text{Transformed}, \\
                                  &\text{Data}, \text{Message} \big\rangle
\end{aligned}
\]

The fields represent:

\begin{itemize}

\item \textbf{Success} indicates whether the plugin executed
successfully.

\item \textbf{Transformed} indicates whether the plugin modified the
runtime context.

\item \textbf{Data} contains provider-specific findings, evidence, or
other governance information.

\item \textbf{Message} provides optional diagnostic or execution
information.

\end{itemize}

The distinction between execution success and transformation is
important for governance operations. A plugin may successfully inspect
a request without modifying it, while another plugin may successfully
transform the request before it is forwarded to an AI model or
enterprise system.

\subsubsection{Reference Plugin: Privacy Protection}

The Presidio plugin demonstrates a governance plugin that can detect
and transform sensitive information within an AI request.

The plugin uses a privacy analysis engine to identify sensitive
entities in the runtime prompt. It supports at least two governance
modes:

\begin{itemize}

\item \texttt{detect} --- identifies sensitive entities without
modifying the runtime context.

\item \texttt{redact} --- identifies sensitive entities and transforms
the runtime prompt by replacing or anonymizing the detected values.

\end{itemize}

A policy may declare the provider as follows:

\begin{lstlisting}[language=json,
    caption={Declarative Privacy Redaction Plugin},
    label={lst:presidio-plugin}
]
"plugins": [
    {
        "provider": "presidio",
        "mode": "redact"
    }
]
\end{lstlisting}

In detection mode, the resulting context remains unchanged:

\[
C' = C
\]

while in redaction mode:

\[
C' = Redact_{\text{Presidio}}(C).
\]

The transformed context is subsequently passed to the next governance
stage.

\subsubsection{Reference Plugin: Secret Scanner}

The Secret Scanner plugin demonstrates a deterministic local governance
provider for detecting sensitive credentials and secrets.

The plugin uses locally defined signatures to identify patterns
corresponding to credentials such as cloud access keys, API keys,
repository tokens, and JSON Web Tokens.

The provider supports both detection and redaction modes.

A representative policy configuration is:

\begin{lstlisting}[language=json,
    caption={Declarative Secret Scanner Plugin},
    label={lst:secret-scanner-plugin}
]
"plugins": [
    {
        "provider": "secret_scanner",
        "mode": "redact"
    }
]
\end{lstlisting}

In detection mode, the plugin returns the identified findings while
leaving the runtime context unchanged. In redaction mode, identified
secrets are replaced with typed placeholders before subsequent runtime
execution.

Thus,

\[
C' =
\begin{cases}
C, & \text{if } \text{mode} = \texttt{detect},\\[6pt]
\begin{aligned}
  &\text{Redact}_{\text{SecretScanner}}(C), \\
  &\quad \text{if } \text{mode} = \texttt{redact}.
\end{aligned}
\end{cases}
\]

The Secret Scanner illustrates that governance plugins need not depend
on external services or machine learning models. Deterministic,
locally executed governance mechanisms can participate through the
same standardized plugin contract.

\subsubsection{Plugin Extensibility}

The Plugin Framework is intentionally open-ended. The reference
implementations described above represent examples of the framework
rather than an exhaustive set of supported governance capabilities.

A plugin may implement detection, validation, transformation,
enforcement, evidence generation, external verification, or
organization-specific governance behavior while preserving the same
standardized execution contract.

Consequently, organizations may introduce new governance capabilities
without modifying the Policy Kernel or changing the fundamental DGPL
policy model.

Potential extensions include, but are not limited to:

\begin{itemize}

\item Data Loss Prevention (DLP)

\item Data classification and sensitivity labeling

\item Content safety and policy validation

\item Regulatory compliance verification

\item Model-specific governance controls

\item Data residency enforcement

\item Cost and resource governance

\item Rate and quota enforcement

\item External trust and identity verification

\item Digital signature and integrity verification

\item Enterprise notification and escalation

\item Organization-specific governance controls

\end{itemize}

These capabilities are not assumed to be implemented by the current
reference runtime. Instead, the Plugin Framework provides the
architectural extension boundary through which such capabilities can
be incorporated as independent providers.

This extensibility property is fundamental to UPA because enterprise
governance requirements are expected to evolve as new AI models,
runtime technologies, regulatory requirements, and organizational
controls emerge.

\subsubsection{Plugin and Evaluation Provider Separation}

UPA distinguishes Governance Plugins from Evaluation Providers.

An \textit{Evaluation Provider} contributes evidence or computed
attributes that may influence policy evaluation. A \textit{Governance
Plugin}, in contrast, executes a governance obligation produced by the
policy evaluation process.

The distinction can be represented as:

\[
\begin{aligned}
\text{Evaluation Provider} &\rightarrow \text{Evidence} \\
                           &\rightarrow \text{Policy Evaluation}
\end{aligned}
\]

whereas:

\[
\begin{aligned}
\text{Policy Evaluation} &\rightarrow \text{Governance Obligation} \\
                         &\rightarrow \text{Governance Plugin}.
\end{aligned}
\]

For example, a threat detection provider may produce a prompt-injection
score that is evaluated against a policy threshold. A privacy plugin
such as Presidio may then redact sensitive information when the
applicable governance policy requires such a transformation.

This separation prevents threat analysis, policy evaluation, and
governance enforcement from becoming tightly coupled within the Policy
Kernel.

\subsubsection{Declarative Plugin Execution Example}

Consider the following runtime request:

\begin{tcolorbox}[
    title=Example: Declarative Plugin Execution,
    colback=gray!5,
    colframe=black!40,
    fonttitle=\bfseries
]

\textbf{Runtime Request}

\begin{quote}
Summarize the uploaded employee contract.
\end{quote}

\vspace{0.3em}

\textbf{Semantic Normalization}

\[
\begin{aligned}
N(e) &= \big\langle \text{User}, \texttt{llm:input}, \\
     &\quad \text{EmployeeContract}, \\
     &\quad \text{EnterpriseContext} \big\rangle
\end{aligned}
\]

\vspace{0.3em}

\textbf{Policy Evaluation}

\[
\operatorname{Eval}(p,\Pi) = (\texttt{Allow},\Gamma)
\]

where the plugin obligations are

\[
\begin{aligned}
\Gamma_P = \big\langle &\text{Presidio}(\texttt{redact}), \\
                       &\text{SecretScanner}(\texttt{redact}) \big\rangle.
\end{aligned}
\]

\vspace{0.3em}

\textbf{Runtime Governance}

The Runtime Governance subsystem resolves the configured providers
through the Plugin Registry and executes them against the current
runtime context.

The Presidio plugin identifies and redacts sensitive entities, while
the Secret Scanner identifies and redacts configured secret patterns.

\vspace{0.3em}

\textbf{Execution Pipeline}

\[
\begin{aligned}
\text{Request} &\rightarrow \text{Semantic Normalization} \\
               &\rightarrow \text{Policy Evaluation} \\
               &\rightarrow \text{Plugin Resolution} \\
               &\rightarrow \text{Plugin Execution} \\
               &\rightarrow \text{Sanitized Context} \\
               &\rightarrow \text{LLM Runtime}.
\end{aligned}
\]

\end{tcolorbox}
The example demonstrates that the application itself does not directly
invoke the privacy or secret-scanning implementations. The governance
requirements are declared by policy, resolved through the Plugin
Registry, and enforced by the UPA Runtime Governance subsystem.

This architecture allows governance capabilities to evolve
independently from the applications and AI models that consume them.
\section{Formal Mathematical Foundation}

This section formalizes the execution semantics of the Unified Policy Architecture (UPA). The objective is to provide a platform-independent mathematical model describing how heterogeneous runtime events are transformed into governed execution outcomes. The formal model establishes a common semantic foundation for semantic normalization, policy evaluation, governance obligations, runtime enforcement, and provider-based extensibility across heterogeneous Enterprise AI Operating Systems.

\subsection{Governance Execution Model}

The Unified Policy Architecture (UPA) models enterprise AI governance as a sequence of deterministic transformations that progressively convert heterogeneous runtime events into governed execution outcomes. Unlike traditional authorization systems that evaluate only access permissions, UPA formalizes governance as a continuous runtime process comprising semantic normalization, policy evaluation, governance obligations, and runtime enforcement.

Let $E$ denote the set of heterogeneous runtime events generated by an Enterprise AI Operating System. Runtime events may originate from user requests, autonomous AI agents, workflow engines, memory operations, tool invocations, external APIs, or inter-agent communications.

The governance execution pipeline is formally defined as
\[
E \xrightarrow{N} P \xrightarrow{\text{Eval}} (D,\Gamma) \xrightarrow{\text{Gov}} R
\]
where:
\begin{itemize}[leftmargin=*]
    \item $N : E \rightarrow P$ denotes the Semantic Normalization function.
    \item $P$ denotes the canonical policy request.
    \item $\text{Eval} : P \times \Pi \times \Phi \rightarrow D \times \Gamma$ denotes the Policy Evaluation function.
    \item $\Pi$ denotes the policy repository.
    \item $\Phi$ denotes the set of evaluation providers.
    \item $D$ denotes the governance decision.
    \item $\Gamma$ denotes the ordered set of governance obligations. 
    \item $\text{Gov} : D \times \Gamma \rightarrow R$ denotes the Runtime Governance function.
    \item $R$ denotes the governed runtime outcome.
\end{itemize}

The governance pipeline consists of four sequential stages:
\begin{enumerate}[leftmargin=*]
\item First, the Semantic Normalization function transforms heterogeneous runtime events into a canonical policy request independent of the underlying AI framework, workflow engine, foundation model, or execution environment.

\item Second, the Policy Evaluation function evaluates the normalized request against the applicable declarative governance policies. During this stage, one or more Evaluation Providers may compute additional governance attributes required for policy evaluation, including threat detection, risk scoring, semantic classification, or contextual enrichment.

\item Third, the Policy Evaluation function returns both a governance decision $d \in D$ and an ordered set of governance obligations $\Gamma = \{ \gamma_1, \gamma_2, \ldots, \gamma_n \}$.

\item Finally, the Runtime Governance function coordinates execution by enforcing the governance decision together with the associated governance obligations, thereby producing the governed runtime outcome $r \in R$.
\end{enumerate}

The governance decision space is defined as
\[
D = \{ \texttt{Allow}, \texttt{Deny}, \texttt{Approve} \},
\]
where:
\begin{itemize}[leftmargin=*]
\item \texttt{Allow} permits autonomous execution.
\item \texttt{Deny} terminates execution.
\item \texttt{Approve} suspends execution pending completion of a human approval workflow.
\end{itemize}

The runtime outcome space is defined as
\[
R = \{ \texttt{Executed}, \texttt{Blocked}, \texttt{PendingApproval}, \texttt{Failed} \}.
\]

Unlike traditional authorization architectures that evaluate access permissions only once before execution, the governance execution model treats policy enforcement as a continuous runtime process. This enables UPA to govern autonomous AI systems consistently across heterogeneous execution environments while remaining independent of individual AI frameworks, orchestration platforms, and enterprise applications.

\subsection{Runtime Event Model}

\textbf{Enterprise AI Operating Systems} continuously generate heterogeneous runtime events originating from \textbf{users}, \textbf{LLMs \& SLMs}, \textbf{agentic apps}, \textbf{autonomous agents}, \textbf{workflow engines}, \textbf{ML Models}, \textbf{enterprise applications}, \textbf{memory systems}, \textbf{external tools}, and \textbf{foundation models}. These events differ significantly in structure, semantics, and execution context, making direct policy evaluation impractical.

To provide a unified governance model, UPA represents every execution request as a runtime event belonging to the set $E$. A runtime event is formally represented as
\[
e = \langle \text{src},\, \text{op},\, \text{res},\, \text{ctx},\, \text{meta} \rangle
\]
where $\text{src}$ denotes the source of the event, $\text{op}$ denotes the requested operation, $\text{res}$ denotes the target resource, $\text{ctx}$ denotes the execution context, and $\text{meta}$ denotes additional runtime metadata. Therefore, $e \in E$.

Runtime event sources are defined as $\text{src} \in S$, where
\[
S = \left\{
\begin{aligned}
&\text{User}, \text{Agent}, \text{Workflow}, \text{Tool}, \\
&\text{Memory}, \text{API}, \text{System}
\end{aligned}
\right\}.
\]
These sources collectively represent the heterogeneous execution environment of Enterprise AI Operating Systems.

The requested operation is defined as $\text{op} \in O$, where
\[
O = \left\{
\begin{aligned}
&\text{Prompt}, \text{ToolInvocation}, \text{MemoryRead}, \\
&\text{MemoryWrite}, \text{WorkflowTransition}, \\
&\text{APIInvocation}, \text{AgentCommunication}
\end{aligned}
\right\}.
\]

Resources represent enterprise assets that may be governed by declarative policies, $\text{res} \in R_s$, where typical resources include:
\[
R_s = \left\{
\begin{aligned}
&\text{LLM}, \text{SLM}, \text{ML}, \text{Memory}, \text{Tool}, \text{Database}, \\
&\text{API}, \text{Workflow}, \text{Document}, \text{KnowledgeBase}
\end{aligned}
\right\}.
\]

The execution context contains attributes required during policy evaluation:
\[
\text{ctx} = \langle \text{principal}, \text{environment}, \text{session}, \text{tenant}, \text{timestamp} \rangle.
\]
The execution context may be enriched dynamically by Evaluation Providers during policy evaluation.

Runtime metadata contains implementation-specific information that does not directly influence governance semantics but may contribute to observability, auditing, or governance evidence generation. Typical metadata includes request identifiers, execution identifiers, framework identifiers, model versions, latency metrics, and trace identifiers.

\subsection{Semantic Normalization Function}

Enterprise AI Operating Systems generate heterogeneous runtime events originating from users, autonomous agents, workflow engines, memory systems, external tools, APIs, and foundation models. Since these events differ significantly in structure and semantics, they cannot be evaluated directly by a declarative policy engine.

The Semantic Normalization function transforms heterogeneous runtime events into a canonical policy request representation that is independent of the underlying execution framework. This abstraction enables a single policy model to govern heterogeneous Enterprise AI Operating Systems without requiring framework-specific policy definitions.

Formally, Semantic Normalization is defined as
\[
N : E \rightarrow P
\]
where $e \in E$ is an arbitrary runtime event and $p = N(e)$ is the corresponding canonical policy request. The canonical policy request is represented as
\[
p = \langle \text{pr},\, \text{ac},\, \text{rs},\, \text{ctx},\, \text{meta} \rangle
\]
where $\text{pr}$ denotes the principal, $\text{ac}$ denotes the requested action, $\text{rs}$ denotes the target resource, $\text{ctx}$ denotes the execution context, and $\text{meta}$ denotes runtime metadata.

The Semantic Normalization function satisfies
\[
\forall e \in E,\; \exists!\, p \in P \quad \text{such that} \quad N(e) = p.
\]
\textit{For every runtime event $e$ in the set of runtime events $E$, there exists exactly one canonical policy request $p$ in the set of policy requests $P$. Applying $N$ to event $e$ produces $p$.}

This property guarantees that every runtime event is transformed into a unique canonical policy request suitable for policy evaluation.

The execution context $\text{ctx}$ may contain attributes including identity, environment, workflow state, session information, organizational context, temporal constraints, and tenant information. Additional contextual attributes may be produced dynamically by Evaluation Providers during Policy Evaluation without altering the normalized representation itself.

By separating runtime event interpretation from policy evaluation, UPA enables governance policies to remain independent of specific agent frameworks, orchestration platforms, programming languages, deployment models, or foundation model providers.

\paragraph{Normalization Property.} For every runtime event $e \in E$, Semantic Normalization produces exactly one canonical policy request $p \in P$, thereby ensuring deterministic policy evaluation independent of the originating execution framework. The output of Semantic Normalization is a canonical policy request consisting of the Principal, Action, Resource, and Context (PARC).

\subsection{Policy Evaluation Function}

The Policy Evaluation function determines whether a canonical policy request satisfies the applicable governance policies defined by an organization. Unlike traditional authorization engines that evaluate a single policy in isolation, UPA evaluates a collection of declarative policies, incorporates dynamic governance attributes produced by Evaluation Providers, resolves policy conflicts deterministically, and produces both a governance decision and an ordered set of governance obligations.

Formally, let $P$ denote the set of canonical policy requests, $\Pi$ the policy repository, and $\Phi$ the set of Evaluation Providers. The Policy Evaluation function is defined as
\[
\text{Eval} : P \times \Pi \times \Phi \rightarrow D \times \Gamma
\]
where $\text{Eval}(p,\Pi,\Phi) = (d,\Gamma)$ for $p \in P$.

The evaluation process consists of four sequential stages:
\begin{enumerate}[leftmargin=*]
\item Policy Matching
\item Evaluation Provider Execution
\item Conflict Resolution
\item Governance Decision Generation
\end{enumerate}

\subsubsection{Policy Matching}

Given a canonical policy request $p \in P$, the Policy Matching function identifies all policy statements whose conditions are satisfied:
\[
\text{Match} : P \times \Pi \rightarrow M \quad \text{where} \quad M = \{ s_1, s_2, \ldots, s_n \}
\]
denotes the set of matched policy statements.

Unlike traditional authorization systems that frequently terminate after locating the first matching rule, UPA evaluates all applicable policy statements to support policy composition, layered governance, compliance verification, and organizational policy inheritance.

\subsubsection{Evaluation Providers}

Certain governance decisions cannot be determined solely from the canonical policy request. Enterprise AI systems frequently require additional runtime information such as threat assessments, semantic classifications, risk scores, contextual attributes, or organization-specific intelligence before a governance decision can be produced.

UPA introduces \textit{Evaluation Providers}, a collection of pluggable evaluation services that compute dynamic governance attributes during policy evaluation. Let $\Phi = \{ \phi_1, \phi_2, \ldots, \phi_n \}$ denote the set of Evaluation Providers. Each provider is represented as
\[
\phi_i : P \rightarrow A_i
\]
where $A_i$ denotes the set of governance attributes produced by the provider.

For a canonical policy request $p \in P$, the collective provider output is defined as
\[
A = \bigcup_{\phi_i \in \Phi} \phi_i(p) = \{ a_1, a_2, \ldots, a_m \},
\]
which represents the complete set of governance attributes available during policy evaluation.

By separating runtime attribute generation from declarative policy evaluation, UPA enables organizations to introduce new evaluation capabilities without changing the Policy Kernel or existing policy semantics.

\subsubsection{Conflict Resolution}

Policy evaluation frequently produces multiple matching policy statements that may express different governance decisions or runtime obligations. Let $M = \{ s_1, s_2, \ldots, s_n \}$ denote the set of matched policy statements. Conflict resolution is formally defined as
\[
\text{Resolve} : M \rightarrow D \times \Gamma \quad \text{where} \quad \text{Resolve}(M) = (d,\Gamma).
\]

The conflict resolution process evaluates matched policies using a deterministic precedence strategy consisting of:
\begin{enumerate}[leftmargin=*]
\item Policy Priority
\item Policy Effect
\item Policy Specificity
\item Policy Version
\end{enumerate}

Policy priority is evaluated first, with higher priority values taking precedence. When multiple policies have the same priority, governance effects are evaluated according to the precedence
\[
\texttt{Deny} > \texttt{Approve} > \texttt{Allow}.
\]
If multiple policies remain applicable, the most specific policy is selected. Policy version is considered only when all preceding precedence rules produce equivalent results.

\subsubsection{Governance Decision}

The final governance decision returned by the Policy Evaluation function belongs to the decision space $D = \{ \texttt{Allow}, \texttt{Deny}, \texttt{Approve} \}$. The resulting decision is defined as:
\[
d = \begin{cases}
\texttt{Allow}, & \parbox[t]{2.1in}{\raggedright if all applicable governance conditions are satisfied,}\\[10pt]
\texttt{Deny},  & \parbox[t]{2.1in}{\raggedright if one or more mandatory governance policies prohibit execution,}\\[10pt]
\texttt{Approve}, & \parbox[t]{2.1in}{\raggedright if execution requires explicit organizational approval before continuation.}
\end{cases}
\]

Unlike conventional authorization engines, UPA introduces \texttt{Approve} as a first-class declarative governance decision, enabling Human-in-the-Loop governance by suspending execution until approval completes.

\subsubsection{Governance Obligations}

In addition to the governance decision, Policy Evaluation returns an ordered set of governance obligations $\Gamma = \{ \gamma_1, \gamma_2, \ldots, \gamma_n \}$. Governance obligations describe runtime actions that must be coordinated by the Runtime Governance subsystem following policy evaluation, such as plugin execution, audit logging, compliance verification, evidence generation, notification services, and workflow metadata.

Unlike Evaluation Providers, governance obligations execute \emph{after} the decision has been determined.

\subsection{Runtime Governance Function}

The Runtime Governance function is responsible for enforcing the governance decision together with the associated governance obligations.

Formally, the Runtime Governance function is defined as
\[
\text{Gov} : D \times \Gamma \rightarrow R \quad \text{where} \quad \text{Gov}(d,\Gamma) = r,
\]
with $d \in D$, $\Gamma = \{ \gamma_1, \gamma_2, \ldots, \gamma_n \}$, and $r \in R$.

The runtime outcome space is defined as
\[
r = \begin{cases}
\texttt{Executed}, & \parbox[t]{1.8in}{\raggedright if $d = \texttt{Allow}$ and all governance obligations succeed,}\\[14pt]
\texttt{Blocked},  & \parbox[t]{1.8in}{\raggedright if $d = \texttt{Deny}$,}\\[8pt]
\texttt{PendingApproval}, & \parbox[t]{1.8in}{\raggedright if $d = \texttt{Approve}$,}\\[8pt]
\texttt{Failed},   & \parbox[t]{1.8in}{\raggedright if mandatory governance obligations cannot be completed.}
\end{cases}
\]

For $d = \texttt{Allow}$, Runtime Governance executes obligations in $\Gamma$ before execution continues. For $d = \texttt{Deny}$, execution is terminated immediately. For $d = \texttt{Approve}$, execution is suspended pending workflow completion.

\subsection{Properties of the Governance Model}

The formal execution model of UPA satisfies several key properties:

\paragraph{Property 1 (Deterministic Normalization).}
For every runtime event $e \in E$, there exists exactly one canonical policy request $p \in P$ such that $N(e) = p$. Identical runtime events always produce identical canonical representations independent of the originating framework.

\paragraph{Property 2 (Deterministic Policy Evaluation).}
Given an identical request $p$, repository $\Pi$, and provider outputs, $\text{Eval}(p,\Pi,\Phi)$ always produces the same governance decision $d$ and obligations $\Gamma$.

\paragraph{Property 3 (Conflict Resolution Consistency).}
Whenever multiple policy statements match, UPA resolves conflicts using a deterministic precedence model ($\text{Priority} > \text{Effect} > \text{Specificity} > \text{Version}$).

\paragraph{Property 4 (Governance Completeness).}
Every canonical policy request $p \in P$ produces exactly one governance decision $d \in D$, ensuring policy evaluation always terminates cleanly.

\paragraph{Property 5 (Framework Independence).}
Governance decisions depend exclusively on the canonical policy request rather than the originating framework or language.

\paragraph{Property 6 (Extensibility).}
Evaluation Providers and Obligation Providers may be added or modified without changing the Policy Kernel or existing policy semantics.

\section{Declarative Governance Policy Language}

The Unified Policy Architecture introduces the
\textit{Declarative Governance Policy Language (DGPL)}, a
platform-independent policy language for expressing enterprise AI
governance. DGPL enables governance requirements to be specified
declaratively rather than embedded within application logic, AI agent
implementations, or runtime-specific control mechanisms.

Unlike traditional authorization languages that primarily express access
control decisions, DGPL provides a unified policy model for
authorization, AI safety, runtime governance, threat detection,
governance obligations, human approval workflows, and extensible
enterprise governance.

The design of DGPL is guided by the following principles:

\begin{enumerate}

\item \textbf{Declarative Governance.}
Policies specify governance requirements and desired governance outcomes
rather than procedural execution logic.

\item \textbf{Platform Independence.}
Policies operate on canonical policy requests and remain independent of
specific AI frameworks, programming languages, orchestration platforms,
and model providers.

\item \textbf{Extensibility.}
Evaluation Providers and Governance Plugins can be introduced without
modifying the core Policy Kernel or the fundamental DGPL model.

\item \textbf{Deterministic Evaluation.}
Equivalent runtime requests evaluated against an equivalent policy
repository produce deterministic governance outcomes.

\item \textbf{Human Readability.}
Policies use a concise and structured representation that can be
understood by policy authors and processed by automated systems.

\item \textbf{Machine Processability.}
DGPL can be serialized using structured formats such as JSON or YAML
while preserving a common underlying semantic model.

\item \textbf{Enterprise Composability.}
Multiple policy statements may be composed to express layered
governance spanning authorization, AI safety, runtime controls,
compliance, approval workflows, and organization-specific requirements.

\end{enumerate}

The DGPL specification defines the logical policy model independently
of its concrete serialization format. JSON and YAML are therefore
representations of the same underlying declarative governance semantics.


\subsection{Policy Document Model}

A DGPL policy document represents a collection of declarative governance
statements evaluated by the UPA Policy Kernel.

A policy document consists of a language version and one or more policy
statements:

\[
Policy =
\left\langle
Version,\ Statements
\right\rangle
\]

where $Version$ identifies the DGPL specification version and
$Statements$ denotes the collection of policy statements contained in
the document.

The statement collection is represented as

\[
Statements =
\left\{
S_1,S_2,\ldots,S_n
\right\}.
\]

Each statement defines an independent declarative governance rule.

A policy document may therefore contain multiple statements that
collectively express authorization, contextual constraints, threat
detection requirements, governance obligations, approval workflows,
compliance requirements, and organization-specific controls.

The policy document does not contain procedural execution logic.
Instead, the UPA Policy Kernel interprets the statements according to
the governance execution semantics defined in Section~5.

A minimal DGPL policy document can be represented as:

\begin{lstlisting}[language=json,
    caption={Generic DGPL Policy Document},
    label={lst:dgpl-policy-document}
]
{
  "version": "2026-01-01",
  "statements": [
    ...
  ]
}
\end{lstlisting}

The \texttt{version} field identifies the language specification version
and enables the language to evolve while maintaining compatibility
across policy implementations.

The \texttt{statements} array contains the independent governance
statements. During policy evaluation, the UPA Policy Kernel evaluates
the applicable statements against a single canonical policy request.

The logical policy model is independent of JSON serialization.
Consequently, equivalent DGPL policies may be represented using JSON,
YAML, or another structured serialization format without changing their
underlying semantics.


\subsection{Policy Statement Model}

A Policy Statement represents the fundamental governance unit within
DGPL. Each statement defines an independent declarative rule that may
authorize, deny, approve, or impose governance requirements on a runtime
operation.

A policy statement is formally represented as

\[
S =
\left\langle
SID,\ Effect,\ PARC,\ Extensions
\right\rangle
\]

where:

\begin{itemize}

\item $SID$ identifies the policy statement.

\item $Effect$ specifies the governance decision associated with the
statement.

\item $PARC$ represents the Principal--Action--Resource--Context
authorization model.

\item $Extensions$ represents optional governance capabilities
associated with the statement.

\end{itemize}

The three governance effects defined by DGPL are:

\[
Effect \in
\{
\texttt{Allow},
\texttt{Deny},
\texttt{Approve}
\}.
\]

The effect semantics are:

\begin{itemize}

\item \textbf{Allow} permits execution subject to applicable governance
obligations.

\item \textbf{Deny} prohibits execution.

\item \textbf{Approve} suspends autonomous execution and requires the
specified approval workflow to complete before execution can continue.

\end{itemize}

A statement may contain only the fields required for its governance
purpose. For example, a general authorization policy may specify
Principal, Action, and Resource without a Context condition, while a
runtime governance policy may additionally specify Evaluation
Providers or Governance Plugins.


\subsection{PARC Model}

DGPL adopts the
\textbf{Principal--Action--Resource--Context (PARC)} model as its core
authorization abstraction.

Formally,

\[
PARC =
\left\langle
Principal,\ Action,\ Resource,\ Context
\right\rangle.
\]

PARC provides a framework-independent representation of the
authorization relationship between a runtime request and a governance
policy.

\subsubsection*{Principal}

The Principal identifies the subject initiating the runtime request.

A Principal may represent:

\begin{itemize}
\item a human user,
\item an autonomous AI agent,
\item an application,
\item an enterprise service,
\item an infrastructure component, or
\item another authorized execution identity.
\end{itemize}

When the Principal element is omitted, the statement may apply to all
principals, subject to the remaining matching criteria.

\subsubsection*{Action}

The Action identifies the operation being performed.

DGPL actions may represent operations such as:

\begin{itemize}
\item LLM input or output operations,
\item tool invocations,
\item API invocations,
\item workflow transitions,
\item memory reads and writes,
\item RAG queries, and
\item organization-specific operations.
\end{itemize}

Actions are represented using standardized identifiers such as
\texttt{llm:input}, \texttt{rag:query}, and \texttt{db:delete}.

\subsubsection*{Resource}

The Resource identifies the enterprise asset against which the action
is performed.

Resources may represent:

\begin{itemize}
\item foundation models,
\item AI agents,
\item enterprise knowledge bases,
\item APIs,
\item databases,
\item workflows,
\item tools,
\item memory systems, or
\item other enterprise resources.
\end{itemize}

Examples include \texttt{knowledgebase:hr},
\texttt{database:production}, and \texttt{*}.

\subsubsection*{Context}

The Context represents runtime information used to determine whether
additional governance predicates are satisfied.

Context may include temporal attributes, environmental attributes,
organizational attributes, user attributes, resource attributes, or
facts generated by Evaluation Providers.

For example, a policy may require:

\[
system.hour \in [9,17].
\]

Context therefore enables governance policies to depend on runtime
conditions without embedding procedural logic into the policy.

\subsubsection*{Governance Extensions}

Beyond PARC, DGPL provides extensible Governance Extensions.

These include:

\begin{itemize}

\item Evaluation Providers, such as Threat Detection,

\item Governance Plugins, such as PII redaction and secret scanning,

\item Human Approval Workflows,

\item and future organization-specific governance extensions.

\end{itemize}

This separation allows the PARC model to remain stable while enabling
the governance language to evolve with enterprise AI requirements.

\begin{tcolorbox}[
title=\textbf{Example: PARC Authorization Policy},
colback=gray!5,
colframe=black!60,
fonttitle=\bfseries
]

Consider the policy requirement:

\begin{quote}
\emph{Allow employees to query the HR knowledge base.}
\end{quote}

The corresponding PARC representation is:

\[
\begin{aligned}
Principal &= Employee \\
Action &= \texttt{rag:query} \\
Resource &= \texttt{knowledgebase:hr} \\
Context &= \text{None}
\end{aligned}
\]

The corresponding DGPL statement is:

\begin{lstlisting}[language=json]
{
  "sid": "HRKnowledgeAccess",
  "effect": "Allow",
  "principal": {
    "Role": ["Employee"]
  },
  "action": [
    "rag:query"
  ],
  "resource": [
    "knowledgebase:hr"
  ]
}
\end{lstlisting}

The example demonstrates that the core authorization semantics are
captured by PARC, while additional governance capabilities can be
introduced through Governance Extensions.

\end{tcolorbox}


\subsection{Governance Extensions}

While PARC captures the core authorization semantics of DGPL,
Enterprise AI Operating Systems require governance capabilities beyond
traditional access control.

DGPL therefore introduces \textit{Governance Extensions} that augment
the core policy model without modifying its fundamental PARC structure.

Governance Extensions provide declarative mechanisms for expressing:

\begin{itemize}
\item contextual constraints,
\item threat detection,
\item runtime governance obligations,
\item human approval workflows, and
\item future organization-specific governance capabilities.
\end{itemize}


\subsubsection{Conditions}

Conditions specify additional predicates that must evaluate to
\texttt{true} before a policy statement becomes applicable.

A condition consists conceptually of:

\[
Condition =
\left\langle
Fact,\ Operator,\ Value
\right\rangle.
\]

For example:

\begin{lstlisting}[language=json]
"condition": {
  "all": [
    {
      "fact": "system.hour",
      "operator": "between",
      "value": [9, 17]
    }
  ]
}
\end{lstlisting}

Multiple conditions may be combined using logical operators such as
\texttt{all} and \texttt{any}.

Conditions enable context-aware governance without introducing
procedural execution logic into the policy.


\subsubsection{Threat Detection}

Threat Detection is an Evaluation Provider that enables DGPL policies
to incorporate runtime security analysis into governance decisions.

DGPL supports two principal classes of Threat Detectors:

\begin{enumerate}

\item \textbf{Static Threat Detectors}

\item \textbf{LLM-Based Threat Detectors}

\end{enumerate}

Static Threat Detectors detect known threats by matching runtime
content against locally maintained signatures or pattern definitions.

A representative static signature configuration may contain:

\begin{lstlisting}
version: 1

patterns:

  - "rm -rf /"
  - "rm -rf *"
  - "rm -rf .*"
  - "rm -rf ~"
  - "rm -fr /"
  - "find / -delete"
  - "shred -u"
  - "wipefs -a"
\end{lstlisting}

Static detection provides deterministic and low-latency detection for
known patterns.

LLM-Based Threat Detectors perform semantic analysis using a foundation
model. They are intended to identify threats that cannot be reliably
detected using static signatures alone, including prompt injection,
jailbreak attempts, indirect prompt attacks, and emerging semantic
attack patterns.

A DGPL statement may declare an LLM-based threat detector as:

\begin{lstlisting}[language=json]
{
  "sid": "PromptInjection",
  "effect": "Deny",
  "action": [
    "llm:input_review"
  ],
  "resource": [
    "*"
  ],
  "threat": {
    "provider": "llm",
    "model": "gemini/gemini-2.5-flash",
    "threshold": 0.80,
    "categories": [
      "PromptInjection",
      "Jailbreak"
    ]
  }
}
\end{lstlisting}

The output of Threat Detection becomes a governance attribute that can
participate in Policy Evaluation.

Conceptually,

\[
ThreatDetect(e)
\rightarrow
ThreatFacts
\rightarrow
PolicyEvaluation.
\]

This distinguishes Threat Detection from Governance Plugins: the
former contributes evidence to policy evaluation, while the latter
executes governance obligations produced by the policy decision.


\subsubsection{Governance Plugins}

Governance Plugins represent extensible governance obligations that are
executed by the Runtime Governance subsystem.

A policy may declaratively specify one or more plugins:

\begin{lstlisting}[language=json]
{
  "sid": "EnterpriseProtection",
  "effect": "Allow",
  "action": [
    "llm:input"
  ],
  "resource": [
    "*"
  ],
  "plugins": [
    {
      "provider": "presidio",
      "mode": "redact"
    },
    {
      "provider": "secret_scanner",
      "mode": "redact"
    }
  ]
}
\end{lstlisting}

The plugin framework is intentionally open-ended. Presidio and Secret
Scanner are reference implementations rather than an exhaustive set
of governance capabilities.

Potential future governance plugins may provide capabilities such as:

\begin{itemize}
\item data loss prevention,
\item data classification,
\item compliance verification,
\item content governance,
\item audit and evidence generation,
\item notification and escalation,
\item data residency enforcement,
\item cost governance,
\item resource and quota controls,
\item organization-specific governance controls.
\end{itemize}

The Plugin Framework therefore provides a stable extension boundary
through which new governance capabilities can be introduced without
modifying the Policy Kernel or the fundamental DGPL policy model.


\subsubsection{Approval Workflows}

DGPL introduces \texttt{Approve} as a first-class governance effect for
Human-in-the-Loop execution.

A policy may require organizational approval for high-risk operations:

\begin{lstlisting}[language=json]
{
  "sid": "ProductionDelete",
  "effect": "Approve",
  "action": [
    "db:delete"
  ],
  "resource": [
    "database:production"
  ],
  "approval": {
    "role": "Admin"
  }
}
\end{lstlisting}

When the applicable governance decision is \texttt{Approve}, autonomous
execution is suspended and the required approval workflow is initiated.

Conceptually,

\[
\begin{aligned}
\texttt{Approve} &\rightarrow \text{ApprovalWorkflow} \\
                 &\rightarrow \begin{cases}
                   \texttt{Approved} \rightarrow \text{ContinueExecution} \\[4pt]
                   \texttt{Rejected} \rightarrow \text{DenyExecution}
                 \end{cases}
\end{aligned}
\]

Approval workflows are particularly applicable to financial
transactions, production changes, privileged administrative operations,
and other high-risk enterprise activities.


\subsection{Policy Composition and Conflict Resolution}

A DGPL policy repository may contain multiple independent policy
statements. A single canonical policy request may therefore match
multiple statements simultaneously.

The Policy Kernel evaluates the statements collectively rather than
terminating after the first matching policy.

Given a canonical policy request $p$, the matched policy set is:

\[
M(p)=
\left\{
S_i \mid Match(S_i,p)=true
\right\}.
\]

The resulting matched policy set may contain statements contributing
different governance requirements.

For example, a single request may match:

\begin{center}
\small 
\begin{tabular}{ll}
\hline
\textbf{Policy} & \textbf{Governance Role} \\ \hline
AllowLLMRequests & Authorization \\
BusinessHours & Contextual Constraint \\
EnterpriseProtection & Governance Plugin \\
PromptInjection & Threat Detection \\ \hline
\end{tabular}
\end{center}

Policy composition therefore allows authorization, security, runtime
governance, and compliance requirements to be expressed independently.

\subsubsection{Policy Matching}

Each policy statement is evaluated independently against the same
canonical policy request.

A statement is applicable when its declared PARC attributes and
conditions match the request.

The Policy Kernel therefore computes:

\[
M =
\left\{
S_1,S_2,\ldots,S_m
\right\}.
\]

Only the statements contained in $M$ participate in subsequent
governance evaluation.

\subsubsection{Conflict Resolution}

When multiple applicable statements specify different governance
effects, DGPL requires deterministic conflict resolution.

The governance effects have the following precedence:

\[
\texttt{Deny}
>
\texttt{Approve}
>
\texttt{Allow}.
\]

Thus, when a request simultaneously matches an \texttt{Allow} and a
\texttt{Deny} statement, the resulting governance decision is
\texttt{Deny}.

Similarly, an applicable \texttt{Approve} requirement takes precedence
over an \texttt{Allow} requirement when no higher-precedence
\texttt{Deny} applies.

Policy specificity is used to distinguish policies when multiple
statements with the same effect apply at different levels of scope.
More specific resource, action, or principal constraints take
precedence over broader wildcard policies.

The conflict resolution semantics are summarized as:

\[
Decision =
Resolve(M).
\]

The resolution function is deterministic:

\[
Resolve(M)=d
\]

where

\[
d \in
\{
\texttt{Allow},
\texttt{Deny},
\texttt{Approve}
\}.
\]

The separation of Policy Composition from Conflict Resolution allows
multiple independent policies to participate in governance while
preserving a single deterministic governance decision.

\subsubsection{Default Deny}

DGPL adopts a default-deny governance model.

If no policy statement matches the canonical policy request, the Policy
Kernel returns:

\[
Decision=\texttt{Deny}.
\]

Default deny ensures that autonomous execution is permitted only when
explicitly authorized by the applicable governance policies.

\begin{tcolorbox}[
title=Example: Policy Composition,
colback=gray!5,
colframe=black!60,
fonttitle=\bfseries
]

Consider a request:

\begin{quote}
\emph{Summarize the uploaded employee contract.}
\end{quote}

The request may simultaneously match:

\begin{center}
\small
\begin{tabularx}{\linewidth}{@{} l X @{}}
\toprule
\textbf{Statement} & \textbf{Effect / Obligation} \\ \midrule
\texttt{AllowLLMRequests} & \texttt{Allow} \\
\texttt{EnterpriseProtection} & Presidio + Secret Scanner \\
\texttt{AuditPolicy} & Audit obligation \\ \bottomrule
\end{tabularx}
\end{center}

The Policy Kernel does not select the first matching statement.
Instead, all applicable statements participate in the governance
evaluation. The resulting decision and governance obligations are then
enforced by Runtime Governance.

\end{tcolorbox}


\subsection{Expression Semantics}

DGPL conditions provide a declarative mechanism for expressing
context-dependent policy predicates. This section defines the basic
semantics of DGPL expressions independently of their concrete JSON or
YAML representation.

\subsubsection{Facts}

A \textit{Fact} identifies a runtime attribute used during condition
evaluation.

Formally,

\[
Fact :
Context \rightarrow Value.
\]

For example:

\[
Fact(\texttt{system.hour},C)=14.
\]

A fact may originate from the canonical policy request, runtime
context, enterprise context, or an Evaluation Provider.

\subsubsection{Primitive Expressions}

A primitive DGPL expression can be represented as:

\[
e =
\left\langle
f,o,v
\right\rangle
\]

where:

\begin{itemize}
\item $f$ is a fact,
\item $o$ is an operator, and
\item $v$ is the comparison value.
\end{itemize}

The expression evaluator is defined as:

\[
Eval(e,C)
\rightarrow
\{\texttt{true},\texttt{false}\}.
\]

Representative operators include:

\begin{itemize}
\item \texttt{equals}
\item \texttt{not\_equals}
\item \texttt{in}
\item \texttt{contains}
\item \texttt{between}
\item \texttt{greater\_than}
\item \texttt{less\_than}
\end{itemize}

\subsubsection{Between Operator}

For a fact $f$, a \texttt{between} expression with bounds
$[l,u]$ evaluates to true when:

\[
Eval(
\langle f,\texttt{between},[l,u]\rangle,C)
=
true
\]

iff

\[
l \leq f(C) \leq u.
\]

For example:

\begin{lstlisting}[language=json]
{
  "fact": "system.hour",
  "operator": "between",
  "value": [9, 17]
}
\end{lstlisting}

is satisfied when the runtime hour is within the declared interval.

\subsubsection{Logical Expressions}

DGPL supports logical composition of expressions.

For an \texttt{all} expression:

\[
Eval(
all(e_1,\ldots,e_n),C
)
=
true
\]

iff

\[
\forall i,\ Eval(e_i,C)=true.
\]

For an \texttt{any} expression:

\[
Eval(
any(e_1,\ldots,e_n),C
)
=
true
\]

iff

\[
\exists i,\ Eval(e_i,C)=true.
\]

This permits multiple runtime conditions to be expressed declaratively
without introducing procedural control flow.

\subsubsection{Condition Evaluation}

A policy condition $c$ is satisfied when:

\[
Eval(c,C)=true.
\]

The condition therefore contributes to policy applicability:

\[
Applicable(S,p)
=
PARCMatch(S,p)
\land
Eval(Condition_S,C).
\]

When a statement does not specify a condition, the condition component
is considered satisfied by default.


\subsection{Reference Policy Examples}

DGPL supports a broad range of enterprise governance scenarios through
the composition of PARC semantics and Governance Extensions.

Representative policies include:

\begin{itemize}

\item Basic authorization policies using the PARC model.

\item Context-aware policies using temporal or environmental
conditions.

\item Threat Detection policies using static or LLM-based Evaluation
Providers.

\item Governance policies invoking runtime plugins for privacy
protection and secret detection.

\item Human Approval policies using the \texttt{Approve} effect.

\item Multi-statement policies combining authorization, threat
detection, conditions, and governance obligations.

\end{itemize}

\subsubsection{Business Hours Policy}

A policy may restrict execution to a defined time window:

\begin{lstlisting}[language=json]
{
  "version": "2026-01-01",
  "statements": [
    {
      "sid": "BusinessHours",
      "effect": "Allow",
      "action": [
        "*"
      ],
      "resource": [
        "*"
      ],
      "condition": {
        "all": [
          {
            "fact": "system.hour",
            "operator": "between",
            "value": [9, 17]
          }
        ]
      }
    }
  ]
}
\end{lstlisting}

\subsubsection{Human Approval Policy}

A high-risk database operation may require explicit approval:

\begin{lstlisting}[language=json]
{
  "version": "2026-01-01",
  "statements": [
    {
      "sid": "ProductionDelete",
      "effect": "Approve",
      "action": [
        "db:delete"
      ],
      "resource": [
        "database:production"
      ],
      "approval": {
        "role": "Admin"
      }
    }
  ]
}
\end{lstlisting}

\subsubsection{Runtime Protection Policy}

Sensitive information may be protected through declarative governance
plugins:

\begin{lstlisting}[language=json]
{
  "version": "2026-01-01",
  "statements": [
    {
      "sid": "EnterpriseProtection",
      "effect": "Allow",
      "action": [
        "llm:input"
      ],
      "resource": [
        "*"
      ],
      "plugins": [
        {
          "provider": "presidio",
          "mode": "redact"
        },
        {
          "provider": "secret_scanner",
          "mode": "redact"
        }
      ]
    }
  ]
}
\end{lstlisting}

\subsubsection{Threat Detection Policy}

A policy may deny an LLM request when an LLM-based threat detector
identifies a prompt injection or jailbreak above a defined threshold:

\begin{lstlisting}[language=json]
{
  "version": "2026-01-01",
  "statements": [
    {
      "sid": "PromptInjection",
      "effect": "Deny",
      "action": [
        "llm:input_review"
      ],
      "resource": [
        "*"
      ],
      "threat": {
        "provider": "llm",
        "model": "gemini/gemini-2.5-flash",
        "threshold": 0.80,
        "categories": [
          "PromptInjection",
          "Jailbreak"
        ]
      }
    },
    {
      "sid": "AllowLLMRequests",
      "effect": "Allow",
      "action": [
        "llm:input_review"
      ],
      "resource": [
        "*"
      ]
    }
  ]
}
\end{lstlisting}

In this example, the two statements may match the same canonical
request. The threat evaluation produces a threat result that
participates in Policy Evaluation. If the configured threat condition
is satisfied, the applicable \texttt{Deny} statement takes precedence
over the general \texttt{Allow} statement.

\subsubsection{Role-Based RAG Authorization}

DGPL can also express role-based access to enterprise knowledge:

\begin{lstlisting}[language=json]
{
  "version": "2026-01-01",
  "statements": [
    {
      "sid": "WiFiAccessForITAdmin",
      "effect": "Allow",
      "principal": {
        "Role": [
          "ITAdmin"
        ]
      },
      "action": [
        "rag:query"
      ],
      "resource": [
        "knowledgebase:wifi"
      ]
    },
    {
      "sid": "HRAccessForEmployeesAndAdmins",
      "effect": "Allow",
      "principal": {
        "Role": [
          "Employee"
        ]
      },
      "action": [
        "rag:query"
      ],
      "resource": [
        "knowledgebase:hr"
      ]
    }
  ]
}
\end{lstlisting}

These examples demonstrate that the same declarative language can
express authorization, contextual governance, threat detection,
runtime transformations, and human approval without embedding
governance logic within the application implementation.

The reference policies in Appendix~A provide additional examples of
DGPL constructs and their corresponding governance behavior.
\section{Industry Policy Packs}
\subsection{Industry Policy Pack Model}

Enterprise AI governance requirements vary across industries because
organizations operate under different business processes, risk models,
regulatory requirements, and operational constraints. A policy that is
appropriate for a general enterprise environment may be insufficient
for healthcare, financial services, manufacturing, or critical
infrastructure.

Writing these controls independently for every application can result
in fragmented authorization logic, duplicated governance mechanisms,
and inconsistent enforcement. UPA addresses this problem through
\textit{Industry Policy Packs}.

\textit{\textbf{Industry Policy Pack}} is a pre-built, versioned
collection of DGPL policy statements designed for a particular industry
or governance domain. A pack provides reusable authorization rules,
contextual conditions, governance obligations, approval requirements,
and provider bindings that can be incorporated into an organization's
existing policy repository.

Industry Policy Packs do not introduce a separate policy language or
enforcement mechanism. Instead, they provide reusable DGPL statements
that are evaluated by the same Policy Kernel used for organization-
specific policies.

Consequently, an organization can adopt domain-specific governance
without modifying the Policy Kernel, Semantic Normalization, Policy
Evaluation, Conflict Resolution, or Runtime Governance mechanisms.

\subsection{Formal Definition}

An Industry Policy Pack is formally represented as

\[
\mathit{IndustryPack}
=
\left\langle
\mathit{Domain},
\mathit{Version},
\mathit{Statements}
\right\rangle
\]

where

\begin{itemize}

\item $\mathit{Domain}$ identifies the industry or governance domain
for which the pack is intended, such as Healthcare, Banking,
Manufacturing, or Oil and Gas.

\item $\mathit{Version}$ identifies the version of the policy pack and
allows the pack to evolve independently while maintaining reproducible
governance configurations.

\item $\mathit{Statements}$ is the collection of DGPL policy statements
contained in the pack.

\end{itemize}

The statements are ordinary DGPL statements and therefore follow the
policy statement model defined in Section~6.

A policy pack is incorporated into an organization's policy repository
by adding its statements to the existing policy set:

\[
\Pi =
\Pi_{\mathrm{core}}
\cup
\bigcup_i
\mathit{Pack}_i.\mathit{Statements}
\]

where $\Pi_{\mathrm{core}}$ represents organization-specific policies
and $\mathit{Pack}_i.\mathit{Statements}$ represents the statements
provided by the $i$-th policy pack.

Once incorporated, the Policy Kernel does not distinguish between a
policy authored internally and one originating from an Industry Policy
Pack.

The resulting evaluation follows the same governance pipeline:

\[
\begin{aligned}
N(e)
&\rightarrow
\text{Policy Matching}
\rightarrow
\Phi
\\
&\rightarrow
\text{Evaluation Providers}
\rightarrow
\text{Conflict Resolution}
\\
&\rightarrow
(d,\Gamma)
\rightarrow
\mathit{Gov}(d,\Gamma).
\end{aligned}
\]

This design allows multiple policy packs to coexist within the same
policy repository. For example, an organization operating a healthcare
business with an internal financial-services division may simultaneously
activate Healthcare and Finance policy packs.

No industry-specific evaluation path is required.

\subsection{Anatomy of an Industry Policy Pack}

An Industry Policy Pack is constructed using the governance constructs
already defined by DGPL.

The principal components are:

\begin{itemize}

\item \textbf{Authorization Statements} --- PARC-based statements that
define which principals may perform specific actions on specific
resources.

\item \textbf{Conditions} --- contextual predicates that restrict
policy applicability based on runtime facts such as business state,
environment, temporal constraints, or organizational context.

\item \textbf{Governance Obligations} --- obligations generated by
policy evaluation that invoke governance capabilities such as
auditing, privacy protection, secret detection, or other runtime
controls.

\item \textbf{Plugins and Provider Bindings} --- references to
governance providers that extend runtime enforcement without modifying
the Policy Kernel.

\item \textbf{Approval Requirements} --- policies using the
\texttt{Approve} effect to require human authorization for high-impact
or irreversible operations.

\end{itemize}

Regulatory references, ownership information, documentation, and
change-management information may accompany a policy pack as
distribution or governance documentation. Such information is not part
of the core DGPL evaluation semantics.

This distinction keeps DGPL policy evaluation implementation-independent
while allowing policy packs to provide the documentation necessary for
enterprise governance and audit processes.

\subsection{Policy Packs for Key Industries}

Industry Policy Packs can be defined for a wide range of enterprise
domains. Table~\ref{tab:packs} presents representative examples.

\begin{table*}[!t]
\centering
\small
\begin{tabular}{@{}p{0.17\textwidth}p{0.45\textwidth}p{0.28\textwidth}@{}}
\toprule
\textbf{Industry} &
\textbf{Primary Governance Focus} &
\textbf{Example Regulatory / Standards Context} \\
\midrule

Healthcare &
Patient-record access, treatment relationship,
sensitive-data disclosure &
HIPAA Privacy and Security Rules \cite{hipaa} \\

Finance / Banking &
Transaction authorization, fraud response,
high-risk transaction approval &
PCI DSS and applicable financial-sector requirements \cite{kpmg} \\

Manufacturing &
Production-system access, machinery control,
operational safety &
ISO~12100 and applicable safety requirements \\

Government &
Citizen-data access, sensitive records,
controlled administrative operations &
Applicable public-sector and data-protection requirements \\

Human Resources &
Employee records, payroll operations,
sensitive workforce information &
Applicable employment and data-protection requirements \\

Oil and Gas &
Pipeline operations, equipment control,
emergency response and safety-critical actions &
Applicable process-safety and HSE requirements \\

GRC &
Audit evidence, policy exceptions,
compliance and reporting controls &
SOX, ISO~27001, and applicable governance frameworks \\

\bottomrule
\end{tabular}
\caption{Representative Industry Policy Packs and governance focus.}
\label{tab:packs}
\end{table*}

The regulatory and standards references in Table~\ref{tab:packs} describe
the governance context in which a pack may be used. They do not imply
that importing a policy pack alone constitutes certification or
regulatory compliance.

Across industries, the same underlying DGPL mechanisms remain
applicable:

\[
\begin{aligned}
  &\texttt{Allow}, \quad \texttt{Deny}, \quad \texttt{Approve}, \\
  &\text{Conditions}, \\
  &\text{Governance Obligations}, \\
  &\text{Evaluation Providers}.
\end{aligned}
\]

The difference between policy packs is therefore primarily the
domain-specific combination of principals, actions, resources,
conditions, effects, and governance obligations rather than a different
policy evaluation mechanism.

\subsection{Industry Policy Pack Examples}

The following examples demonstrate how Industry Policy Packs express
domain-specific governance using the same DGPL constructs.

Each request follows the general UPA execution model:

\[
E
\xrightarrow{N}
P
\xrightarrow{Eval}
(D,\Gamma)
\xrightarrow{Gov}
R.
\]
\newpage
\subsubsection{Healthcare --- Patient Record Access}

A Healthcare Policy Pack may restrict patient-record access to
authorized members of a patient's care team \cite{hipaa}. Additional contextual
conditions may require verified authentication and an established
treatment relationship.

\begin{lstlisting}[language=json]
{
  "version": "2026-06-05",
  "statements": [
    {
      "sid": "AllowPatientRecordAccess",
      "effect": "Allow",
      "principal": { "Role": ["Doctor", "Nurse"] },
      "action": ["patient-record:read"],
      "resource": ["healthcare-record:assigned-patient"],
      "condition": {
        "all": [
          { "fact": "context.authentication_verified",
            "operator": "equal", "value": true },
          { "fact": "context.treatment_relationship",
            "operator": "equal", "value": true }
        ]
      },
      "plugins": [ { "provider": "audit_logger", 
                     "mode": "record" } ]
    }
  ]
}
\end{lstlisting}

Consider a nurse attempting to access the record of an assigned
patient.

The normalized request is represented using PARC:

\begin{tcolorbox}[
  colback=gray!5,
  colframe=gray!30,
  arc=2pt,
  boxrule=0.5pt,
  top=4pt, bottom=4pt, left=8pt, right=8pt
]
\[
N(e) = \left\langle
  \begin{aligned}
    &\text{Nurse}, \\
    &\texttt{patient-record:read}, \\
    &\text{assigned-patient}, \\
    &ctx
  \end{aligned}
\right\rangle
\]

where

\[
\begin{aligned}
  ctx = \{ &\text{authentication\_verified} = \text{true}, \\
           &\text{treatment\_relationship} = \text{true} \}
\end{aligned}
\]

\end{tcolorbox}

The statement matches the canonical request and produces:

\[
Eval(p,\Pi)
=
(\texttt{Allow},\Gamma)
\]

where

\[
\Gamma =
\{
\text{AuditLog}
\}.
\]

Runtime Governance executes the resulting obligation and the request
continues to the protected resource.

If either contextual condition is false, the statement does not match.
If no other policy authorizes the request, the default-deny rule applies.

\subsubsection{Cross-Industry --- LLM Input Protection}

Some governance controls are not specific to an industry but protect
the common AI runtime layer. Such controls can be distributed as
cross-industry policy components and composed with Industry Policy
Packs.

For example, an organization may require sensitive information to be
detected and redacted before an LLM receives input.

\begin{lstlisting}[language=json]
{
  "version": "2026-01-01",
  "statements": [
    {
      "sid": "EnterpriseProtection",
      "effect": "Allow",
      "action": ["llm:input"],
      "resource": ["*"],
      "plugins": [
        {
          "provider": "presidio",
          "mode": "redact"
        },
        {
          "provider": "secret_scanner",
          "mode": "redact"
        }
      ]
    }
  ]
}
\end{lstlisting}

The policy itself does not contain implementation logic for either
provider. Instead, the matched statement produces governance
obligations that are executed by the Runtime Governance subsystem.

This allows the same protection policy to be combined with Healthcare,
Finance, Oil and Gas, or other Industry Policy Packs.

\subsubsection{Banking --- Fraud Detection Agent}

A Banking Policy Pack may allow an autonomous fraud-detection agent to
perform investigative operations while requiring human approval for
irreversible actions.

\begin{lstlisting}[language=json]
{
  "version": "2026-06-06",
  "statements": [
    {
      "sid": "FraudDetectionAgentInvestigate",
      "effect": "Allow",
      "principal": {
        "Type": ["AIAgent"],
        "Role": ["FraudDetectionAgent"]
      },
      "action": [
        "transaction:analyse",
        "fraud-risk:calculate",
        "transaction:temporary-hold",
        "fraud-team:notify",
        "incident:create"
      ],
      "resource": [
        "bank-transaction:*"
      ],
      "condition": {
        "all": [
          {
            "fact": "context.fraudRisk",
            "operator": "equal",
            "value": "High"
          },
          {
            "fact": "context.transactionDataValidated",
            "operator": "equal",
            "value": true
          }
        ]
      },
      "plugins": [
        {
          "provider": "audit_logger",
          "mode": "record"
        }
      ]
    },
    {
      "sid": "FraudDetectionAgentIrreversibleAction",
      "effect": "Approve",
      "principal": {
        "Type": ["AIAgent"],
        "Role": ["FraudDetectionAgent"]
      },
      "action": [
        "transaction:reject",
        "account:block"
      ],
      "resource": [
        "bank-transaction:*",
        "bank-account:*"
      ],
      "approval": {
        "role": "FraudOfficer"
      }
    }
  ]
}
\end{lstlisting}

The first statement allows the agent to investigate a high-risk
transaction and perform reversible investigative actions.

A request to permanently reject the transaction or block the account
matches the second statement and produces:

\[
Eval(p,\Pi)
=
(\texttt{Approve},\Gamma).
\]

Runtime Governance suspends autonomous execution and initiates the
approval workflow for the designated \texttt{FraudOfficer}.

This illustrates how Industry Policy Packs can preserve autonomous
execution for lower-risk actions while introducing human control at
specific decision boundaries.

\subsubsection{Oil and Gas --- LNG Safety Agent}

An Oil and Gas Policy Pack can govern safety-critical autonomous
operations.

For example, an LNG Safety Agent may be permitted to stop an LNG
transfer when a validated leak is detected, while requiring human
approval before restarting the transfer.

\begin{lstlisting}[language=json]
{
  "version": "2026-06-06",
  "statements": [
    {
      "sid": "LNGSafetyEmergencyResponse",
      "effect": "Allow",
      "principal": {
        "Type": ["AIAgent"],
        "Role": ["LNGSafetyAgent"]
      },
      "action": [
        "alarm:create",
        "control-room:notify",
        "lng-transfer:stop",
        "incident:create"
      ],
      "resource": [
        "lng-terminal:*"
      ],
      "condition": {
        "all": [
          {
            "fact": "context.gasLeakDetected",
            "operator": "equal",
            "value": true
          },
          {
            "fact": "context.sensorValidated",
            "operator": "equal",
            "value": true
          }
        ]
      }
    },
    {
      "sid": "LNGSafetyRestartTransfer",
      "effect": "Approve",
      "principal": {
        "Type": ["AIAgent"],
        "Role": ["LNGSafetyAgent"]
      },
      "action": [
        "lng-transfer:restart"
      ],
      "resource": [
        "lng-terminal:*"
      ],
      "approval": {
        "role": "ControlRoomSupervisor"
      }
    }
  ]
}
\end{lstlisting}

For a validated emergency condition, the first statement produces:

\[
Eval(p,\Pi)
=
(\texttt{Allow},\Gamma)
\]

and the safety action can be executed immediately.

A later restart request produces:

\[
Eval(p,\Pi)
=
(\texttt{Approve},\Gamma)
\]

and Runtime Governance transfers control to the designated
\texttt{ControlRoomSupervisor}.

This demonstrates that Industry Policy Packs can express different
governance boundaries for autonomous actions within the same workflow.

\subsection{Human Approval Within Industry Policy Packs}

Industry Policy Packs can use the \texttt{Approve} effect when an action
is considered sufficiently sensitive, irreversible, or high impact that
autonomous execution should be suspended pending human authorization.

Examples include:

\begin{itemize}

\item permanently blocking a bank account;

\item restarting a safety-critical industrial process;

\item releasing sensitive information outside an organization;

\item approving a high-value financial transaction; and

\item executing an irreversible administrative operation.

\end{itemize}

The approval mechanism separates the ability of an agent to initiate
an operation from the organization's authority to authorize its
execution.

Formally:

\[
\begin{aligned}
  \texttt{Approve} &\rightarrow \text{Approval Workflow} \\
  &\rightarrow \begin{cases}
    \texttt{Approved} &\rightarrow \text{Continue Execution} \\
    \texttt{Rejected} &\rightarrow \text{Terminate} \\
    \texttt{Timeout}  &\rightarrow \text{Terminate or Hold}
  \end{cases}
\end{aligned}
\]

The exact timeout and rejection behavior is determined by the
organization's governance configuration.

This allows autonomous agents to perform low-risk and reversible
operations while retaining human control over high-impact decisions.

\subsection{Composition of Industry Policy Packs}

Industry Policy Packs can be composed with organization-specific
policies and other packs.

Let

\[
\Pi_{\mathrm{org}}
\]

represent organization-specific policies and let

\[
\mathit{Pack}_1,\ldots,\mathit{Pack}_n
\]

represent active Industry Policy Packs. The effective policy repository
is:

\[
\Pi_{\mathrm{effective}}
=
\Pi_{\mathrm{org}}
\cup
\bigcup_{i=1}^{n}
\mathit{Pack}_i.\mathit{Statements}.
\]

A single canonical policy request is evaluated against this effective
repository.

The Policy Kernel therefore does not evaluate a separate request for
each policy pack. Instead, the same canonical request is matched
against all applicable statements.

The resulting set of matching statements is subsequently processed by
Evaluation Providers and Conflict Resolution.

This preserves the policy composition semantics defined by DGPL.

For example, a request to process a healthcare payment may simultaneously
be subject to:

\begin{itemize}

\item Healthcare access controls;

\item Finance transaction controls;

\item Organization-specific approval policies; and

\item Cross-industry LLM protection policies.

\end{itemize}

All applicable statements participate in the same governance
evaluation.

\subsection{Threat-Aware Governance Within Industry Policy Packs}

Industry Policy Packs can also reference shared Evaluation Providers
for threat detection.

The policy pack does not need to implement threat-detection algorithms
itself. Instead, the Policy Evaluation subsystem can invoke the
appropriate Evaluation Provider and use the resulting evaluation
evidence as part of governance processing.

Representative threat categories include:

\begin{itemize}

\item Prompt injection and jailbreak attempts;

\item Unauthorized agent actions;

\item Attempts to bypass approval requirements;

\item Data exfiltration and privilege escalation;

\item Malicious tool invocation; and

\item Unsafe or socially engineered instructions.

\end{itemize}

This design separates threat-detection implementation from
domain-specific governance policy.

For example, the same prompt-injection detection capability can be
shared by Healthcare, Banking, and Oil and Gas policy packs while each
domain independently determines the resulting governance action.

Thus:

\[
\text{Threat Detection}
\neq
\text{Industry Policy}.
\]

Instead, threat detection provides evaluation evidence that can
participate in the common governance evaluation process.

\subsection{Benefits of Industry Policy Packs}

Industry Policy Packs provide several architectural benefits.

\begin{itemize}

\item \textbf{Policy Reuse} --- common industry governance controls can
be reused across multiple applications and AI agents.

\item \textbf{Consistent Enforcement} --- domain-specific policies are
evaluated through the same Policy Kernel and Runtime Governance
mechanisms.

\item \textbf{Composability} --- multiple policy packs can be combined
with organization-specific policies.

\item \textbf{Versionability} --- policy packs can evolve independently
and provide reproducible governance configurations.

\item \textbf{Extensibility} --- new industry packs can be introduced
without modifying the Policy Kernel.

\item \textbf{Operational Separation} --- domain governance rules remain
declarative while provider implementations remain external to the
policy language.

\end{itemize}

The key architectural property is that Industry Policy Packs extend
the \textit{policy repository}, not the \textit{policy evaluation
architecture}.


\subsection{Summary}

Industry Policy Packs provide a reusable mechanism for expressing
domain-specific enterprise AI governance using the Declarative
Governance Policy Language.

A policy pack is composed of ordinary DGPL statements and is incorporated
into the organization's policy repository:

\[
\begin{aligned}
  \mathit{Pack} &\rightarrow \Pi \\
  &\rightarrow \text{Policy Matching} \\
  &\rightarrow \text{Evaluation} \\
  &\rightarrow \text{Conflict Resolution} \\
  &\rightarrow \text{Runtime Governance}.
\end{aligned}
\]

Because the same Policy Kernel evaluates all applicable statements,
Industry Policy Packs do not require specialized enforcement logic.

Healthcare, Banking, Oil and Gas, and other domains can therefore
define different governance requirements while sharing the same
underlying PARC model, policy evaluation semantics, governance
obligations, approval mechanisms, and provider framework.

This establishes Industry Policy Packs as a distribution and reuse
mechanism for DGPL rather than a separate governance subsystem.

\section{SID Registry, Indexing, and Resolution}
\subsection{DGPL Policy Statements}

Each DGPL policy statement is identified by a unique Statement
Identifier (\textit{SID}). The SID provides a stable identifier for
policy statements and enables policy evaluation results, governance
obligations, audit records, and administrative operations to reference
the originating policy statement.

For example, a DGPL statement may define:

\begin{lstlisting}[language=json]
{
  "sid": "ProductionDelete",
  "effect": "Approve",
  "action": [
    "db:delete"
  ],
  "resource": [
    "database:production"
  ],
  "approval": {
    "role": "Admin"
  }
}
\end{lstlisting}

In this example, \texttt{ProductionDelete} identifies the policy
statement but does not itself determine whether the statement matches a
request. Matching is performed using the applicable policy attributes,
while the SID provides the identity of the matched statement.

\subsubsection{SID Registry}

The SID Registry maintains the relationship between a statement
identifier and its corresponding policy statement. Conceptually, the
registry can be represented as:

\[
R_{\mathrm{SID}} :
SID \rightarrow Statement
\]

where $SID$ is the statement identifier and $Statement$ represents the
corresponding DGPL policy statement.

The registry enables the Policy Kernel and supporting governance
components to resolve the origin of an evaluated policy decision.

For example:

\[
R_{\mathrm{SID}}(\texttt{ProductionDelete})
=
p_{\texttt{ProductionDelete}}
\]

The registry therefore provides a stable reference for policy
statements without requiring the evaluation subsystem to depend on the
physical storage location of the policy.

\subsubsection*{Policy Indexing}

Evaluating every policy statement against every runtime request can
become inefficient as the policy repository grows. UPA therefore
supports indexing mechanisms that reduce the number of candidate
statements considered during policy matching.

Given a normalized PARC request:

\[
q =
\langle
Principal,\ Action,\ Resource,\ Context
\rangle
\]

the indexing layer identifies a candidate policy set:

\[
I(q) = P_q
\]

where $P_q \subseteq P$ and $P$ represents the complete set of policy
statements.

The index may use policy attributes such as:

\begin{itemize}
    \item Principal
    \item Action
    \item Resource
    \item Policy effect
    \item Condition presence
\end{itemize}

The purpose of indexing is to identify potentially applicable
statements efficiently. Final policy applicability remains the
responsibility of the Policy Matching and Evaluation subsystems.

Thus:

\[
P_q
\subseteq
P
\]

represents the candidate set rather than the final set of matching
policies.

\subsubsection*{SID Resolution}

After candidate statements have been identified and evaluated, the
Policy Kernel can associate the resulting decision with the SIDs of the
applicable statements.

Let:

\[
M(q) =
\{p_1,p_2,\ldots,p_n\}
\]

represent the set of statements that match the normalized request.

The corresponding SID set is:

\[
SID(M(q)) =
\{
sid(p_1),sid(p_2),\ldots,sid(p_n)
\}
\]

This provides a direct relationship between the runtime decision and
the policy statements that contributed to that decision.

For example:

\[
\operatorname{Eval}(q,P)
=
(\texttt{Approve},\Gamma,\{ \texttt{ProductionDelete} \})
\]

The SID therefore provides traceability without becoming a separate
policy request or an independent evaluation object.

\subsubsection*{Resolution Flow}

The complete resolution process can be represented as:

\[
\begin{aligned}
Runtime\ Event
&\rightarrow
Semantic\ Normalization \\
&\rightarrow
PARC\ Request \\
&\rightarrow
Policy\ Index \\
&\rightarrow
Candidate\ Statements \\
&\rightarrow
Policy\ Matching \\
&\rightarrow
Matched\ Statements \\
&\rightarrow
SID\ Resolution \\
&\rightarrow
Policy\ Evaluation
\end{aligned}
\]

The resulting governance decision may subsequently reference the
resolved SIDs for auditability, explanation, approval workflows, and
governance evidence.

\begin{tcolorbox}[
title=Example: SID Resolution,
colback=gray!5,
colframe=black!40,
fonttitle=\bfseries
]

\textbf{Runtime Request}

\[
q = \left\langle
  \begin{aligned}
    &\text{User}, \\
    &\texttt{db:delete}, \\
    &\texttt{database:production}, \\
    &\text{EnterpriseContext}
  \end{aligned}
\right\rangle
\]

\textbf{Candidate Policy}

\begin{lstlisting}[language=json]
{
  "sid": "ProductionDelete",
  "effect": "Approve",
  "action": [
    "db:delete"
  ],
  "resource": [
    "database:production"
  ],
  "approval": {
    "role": "Admin"
  }
}
\end{lstlisting}

\textbf{Matching}

The action and resource match the normalized request. The statement
therefore becomes applicable.

\[
M(q) =
\{
p_{\texttt{ProductionDelete}}
\}
\]

\textbf{SID Resolution}

\[
SID(M(q))
=
\{
\texttt{ProductionDelete}
\}
\]

\textbf{Governance Decision}

\[
\operatorname{Eval}(q,P) = \left(
  \begin{aligned}
    &\texttt{Approve}, \\
    &\Gamma, \\
    &\{\texttt{ProductionDelete}\}
  \end{aligned}
\right)
\]

The SID provides the identity of the policy statement responsible for
the resulting governance decision.

\end{tcolorbox}

The SID Registry and indexing mechanism therefore serve different but
complementary purposes. Indexing improves the efficiency of policy
candidate retrieval, while the SID Registry provides stable statement
identity and resolution. Separating these responsibilities allows the
Policy Kernel to optimize policy retrieval without changing the
semantic identity of DGPL policy statements.

\section{Experimental Evaluation}

This section evaluates the implemented Unified Policy Architecture (UPA) using six complementary dimensions: correctness, latency, policy repository scalability, concurrent request scalability, plugin execution overhead, and interoperability through Semantic Normalization. The order follows the evaluation flow from validating decision behaviour, through performance and scalability, to the framework-independent normalization layer.

\begin{table}[tb]
\centering
\caption{Evaluation Dimensions}
\label{tab:evaluation-dimensions}
\renewcommand{\arraystretch}{1.15}
\scriptsize
\begin{tabularx}{\linewidth}{@{} c l X @{}}
\toprule
\textbf{No.} & \textbf{Evaluation Dimension} & \textbf{Purpose} \\
\midrule
1 & Correctness & Checks whether the Policy Engine returns the expected \texttt{Allow}, \texttt{Deny}, or \texttt{Approve} decision. \\
2 & Latency & Measures the time required to evaluate one authorization request and return a decision. \\
3 & Policy Repository Scalability & Checks performance as the number of policy statements increases. \\
4 & Concurrent Request Scalability & Checks whether multiple authorization requests can be handled without failures. \\
5 & Plugin Execution Overhead & Measures additional latency introduced by governance plugins. \\
6 & Interoperability / Normalization & Checks whether different runtime inputs can be converted into a common authorization request. \\
\bottomrule
\end{tabularx}
\end{table}

\subsection{Correctness}

Correctness evaluates whether the Policy Engine produces the expected governance decision for a canonical authorization request. Non-matching policies are ignored, while the effects of matching policies are resolved using deterministic precedence:
\[
\texttt{Deny} > \texttt{Approve} > \texttt{Allow} > \text{\texttt{Default Deny}}.
\]

\subsubsection*{Representative Policy}

\begin{lstlisting}[language=json, basicstyle=\ttfamily\scriptsize]
{
  "sid": "AllowSalesDatabaseRead",
  "effect": "Allow",
  "action": ["db:read"],
  "resource": ["database:sales"]
}
\end{lstlisting}

\subsubsection*{Authorization Request}

\begin{lstlisting}[language=json, basicstyle=\ttfamily\scriptsize]
{
  "principal": {"id": "user-001", "type": "User"},
  "action": {"name": "db:read"},
  "resource": {"id": "sales", "type": "database"}
}
\end{lstlisting}

\subsubsection*{Evaluation Result}

\begin{lstlisting}[language=json, basicstyle=\ttfamily\scriptsize]
{
  "decision": "Allow",
  "matched_policy": "AllowSalesDatabaseRead"
}
\end{lstlisting}

The request matches the action and resource defined by \texttt{AllowSalesDatabaseRead}; therefore, the engine returns \texttt{Allow}. Additional cases verify mismatches, default deny behaviour, human approval, and conflict resolution.

\begin{table}[tb]
\centering
\caption{Policy decision verification suite}
\label{tab:correctness-suite}
\renewcommand{\arraystretch}{1.15}
\scriptsize
\setlength{\tabcolsep}{3pt}
\begin{tabularx}{\linewidth}{@{} l X c c c @{}}
\toprule
\textbf{Test ID} & \textbf{Scenario} & \textbf{Exp.} & \textbf{Act.} & \textbf{Res.} \\
\midrule
\texttt{COR-01} & Matching Allow policy & \texttt{Allow} & \texttt{Allow} & Pass \\
\texttt{COR-02} & Action mismatch & \texttt{Deny} & \texttt{Deny} & Pass \\
\texttt{COR-03} & Resource mismatch & \texttt{Deny} & \texttt{Deny} & Pass \\
\texttt{COR-04} & Principal mismatch & \texttt{Deny} & \texttt{Deny} & Pass \\
\texttt{COR-05} & Deny overrides Allow & \texttt{Deny} & \texttt{Deny} & Pass \\
\texttt{COR-06} & Approve overrides Allow & \texttt{Approve} & \texttt{Approve} & Pass \\
\texttt{COR-07} & Deny overrides Approve/Allow & \texttt{Deny} & \texttt{Deny} & Pass \\
\texttt{COR-08} & No matching policy & \texttt{Deny} & \texttt{Deny} & Pass \\
\texttt{COR-09} & Human approval policy & \texttt{Approve} & \texttt{Approve} & Pass \\
\bottomrule
\end{tabularx}
\end{table}

All nine structured correctness cases returned the expected decision, producing 9/9 successful cases in the preliminary engine-level verification suite.

\subsection{Latency}

Latency measures the time taken by \texttt{PolicyEngine.authorize()} to evaluate one canonical authorization request and return a governance decision. Ten warm-up executions were excluded, followed by 100 measured requests for each policy repository size.

\subsubsection*{Representative Benchmark Request}

\begin{tcolorbox}[
  colback=gray!5,
  colframe=gray!30,
  arc=2pt,
  boxrule=0.5pt,
  top=3pt, bottom=3pt, left=4pt, right=4pt
]
\scriptsize\ttfamily
Principal = research-agent\\
Action    = tool:execute\\
Resource  = tool:web\_search\\
Matching policy placed at final position.
\end{tcolorbox}

\begin{table}[tb]
\centering
\caption{Authorization latency (ms)}
\label{tab:authorization-latency}
\renewcommand{\arraystretch}{1.15}
\scriptsize
\setlength{\tabcolsep}{2.5pt}
\begin{tabularx}{\linewidth}{@{} r r r r r r @{}}
\toprule
\textbf{Policies} & \textbf{Mean} & \textbf{Median} & \textbf{P95} & \textbf{P99} & \textbf{Max} \\
\midrule
10    & 0.0689  & 0.0612  & 0.0837   & 0.1103   & 0.6170 \\
100   & 0.4650  & 0.4003  & 0.8408   & 1.2113   & 1.5430 \\
500   & 4.0484  & 2.3851  & 3.4456   & 74.9523  & 94.3876 \\
1,000 & 4.7721  & 3.1485  & 4.4085   & 54.3665  & 59.1441 \\
5,000 & 32.1788 & 17.2241 & 108.2480 & 141.9245 & 145.4172 \\
\bottomrule
\end{tabularx}
\end{table}

\begin{figure*}[!t]
\centering
\includegraphics[width=0.85\linewidth]{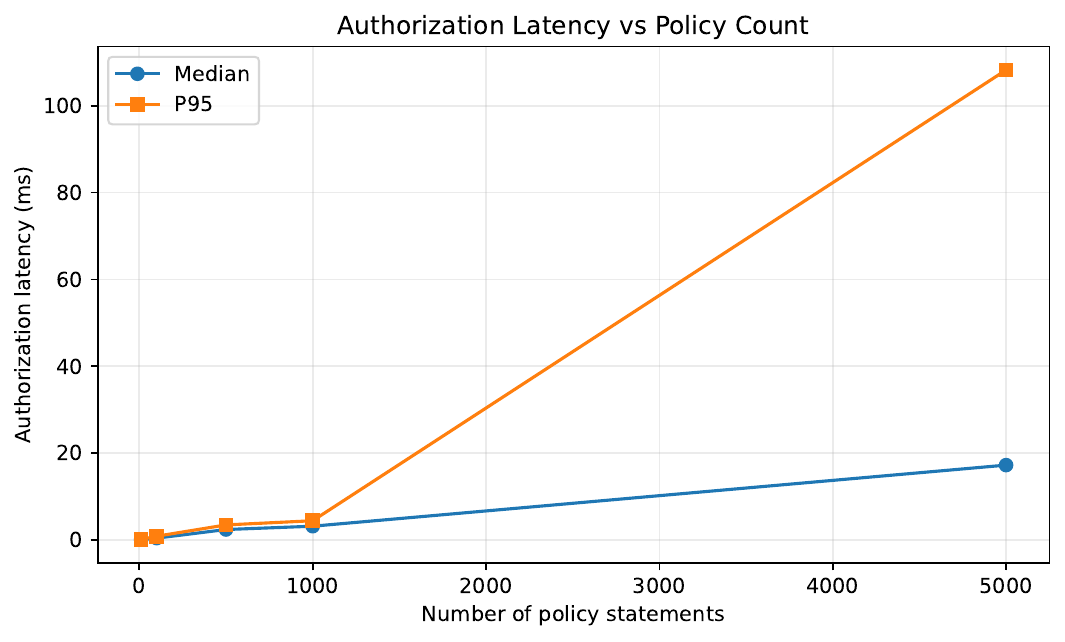}
\caption{Median and P95 authorization latency across increasing policy counts.}
\label{fig:authorization-latency}
\end{figure*}

Median latency remained below approximately 3.2 ms through 1,000 policies and increased to 17.2241 ms at 5,000 policies. P95 reached 108.248 ms at 5,000 policies, indicating greater tail-latency variation at the largest tested repository.

\subsection{Policy Repository Scalability}

Policy Repository Scalability examines how the evaluator behaves as the number of policy statements increases. The matching policy was placed at the end of every repository so that the sequential evaluator traversed the complete policy set before finding the applicable rule.

\subsubsection*{Repository Growth Configuration}

\begin{tcolorbox}[
  colback=gray!5,
  colframe=gray!30,
  arc=2pt,
  boxrule=0.5pt,
  top=4pt, bottom=4pt, left=6pt, right=6pt
]
\scriptsize\ttfamily
10 policies    $\rightarrow$ match at position 10\\
100 policies   $\rightarrow$ match at position 100\\
500 policies   $\rightarrow$ match at position 500\\
1,000 policies $\rightarrow$ match at position 1,000\\
5,000 policies $\rightarrow$ match at position 5,000
\end{tcolorbox}

\begin{table}[tb]
\centering
\caption{Effective sequential throughput}
\label{tab:sequential-throughput}
\renewcommand{\arraystretch}{1.15}
\scriptsize
\setlength{\tabcolsep}{3pt}
\begin{tabularx}{\linewidth}{@{} r r r X @{}}
\toprule
\textbf{Policies} & \textbf{req/s} & \textbf{Median (ms)} & \textbf{Observation} \\
\midrule
10    & 14,519.60 & 0.0612  & Very small repository \\
100   & 2,150.54  & 0.4003  & Low evaluation cost \\
500   & 247.01    & 2.3851  & Higher scan cost \\
1,000 & 209.55    & 3.1485  & Low median latency \\
5,000 & 31.08     & 17.2241 & Scan becomes costly \\
\bottomrule
\end{tabularx}
\end{table}

As repository size increased, effective sequential throughput decreased. The result identifies sequential policy traversal as the main scalability limitation in the current evaluator and motivates future indexing or partitioning strategies.

\subsection{Concurrent Request Scalability}

Concurrent Request Scalability checks whether the Policy Engine can process multiple authorization requests in the same execution period without failures. The repository was fixed at 1,000 policy statements, while concurrency was increased from 1 to 100 using in-process \texttt{asyncio} execution.

\begin{table}[tb]
\centering
\caption{Concurrent authorization results}
\label{tab:concurrency-results}
\renewcommand{\arraystretch}{1.15}
\scriptsize
\setlength{\tabcolsep}{2pt}
\begin{tabularx}{\linewidth}{@{} r r r r r r r @{}}
\toprule
\textbf{Conc.} & \textbf{Reqs} & \textbf{Succ.} & \textbf{Fail} & \textbf{Req/s} & \textbf{Mean(ms)} & \textbf{P95(ms)} \\
\midrule
1   & 10    & 10    & 0 & 101.60 & 9.3856 & 35.1700 \\
10  & 100   & 100   & 0 & 178.03 & 5.2143 & 5.9975 \\
25  & 250   & 250   & 0 & 209.71 & 4.4762 & 5.4775 \\
50  & 500   & 500   & 0 & 237.96 & 3.9737 & 5.1276 \\
100 & 1,000 & 1,000 & 0 & 236.27 & 4.0023 & 4.3574 \\
\bottomrule
\end{tabularx}
\end{table}

\begin{figure*}[!t]
\centering
\includegraphics[width=0.85\linewidth]{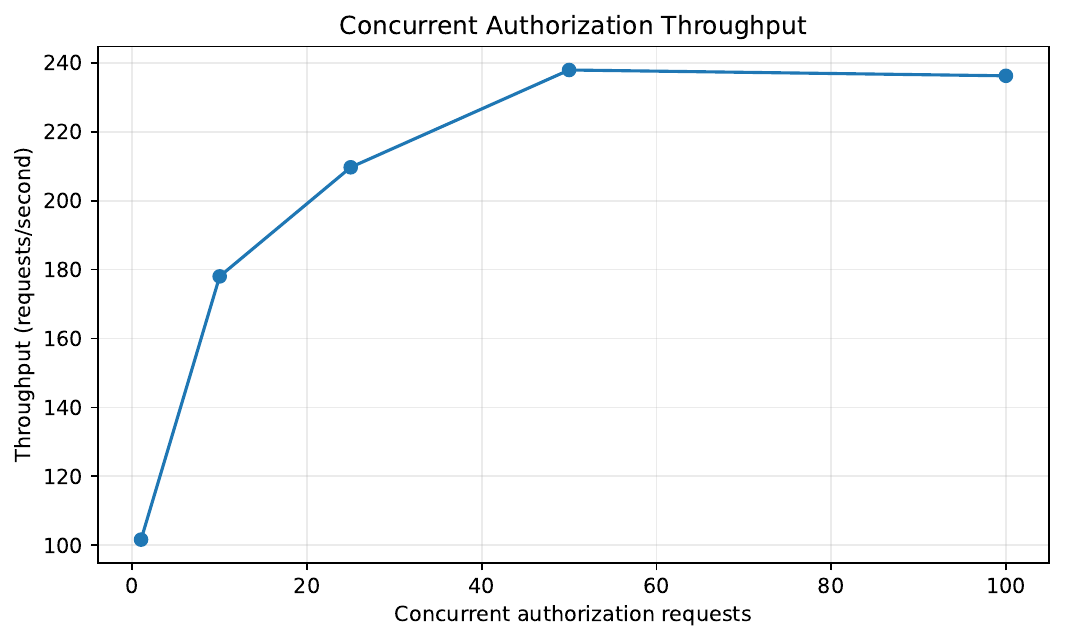}
\caption{Observed throughput across increasing concurrency levels.}
\label{fig:concurrency-throughput}
\end{figure*}

Across all concurrency configurations, 1,860 measured authorization requests completed successfully and no failures were recorded. Throughput increased to 237.96 requests per second at concurrency 50 and remained near that level at concurrency 100.

\subsection{Plugin Execution Overhead}

Plugin Execution Overhead measures the additional latency introduced when governance providers are executed as part of policy evaluation. Four configurations were measured: no plugin, Presidio only, secret scanner only, and Presidio followed by the secret scanner.

\subsubsection*{Representative Plugin Policy}

\begin{lstlisting}[language=json, basicstyle=\ttfamily\scriptsize]
{
  "sid": "EnterpriseProtection",
  "effect": "Allow",
  "action": ["llm:input"],
  "resource": ["*"],
  "plugins": [
    {"provider": "presidio", "mode": "redact"},
    {"provider": "secret_scanner", "mode": "redact"}
  ]
}
\end{lstlisting}

\begin{table}[tb]
\centering
\caption{Plugin execution latency (ms)}
\label{tab:plugin-latency}
\renewcommand{\arraystretch}{1.15}
\scriptsize
\setlength{\tabcolsep}{2pt}
\begin{tabularx}{\linewidth}{@{} X r r r r @{}}
\toprule
\textbf{Configuration} & \textbf{Mean} & \textbf{Median} & \textbf{P95} & \textbf{Overhead} \\
\midrule
No plugins              & 0.0379    & 0.0350    & 0.0604    & 0.0000 \\
Presidio only          & 2398.5800 & 2195.5687 & 4128.9392 & 2398.5421 \\
Secret scanner only    & 0.0471    & 0.0460    & 0.0540    & 0.0092 \\
Presidio + Secret scan & 1916.9712 & 1609.6492 & 3294.6966 & 1916.9333 \\
\bottomrule
\end{tabularx}
\end{table}

\begin{figure*}[!t]
\centering
\includegraphics[width=0.85\linewidth]{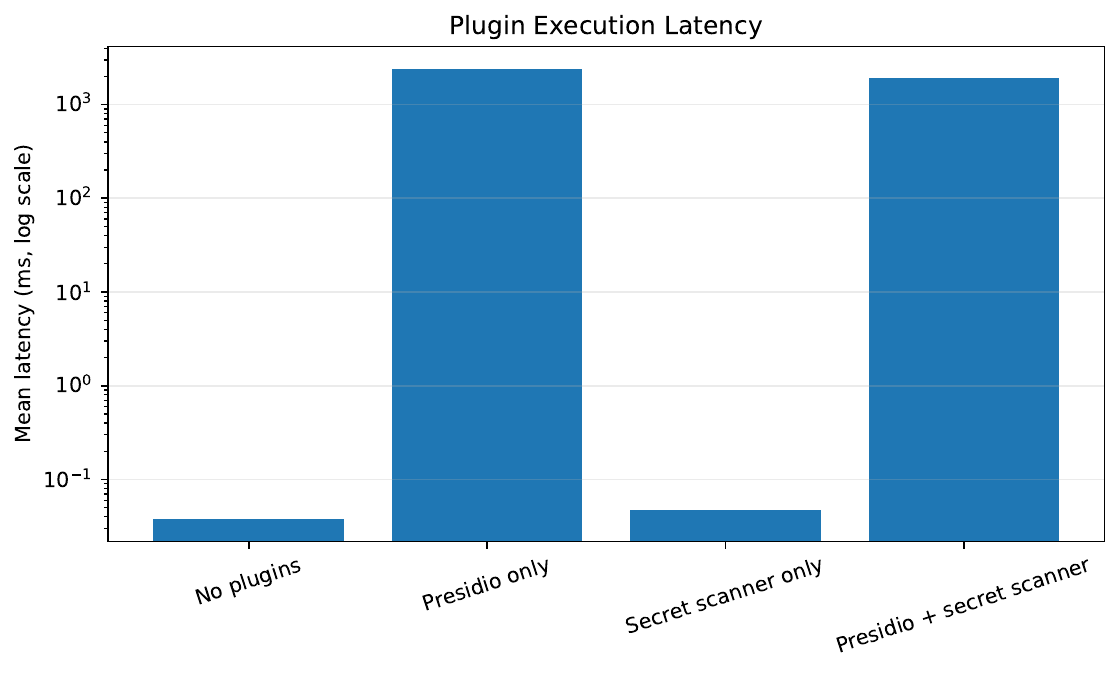}
\caption{Mean plugin latency on a logarithmic scale.}
\label{fig:plugin-latency}
\end{figure*}

The secret scanner introduced only 0.0092 ms of mean overhead relative to the no-plugin baseline. Presidio added approximately 2,398.54 ms of mean overhead and dominated the measured runtime. The combined result is treated as preliminary because warm-up, caching, dependency initialization, and test order may influence plugin measurements.

\subsection{Interoperability and Semantic Normalization}

Interoperability and Semantic Normalization evaluate whether heterogeneous runtime requests can be transformed into a canonical authorization representation containing \texttt{principal}, \texttt{action}, \texttt{resource}, \texttt{context}, and \texttt{metadata}. In the current Context Builder pipeline, the LLM receives solely the natural-language user query. Trusted identity attributes and execution metadata are bound independently by the application runtime to prevent prompt-injection spoofing.

\subsubsection*{Representative Runtime Input}

\begin{tcolorbox}[
  colback=gray!5,
  colframe=gray!40,
  arc=2pt,
  boxrule=0.5pt,
  top=4pt, bottom=4pt, left=6pt, right=6pt
]
\scriptsize\ttfamily
Authenticated principal:\\
\hspace*{1.5em}user-001 / User\\[0.3em]
Natural-language request:\\
\hspace*{1.5em}"Read customer data in the sales database"
\end{tcolorbox}

\subsubsection*{Generated Canonical Authorization Request}

\begin{lstlisting}[language=json, basicstyle=\ttfamily\scriptsize]
{
  "principal": {"id": "user-001", "type": "User"},
  "action": {"name": "db:read"},
  "resource": {"id": "sales", "type": "database"},
  "context": {
    "custom": {
      "intent": "read_data",
      "domain": "database",
      "confidence": 0.9,
      "query_type": "task",
      "data_type": "customer"
    }
  }
}
\end{lstlisting}

\begin{figure*}[!t]
\centering
\includegraphics[width=0.65\linewidth]{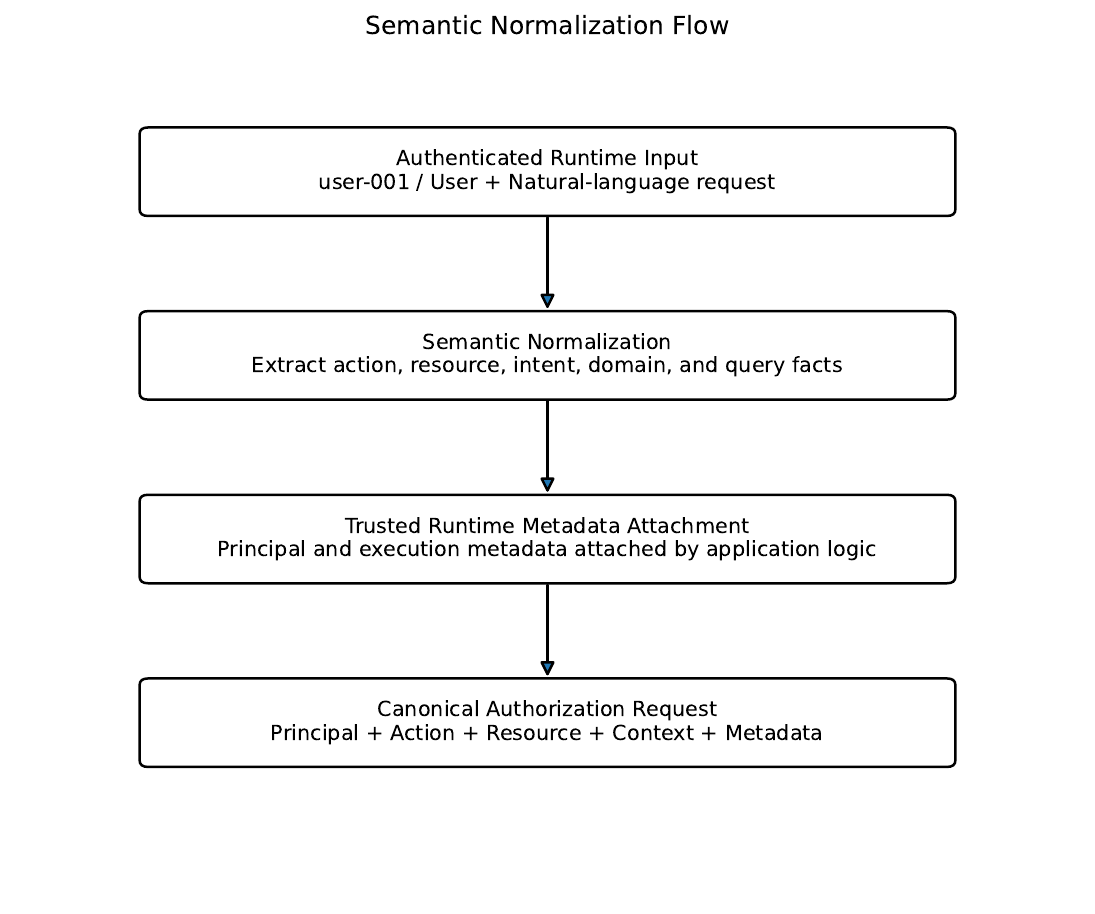}
\caption{Semantic Normalization flow from authenticated runtime input to a canonical authorization request.}
\label{fig:semantic-normalization-flow}
\end{figure*}

\begin{table}[tb]
\centering
\caption{Semantic normalization verification -- \texttt{SN-01}}
\label{tab:semantic-normalization}
\renewcommand{\arraystretch}{1.15}
\scriptsize
\setlength{\tabcolsep}{3pt}
\begin{tabularx}{\linewidth}{@{} l X X c @{}}
\toprule
\textbf{Field} & \textbf{Expected} & \textbf{Generated} & \textbf{Result} \\
\midrule
Principal    & \texttt{user-001 / User}       & \texttt{user-001 / User}       & Pass \\
Action       & \texttt{db:read}               & \texttt{db:read}               & Pass \\
Res. Type    & \texttt{database}              & \texttt{database}              & Pass \\
Res. ID      & \texttt{sales}                 & \texttt{sales}                 & Pass \\
Intent       & \texttt{read\_data}            & \texttt{read\_data}            & Pass \\
Domain       & \texttt{database}              & \texttt{database}              & Pass \\
Query Type   & \texttt{task}                  & \texttt{task}                  & Pass \\
Fact Ext.    & \texttt{data\_type=customer}   & \texttt{data\_type=customer}   & Pass \\
\bottomrule
\end{tabularx}
\end{table}

\texttt{SN-01} shows that the natural-language request was converted into the expected canonical action and resource while preserving the authenticated principal. This representative case demonstrates the intended interoperability role of Semantic Normalization: different runtime sources can be mapped to one common authorization structure before policy evaluation. A broader labelled test suite is required to quantify normalization accuracy across multiple actions, resources, and runtime formats.

\subsection{Summary of Findings}

The preliminary experimental results establish a concrete performance baseline for the UPA prototype across all six evaluation dimensions:

\begin{itemize}[leftmargin=1.2em, itemsep=0.3em]
  \item \textbf{Correctness:} The Policy Engine achieved a 100\% pass rate ($9/9$ test cases), correctly resolving precedence conflicts (\texttt{Deny} $>$ \texttt{Approve} $>$ \texttt{Allow} $>$ \text{\texttt{Default Deny}}).
  \item \textbf{Latency \& Scalability:} Median decision latency remained sub-3.2\,ms for repositories up to 1,000 policies. However, sequential traversal caused latency to scale linearly, reaching 17.2\,ms (median) and 108.2\,ms (P95) at 5,000 policies.
  \item \textbf{Concurrency:} The engine processed 1,860 concurrent requests under multi-threaded execution with zero failures, saturating at a peak throughput of $\approx 238$ req/s.
  \item \textbf{Plugin Overhead:} Lightweight governance plugins (e.g., secret scanning) added negligible delay ($+0.0092$\,ms), whereas deep NLP providers (e.g., Presidio) dominated total execution time ($\approx 2,398$\,ms).
  \item \textbf{Semantic Normalization:} The framework successfully translated unaligned natural-language queries into canonical principal-action-resource tuples while preserving authenticated identity boundaries.
\end{itemize}
\section{Discussion}

The Unified Policy Architecture (UPA) provides a governance-oriented
architectural model for Enterprise AI systems in which policy evaluation,
runtime governance, human intervention, and governance extensions are
treated as coordinated components of a single control plane. The
architecture is designed to address the operational requirements of
autonomous agents and LLM-based applications rather than treating
governance solely as a model-level safety mechanism.

\subsection{Strengths}

A primary strength of UPA is the separation between policy definition,
policy evaluation, and policy execution. The Declarative Governance
Policy Language (DGPL) provides a structured representation of
governance requirements, while the Policy Kernel evaluates a normalized
request against applicable policies. This separation allows governance
rules to remain independent of individual agent implementations.

The use of the PARC model---Principal, Action, Resource, and Context---
provides a common representation for policy matching across heterogeneous
enterprise applications. Consequently, policies can be evaluated against
different types of agents, tools, resources, and runtime operations
without requiring application-specific policy logic.

Another strength is the explicit representation of governance outcomes.
UPA does not restrict policy evaluation to binary allow or deny decisions.
The architecture supports outcomes such as \texttt{Allow},
\texttt{Deny}, \texttt{Approve}, and associated governance obligations.
This enables policies to initiate actions such as human approval,
auditing, notification, inspection, transformation, or other runtime
governance operations.

The Plugin Framework further separates extensible governance capabilities
from the Policy Kernel. Privacy protection, secret detection, compliance
validation, threat detection, and organization-specific controls can
therefore be introduced without modifying the fundamental policy
evaluation mechanism.

\subsection{Architectural Implications}

The architecture implies that Enterprise AI governance should be treated
as a runtime control-plane capability rather than as functionality
embedded independently within each AI application. This distinction
becomes increasingly important as organizations operate multiple agents,
models, tools, knowledge sources, and business applications.

A centralized governance architecture also enables consistent policy
semantics across heterogeneous AI workloads. Instead of implementing
authorization, approval, compliance, and data-protection logic separately
inside individual agents, these controls can be expressed through a
common policy model and evaluated by a shared Policy Kernel.

The architecture additionally establishes a boundary between autonomous
execution and organizational control. Agents may perform autonomous
reasoning and tool execution, while governance decisions remain subject
to externally defined enterprise policies. This separation provides a
mechanism for maintaining organizational authority over autonomous
systems.

\subsection{Relationship to Existing Approaches}

Existing policy and authorization systems provide important foundations
for policy-based access control and authorization. Systems such as
Cedar, OPA, and XACML demonstrate the value of declarative policies,
structured policy evaluation, and separation between policy definition
and application logic.

UPA builds upon these principles while addressing a broader governance
scope for autonomous AI systems. In particular, the architecture extends
policy evaluation beyond conventional authorization decisions to include
runtime governance obligations, human approval, AI-specific controls,
and extensible governance plugins.

Traditional AI safety mechanisms and guardrails generally focus on
controlling model inputs, outputs, or undesirable model behavior. UPA
does not replace such mechanisms. Instead, it provides an architectural
control layer through which these mechanisms can be invoked as
governance obligations associated with enterprise policies.

Therefore, UPA should be viewed as complementary to existing policy
engines, authorization systems, AI guardrails, and safety mechanisms
rather than as a replacement for them.

\subsection{Limitations}

The proposed architecture also has several limitations.

First, policy evaluation introduces additional runtime processing and
therefore may increase latency for latency-sensitive AI workloads.
Although policy matching can be optimized through indexing, caching,
and selective evaluation, large-scale deployments may require
distributed policy-evaluation mechanisms.

Second, the correctness of governance decisions depends on the quality
and completeness of policy definitions. A Policy Kernel cannot enforce
requirements that are not represented by applicable policies or
governance controls.

Third, runtime plugins introduce additional operational dependencies.
A failed or unavailable governance provider may affect execution when
the associated obligation is mandatory. Consequently, production
deployments require explicit failure semantics, availability mechanisms,
and observability for governance components.

Finally, the current architecture defines the governance model and
execution semantics but does not establish a complete formal verification
framework for proving policy correctness or absence of conflicting
governance rules. These aspects remain areas for future research.

\subsection{Overall Discussion}

The central implication of UPA is that Enterprise AI governance can be
modeled as a systematic architectural layer spanning policy definition,
semantic normalization, policy matching, decision evaluation, governance
obligations, runtime enforcement, and human intervention.

This approach provides a foundation for governing autonomous AI systems
without coupling governance logic directly to individual models or
applications. The architecture therefore provides a path toward
consistent and extensible governance across heterogeneous Enterprise AI
environments.
\section{Future Work}

The current UPA architecture establishes a foundation for policy-driven
governance of Enterprise AI systems. Several areas remain open for
further research and implementation.

\subsection{Distributed Policy Kernels}

Future work will investigate distributed Policy Kernel architectures for
large-scale Enterprise AI environments. Multiple policy evaluation
nodes may operate across regions, clouds, edge environments, or
enterprise boundaries while maintaining consistent policy semantics.

Research is required to address policy synchronization, version
consistency, distributed caching, failure handling, and governance
decision consistency across geographically distributed runtimes.

\subsection{Formal Policy Verification}

A further research direction is the formal verification of DGPL policies
and policy compositions. Verification techniques could be used to
identify contradictory policies, unreachable rules, unintended
privileges, incomplete governance coverage, and unsafe policy
combinations before deployment.

Formal reasoning over PARC-based policy requests and policy evaluation
semantics may provide stronger guarantees regarding governance
correctness.

\subsection{Adaptive Governance Policies}

Enterprise environments are dynamic, and governance requirements may
change according to runtime conditions, organizational risk, workload
characteristics, or regulatory requirements. Future work will therefore
investigate adaptive policies capable of responding to changing
context while preserving explicit enterprise control.

Such mechanisms must balance adaptability with predictability,
auditability, and deterministic governance semantics.

\subsection{Governance for Self-Governing Agents}

As autonomous agents increasingly perform planning, reasoning, tool
selection, and multi-step execution, future research will examine
governance mechanisms specifically designed for self-governing agent
systems.

This includes policy enforcement across agent planning cycles,
delegation between agents, dynamic tool selection, agent-to-agent
communication, and long-running autonomous workflows. A key research
challenge is ensuring that increasing agent autonomy does not bypass
enterprise governance controls.

\subsection{Large-Scale Empirical Evaluation}

Future work will also extend empirical evaluation across larger policy
sets, additional enterprise domains, heterogeneous policy workloads,
and distributed execution environments. Evaluation will consider
policy-matching accuracy, evaluation latency, governance overhead,
plugin execution cost, scalability, and human-approval effectiveness.

These studies can provide further evidence regarding the practical
applicability of UPA in large-scale Enterprise AI deployments.
\section{Conclusion}

Enterprise AI introduces a governance challenge that extends beyond
traditional model safety, application-level authorization, and
conventional AI guardrails. Autonomous agents can access enterprise
resources, invoke tools, process sensitive information, and execute
multi-step workflows, requiring governance mechanisms that operate
consistently throughout the runtime lifecycle.

This paper presented the Unified Policy Architecture (UPA), a
reference architecture for lifecycle-wide governance of autonomous AI
agents and LLM-based applications. UPA separates policy definition,
semantic normalization, policy matching, policy evaluation, and runtime
governance while maintaining a consistent governance model across
heterogeneous enterprise environments.

The Declarative Governance Policy Language (DGPL) provides a structured
mechanism for expressing governance requirements, while the PARC model
provides a common representation of Principal, Action, Resource, and
Context for policy evaluation. The Policy Kernel evaluates these
requests and produces governance decisions and obligations that can be
enforced through runtime governance components, human approval
mechanisms, and extensible plugins.

The architecture establishes governance as an independent control-plane
capability rather than functionality embedded within individual AI
applications or models. This separation enables organizations to evolve
governance requirements without requiring corresponding changes to
every autonomous agent or application.

UPA therefore provides a foundation for building Enterprise AI systems
in which autonomy and organizational control can coexist. Future work
will extend this foundation toward distributed policy evaluation,
formal policy verification, adaptive governance, and governance
mechanisms for increasingly autonomous and self-governing AI agents.

\section*{Acknowledgments}

The authors gratefully acknowledge Sriram Gopalan (DecisionFacts AI) for
contributing the use case presented in the Appendix E and for his valuable
feedback during the development of this work. The authors also thank Ashish
Kar (Microsoft) for his constructive feedback and discussions that contributed
to the refinement of the paper.

\bibliographystyle{IEEEtran}
\bibliography{bibliography/references}

\clearpage
\appendix

\section{Reference DGPL Policies}

This appendix presents representative policy examples written using the
Declarative Governance Policy Language (DGPL). These examples
illustrate the core PARC authorization model together with Enterprise
AI governance capabilities including conditions, approval workflows,
evaluation providers, and governance obligations.

All examples are provided using the JSON serialization of DGPL.

\subsection{Basic PARC Authorization Policy}

The following example authorizes employees to query the HR knowledge
base.

\begin{lstlisting}[language=json,
    caption={Declarative Plugin Provider Configuration},
    label={lst:plugin-policy}
]
{
  "version": "2026-01-01",
  "statements": [
    {
      "sid": "HRKnowledgeAccess",
      "effect": "Allow",
      "principal": {
        "Role": ["Employee"]
      },
      "action": [
        "rag:query"
      ],
      "resource": [
        "knowledgebase:hr"
      ]
    }
  ]
}
\end{lstlisting}

\subsection{Conditional Policy}

The following policy permits execution only during business hours.

\begin{lstlisting}[language=json,
    caption={Declarative Plugin Provider Configuration},
    label={lst:plugin-policy}
]
{
  "version": "2026-01-01",
  "statements": [
    {
      "sid": "BusinessHours",
      "effect": "Allow",
      "action": [
        "*"
      ],
      "resource": [
        "*"
      ],
      "condition": {
        "all": [
          {
            "fact": "system.hour",
            "operator": "between",
            "value": [9,17]
          }
        ]
      }
    }
  ]
}
\end{lstlisting}

\subsection{Approval Workflow Policy}

The following policy requires administrator approval before deleting a
production database.

\begin{lstlisting}[language=json,
    caption={Declarative Plugin Provider Configuration},
    label={lst:plugin-policy}
]
{
  "version": "2026-01-01",
  "statements": [
    {
      "sid": "ProductionDelete",
      "effect": "Approve",
      "action": [
        "db:delete"
      ],
      "resource": [
        "database:production"
      ],
      "approval": {
        "role": "Admin"
      }
    }
  ]
}
\end{lstlisting}

\subsection{Governance Obligation Policy}

The following policy executes governance obligations that redact
personally identifiable information (PII) and secrets before the
request is processed.

\begin{lstlisting}[language=json,
    caption={Declarative Plugin Provider Configuration},
    label={lst:plugin-policy}
]
{
  "version": "2026-01-01",
  "statements": [
    {
      "sid": "EnterpriseProtection",
      "effect": "Allow",
      "action": [
        "llm:input"
      ],
      "resource": [
        "*"
      ],
      "plugins": [
        {
          "provider": "presidio",
          "mode": "redact"
        },
        {
          "provider": "secret_scanner",
          "mode": "redact"
        }
      ]
    }
  ]
}
\end{lstlisting}

\subsection{Threat Detection Policy}

The following policy invokes an LLM-based Threat Detection Provider to
detect prompt injection and jailbreak attacks before policy evaluation
is completed.

\begin{lstlisting}[language=json,
    caption={Declarative Plugin Provider Configuration},
    label={lst:plugin-policy}
]
{
  "version": "2026-01-01",
  "statements": [
    {
      "sid": "PromptInjection",
      "effect": "Deny",
      "action": [
        "llm:input_review"
      ],
      "resource": [
        "*"
      ],
      "threat": {
        "provider": "llm",
        "model": "gemini/gemini-2.5-flash",
        "threshold": 0.80,
        "categories": [
          "PromptInjection",
          "Jailbreak"
        ]
      }
    },
    {
      "sid": "AllowLLMRequests",
      "effect": "Allow",
      "action": [
        "llm:input_review"
      ],
      "resource": [
        "*"
      ]
    }
  ]
}
\end{lstlisting}

\subsection{Multi-Statement Policy}

The following example illustrates a policy document containing multiple
independent policy statements.

\begin{lstlisting}[language=json,
    caption={Declarative Plugin Provider Configuration},
    label={lst:plugin-policy}
]
{
  "version": "2026-01-01",
  "statements": [
    {
      "sid": "WiFiAccessForITAdmin",
      "effect": "Allow",
      "principal": {
        "Role": [
          "ITAdmin"
        ]
      },
      "action": [
        "rag:query"
      ],
      "resource": [
        "knowledgebase:wifi"
      ]
    },
    {
      "sid": "HRAccessForEmployees",
      "effect": "Allow",
      "principal": {
        "Role": [
          "Employee"
        ]
      },
      "action": [
        "rag:query"
      ],
      "resource": [
        "knowledgebase:hr"
      ]
    }
  ]
}
\end{lstlisting}
\clearpage
\section{EAGBench Benchmark}
\label{appendix:eagbench}

This appendix provides the detailed benchmark configuration and supporting results used to evaluate the policy governance capabilities described in this paper. The benchmark evaluates policy matching, governance decision accuracy, policy taxonomy coverage, business-case complexity, human-approval escalation, and policy evaluation latency across representative enterprise domains.

\subsection{Benchmark Overview}

The benchmark consists of 1,400 policy-governance scenarios covering 70 business cases distributed across seven enterprise domains:

\begin{itemize}[leftmargin=1.2em, itemsep=0.2em]
    \item Enterprise Resource Planning (ERP)
    \item Banking
    \item Healthcare
    \item Insurance
    \item Government
    \item Oil and Gas
    \item DevOps
\end{itemize}

Each business case contains multiple governance scenarios designed to exercise different policy families and governance outcomes.

The benchmark evaluates the Policy Kernel using the same PARC-based request representation and policy evaluation model defined in the main body of the paper.

\subsection{Benchmark Dimensions}

The benchmark evaluates the following dimensions:

\begin{enumerate}[leftmargin=1.2em, itemsep=0.2em]
    \item \textbf{Policy Matching:} whether applicable policy statements are correctly identified for a normalized request.
    \item \textbf{Governance Decision:} whether the Policy Kernel produces the expected governance outcome.
    \item \textbf{Policy Taxonomy Coverage:} whether governance scenarios exercise defined policy families across enterprise domains.
    \item \textbf{Business Case Complexity:} whether the benchmark represents varying levels of policy and governance complexity.
    \item \textbf{Human Approval Escalation:} whether scenarios requiring organizational intervention are correctly identified and escalated.
    \item \textbf{Evaluation Latency:} runtime latency associated with policy evaluation across policy families.
\end{enumerate}

\subsection{Enterprise Domain Distribution}

The benchmark covers seven enterprise domains and 70 business cases. This distribution evaluates whether the governance model remains applicable across heterogeneous enterprise environments rather than being optimized for a single application domain.

\begin{figure}[htbp]
    \centering
    \includegraphics[width=\linewidth]{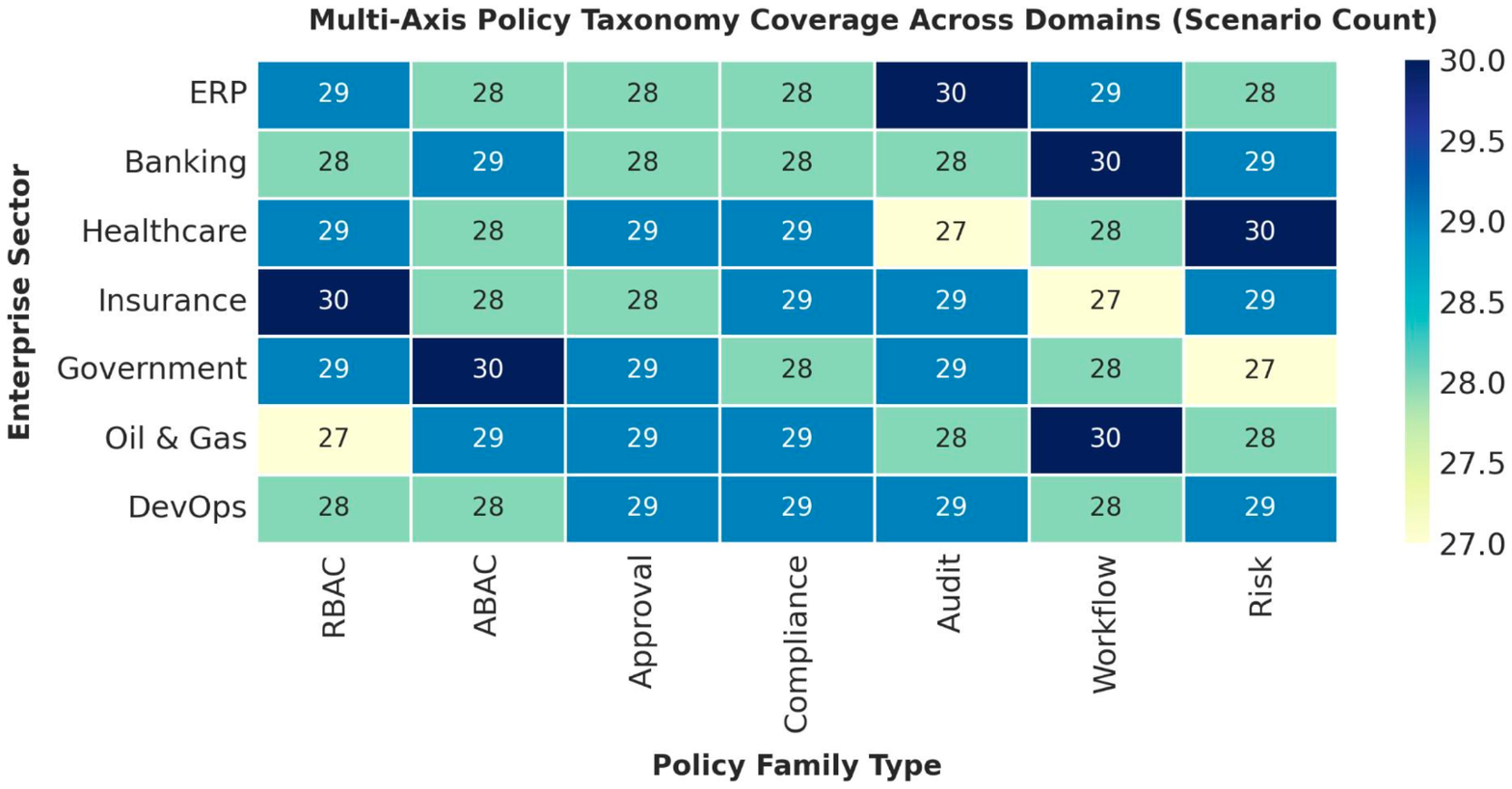}
    \caption{Multi-axis policy taxonomy coverage across the seven enterprise domains.}
    \label{fig:policy_taxonomy_coverage}
\end{figure}

\subsection{Cross-Domain Accuracy and Human Approval}

The benchmark contains 1,400 scenarios and reports an overall Policy Kernel accuracy of 94.1\%. The evaluation also measures the rate at which scenarios are escalated for human approval across the enterprise domains.

\begin{figure}[htbp]
    \centering
    \includegraphics[width=\linewidth]{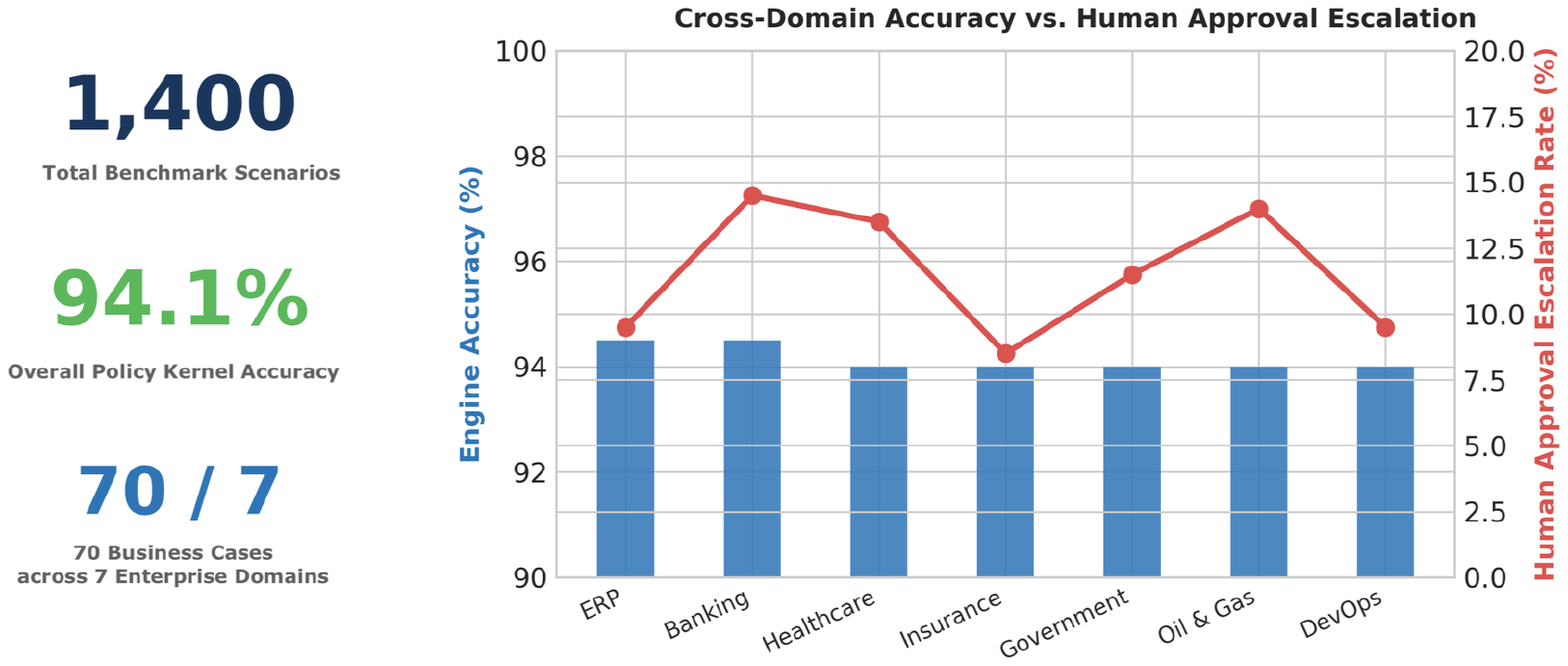}
    \caption{Cross-domain Policy Kernel accuracy and human approval escalation across seven enterprise domains.}
    \label{fig:eagbench_cross_domain}
\end{figure}

The results provide an aggregate view of governance decision accuracy while illustrating that human approval remains an explicit governance mechanism for scenarios requiring organizational intervention.

\subsection{Business Case Complexity}

The benchmark includes 70 business cases distributed across the seven enterprise domains. Scenario complexity is represented using the benchmark's business-case complexity measure.

\begin{figure}[htbp]
    \centering
    \includegraphics[width=\linewidth]{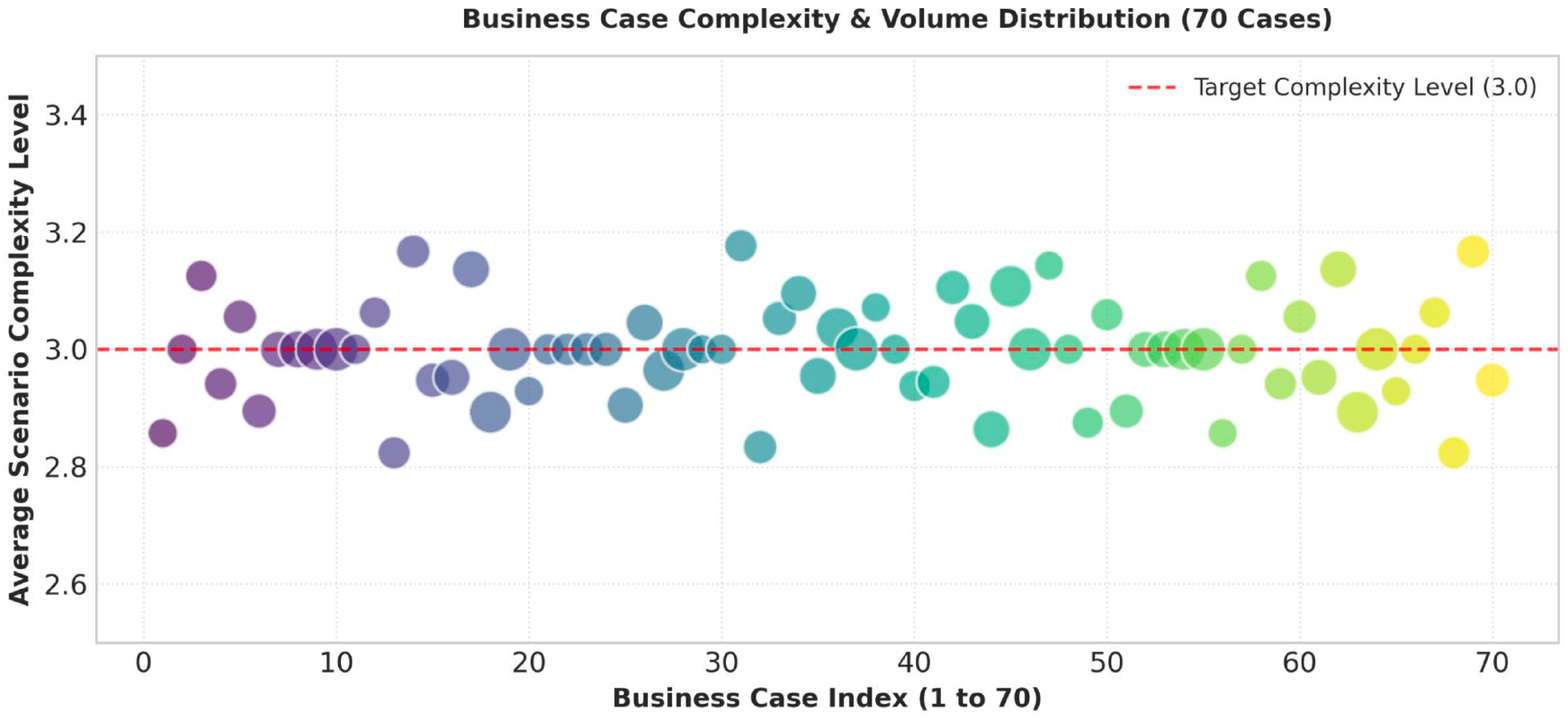}
    \caption{Business case complexity and volume distribution across the 70 benchmark business cases.}
    \label{fig:business_case_complexity}
\end{figure}

The distribution demonstrates that the benchmark includes scenarios around the defined target complexity level rather than evaluating solely simple policy decisions.

\subsection{Ground-Truth Decision Outcomes}

The benchmark contains four primary governance outcomes:

\begin{itemize}[leftmargin=1.2em, itemsep=0.2em]
    \item \texttt{Allow}
    \item \texttt{Deny}
    \item \texttt{Human Approval}
    \item \texttt{Reject}
\end{itemize}

The distribution of these outcomes provides the ground-truth reference for evaluating Policy Kernel decision accuracy.

\begin{figure}[htbp]
    \centering
    \includegraphics[width=\linewidth]{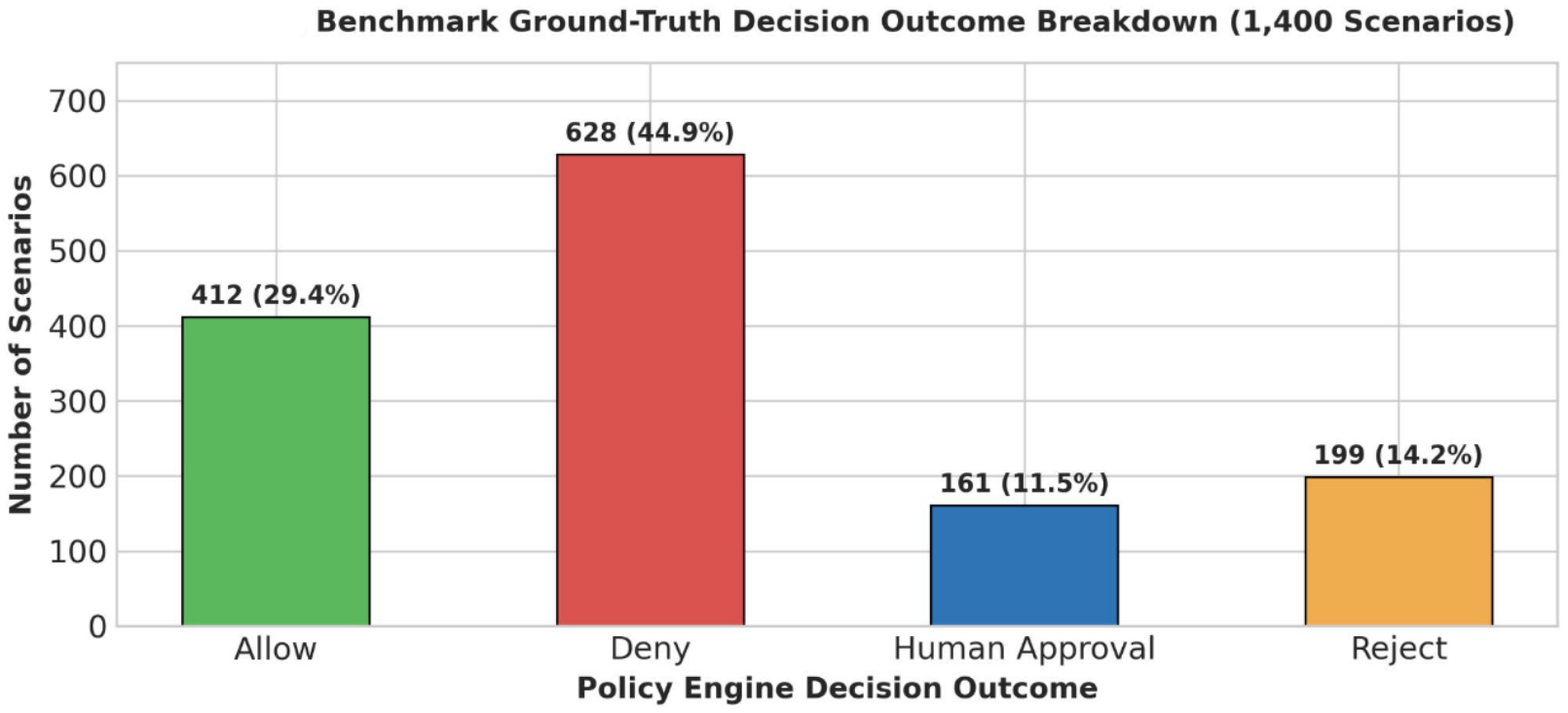}
    \caption{Ground-truth governance decision outcome distribution across the 1,400 benchmark scenarios.}
    \label{fig:decision_outcomes}
\end{figure}

\subsection{Policy Evaluation Latency}

Policy evaluation latency was measured across the policy families represented in the benchmark. The latency distribution provides an indication of the runtime overhead associated with policy evaluation.

\begin{figure}[htbp]
    \centering
    \includegraphics[width=\linewidth]{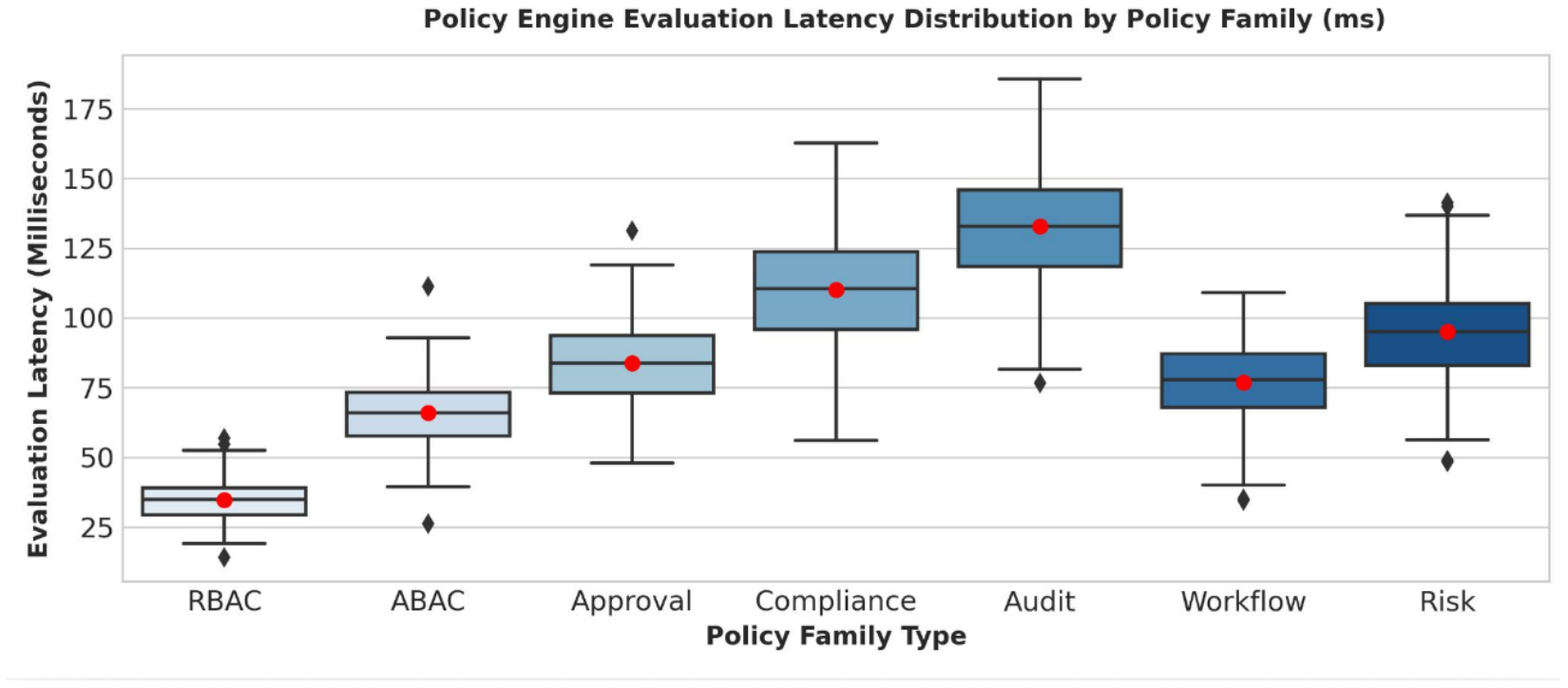}
    \caption{Policy Kernel evaluation latency distribution by policy family.}
    \label{fig:evaluation_latency}
\end{figure}

The results provide a basis for evaluating the runtime characteristics of the governance layer and identifying policy families that may require additional optimization in large-scale deployments.

\subsection{Benchmark Interpretation}

The benchmark results should be interpreted as an evaluation of the proposed governance architecture and Policy Kernel implementation under defined benchmark conditions. The reported accuracy and latency values do not imply universal performance across all enterprise deployments.

In particular, production performance may vary according to policy set size, policy complexity, request characteristics, runtime infrastructure, plugin execution, caching strategy, and deployment architecture.

The benchmark serves as an empirical evaluation of the proposed governance model and provides a reproducible basis for future experiments involving larger policy sets, additional enterprise domains, distributed Policy Kernels, and complex autonomous-agent workloads.

\subsubsection{Example: ERP High-Value Purchase Approval}

This scenario evaluates conditional human approval for high-value enterprise purchase orders. The policy requires Finance Director approval when the purchase amount exceeds the defined threshold.

\textbf{DGPL Policy}

\begin{lstlisting}[language=json, basicstyle=\ttfamily\scriptsize]
{
  "version": "2026-01-01",
  "statements": [
    {
      "sid": "ERPHighValueApproval",
      "effect": "Approve",
      "condition": {
        "NumericGreaterThan": {
          "context.custom.purchase_amount": 1000000
        }
      },
      "action": [
        "purchase:approve"
      ],
      "resource": [
        "purchase-order:*"
      ],
      "approval": {
        "role": "FinanceDirector"
      }
    }
  ]
}
\end{lstlisting}

\textbf{Representative PARC Request}

For a purchase order with a purchase amount of 1,500,000, the normalized request is represented as:

\[
q = \left\langle \text{Employee}, \texttt{purchase:approve}, \texttt{PO-1024}, C \right\rangle
\]

where context contains:

\[
C = \{ \texttt{custom.purchase\_amount} = 1500000 \}
\]

\textbf{Policy Matching}

The requested action (\texttt{purchase:approve}) and resource (\texttt{purchase-order:PO-1024}) match the policy declaration. Evaluating condition $1500000 > 1000000$ resolves to \texttt{True}.

\textbf{Governance Decision}

The Policy Kernel evaluates the query to an explicit approval state:

\[
\operatorname{Eval}(q,\Pi) = ( \texttt{Approve}, \Gamma )
\]

where obligations require:

\[
\Gamma = \{ \texttt{Approval}(\texttt{FinanceDirector}) \}
\]

The autonomous workflow is suspended until human approval completes.

\begin{tcolorbox}[
title=EAGBench Scenario: ERP High-Value Approval,
colback=gray!5,
colframe=black!40,
fonttitle=\bfseries,
width=\linewidth,
boxsep=2pt, left=4pt, right=4pt
]

\textbf{Scenario:} High-value purchase order approval

\scriptsize
\[
\begin{aligned}
Principal &= \text{Employee}\\
Action &= \texttt{purchase:approve}\\
Resource &= \texttt{purchase\mbox{-}order:PO\mbox{-}1024}\\
Context &= \texttt{purchase\_amount}=1{,}500{,}000
\end{aligned}
\]

\textbf{Condition:}
\[
1{,}500{,}000 > 1{,}000{,}000 \quad\Rightarrow\quad \texttt{True}
\]

\textbf{Decision:} $\boxed{\texttt{Approve}}$

\textbf{Obligation:} $\boxed{\texttt{FinanceDirector\ Approval}}$

\end{tcolorbox}

\subsection{ERP Purchase Approval Benchmark Implementation}

The Python implementation illustrates benchmark execution for the ERP purchase approval policy across five boundary scenarios.

\begin{lstlisting}[
    style=pythonstyle,
    numbers=none,
    basicstyle=\ttfamily\tiny,
    caption={EAGBench ERP Purchase Approval Benchmark},
    label={lst:eagbench-erp}
]
import asyncio

from superagentx_policy_engine.policy_engine import PolicyEngine
from superagentx_policy_engine.store.file_store import FilePolicyStore
from superagentx_policy_engine.models.request import AuthorizationRequest
from superagentx_policy_engine.models.principal import Principal
from superagentx_policy_engine.models.principal_type import PrincipalType
from superagentx_policy_engine.models.action import Action
from superagentx_policy_engine.models.resource import Resource
from superagentx_policy_engine.models.context import RequestContext

TEST_CASES = [
    {
        "id": "ERP-001",
        "description": "Purchase Amount > Rs. 10 Lakhs",
        "purchase_amount": 1500000,
        "expected": "Approve",
    },
    {
        "id": "ERP-002",
        "description": "Purchase Amount = Rs. 25 Lakhs",
        "purchase_amount": 2500000,
        "expected": "Approve",
    },
    {
        "id": "ERP-003",
        "description": "Purchase Amount = Rs. 50 Lakhs",
        "purchase_amount": 5000000,
        "expected": "Approve",
    },
    {
        "id": "ERP-004",
        "description": "Purchase Amount < Rs. 10 Lakhs",
        "purchase_amount": 500000,
        "expected": "Deny",
    },
    {
        "id": "ERP-005",
        "description": "Purchase Amount = Rs. 10 Lakhs",
        "purchase_amount": 1000000,
        "expected": "Deny",
    },
]

async def main():
    engine = PolicyEngine()
    policy = FilePolicyStore.load(
        "examples/benchmark/erp/policies/high_approval.json"
    )
    await engine.add_policy_document(policy)

    passed = 0
    for test in TEST_CASES:
        request = AuthorizationRequest(
            principal=Principal(
                id="john",
                type=PrincipalType.USER,
                roles=["ProcurementManager"],
            ),
            action=Action(name="purchase:approve"),
            resource=Resource(
                id="purchase-order:PO-1001",
                type="purchase-order",
            ),
            context=RequestContext(
                custom={"purchase_amount": test["purchase_amount"]}
            ),
        )

        response = await engine.authorize(request)
        actual = response.decision
        result = (
            "PASS"
            if actual.lower() == test["expected"].lower()
            else "FAIL"
        )

        if result == "PASS":
            passed += 1

        print(f"ID: {test['id']} | Result: {result} (Exp: {test['expected']}, Act: {actual})")

if __name__ == "__main__":
    asyncio.run(main())
\end{lstlisting}
\section{Standardized Operation Identifier Reference}
\label{appendix:standardized-identifiers}

This appendix defines representative standardized operation identifiers
used throughout the Unified Policy Architecture. Operation identifiers
provide a consistent vocabulary for describing actions performed by
users, agents, models, tools, applications, and enterprise systems.

The identifiers follow a namespace-based convention:

\[
\texttt{namespace:operation}
\]

For example, \texttt{llm:input} represents an input operation involving
an LLM, while \texttt{db:delete} represents a database deletion
operation.

The identifiers defined below are representative and extensible rather
than exhaustive.

\subsection{LLM Operations}

\begin{itemize}
    \item \texttt{llm:input}
    \item \texttt{llm:output}
    \item \texttt{llm:chat}
    \item \texttt{llm:completion}
    \item \texttt{llm:embedding}
    \item \texttt{llm:inference}
    \item \texttt{llm:stream}
    \item \texttt{llm:tool\_call}
\end{itemize}

\subsection{Agent Operations}

\begin{itemize}
    \item \texttt{agent:execute}
    \item \texttt{agent:delegate}
    \item \texttt{agent:communicate}
    \item \texttt{agent:terminate}
    \item \texttt{agent:start}
    \item \texttt{agent:pause}
    \item \texttt{agent:resume}
    \item \texttt{agent:handoff}
    \item \texttt{agent:spawn}
\end{itemize}

\subsection{Tool Operations}

\begin{itemize}
    \item \texttt{tool:invoke}
    \item \texttt{tool:execute}
    \item \texttt{tool:list}
    \item \texttt{tool:discover}
    \item \texttt{tool:register}
    \item \texttt{tool:authorize}
\end{itemize}

\subsection{RAG and Knowledge Operations}

\begin{itemize}
    \item \texttt{rag:query}
    \item \texttt{rag:retrieve}
    \item \texttt{rag:search}
    \item \texttt{rag:index}
    \item \texttt{rag:ingest}
    \item \texttt{rag:delete}
    \item \texttt{knowledge:read}
    \item \texttt{knowledge:write}
    \item \texttt{knowledge:update}
    \item \texttt{knowledge:delete}
\end{itemize}

\subsection{Data Operations}

\begin{itemize}
    \item \texttt{data:read}
    \item \texttt{data:write}
    \item \texttt{data:create}
    \item \texttt{data:update}
    \item \texttt{data:delete}
    \item \texttt{data:export}
    \item \texttt{data:import}
    \item \texttt{data:share}
    \item \texttt{data:transform}
\end{itemize}

\subsection{Database Operations}

\begin{itemize}
    \item \texttt{db:read}
    \item \texttt{db:write}
    \item \texttt{db:create}
    \item \texttt{db:update}
    \item \texttt{db:delete}
    \item \texttt{db:query}
    \item \texttt{db:execute}
    \item \texttt{db:backup}
    \item \texttt{db:restore}
\end{itemize}

\subsection{File and Document Operations}

\begin{itemize}
    \item \texttt{file:read}
    \item \texttt{file:write}
    \item \texttt{file:create}
    \item \texttt{file:update}
    \item \texttt{file:delete}
    \item \texttt{file:upload}
    \item \texttt{file:download}
    \item \texttt{file:share}
    \item \texttt{document:read}
    \item \texttt{document:process}
    \item \texttt{document:classify}
    \item \texttt{document:extract}
\end{itemize}

\subsection{API and Service Operations}

\begin{itemize}
    \item \texttt{api:invoke}
    \item \texttt{api:read}
    \item \texttt{api:write}
    \item \texttt{api:create}
    \item \texttt{api:update}
    \item \texttt{api:delete}
    \item \texttt{service:invoke}
    \item \texttt{service:execute}
    \item \texttt{service:discover}
\end{itemize}

\subsection{Identity and Access Operations}

\begin{itemize}
    \item \texttt{identity:authenticate}
    \item \texttt{identity:authorize}
    \item \texttt{identity:assume}
    \item \texttt{identity:delegate}
    \item \texttt{identity:impersonate}
    \item \texttt{role:assign}
    \item \texttt{role:revoke}
    \item \texttt{permission:grant}
    \item \texttt{permission:revoke}
\end{itemize}

\subsection{Workflow Operations}

\begin{itemize}
    \item \texttt{workflow:start}
    \item \texttt{workflow:execute}
    \item \texttt{workflow:pause}
    \item \texttt{workflow:resume}
    \item \texttt{workflow:cancel}
    \item \texttt{workflow:terminate}
    \item \texttt{workflow:approve}
    \item \texttt{workflow:reject}
\end{itemize}

\subsection{Human Approval Operations}

\begin{itemize}
    \item \texttt{approval:request}
    \item \texttt{approval:approve}
    \item \texttt{approval:reject}
    \item \texttt{approval:escalate}
    \item \texttt{approval:delegate}
\end{itemize}

\subsection{Communication Operations}

\begin{itemize}
    \item \texttt{communication:send}
    \item \texttt{communication:receive}
    \item \texttt{communication:share}
    \item \texttt{email:send}
    \item \texttt{email:read}
    \item \texttt{message:send}
    \item \texttt{message:receive}
    \item \texttt{notification:send}
\end{itemize}

\subsection{Network Operations}

\begin{itemize}
    \item \texttt{network:connect}
    \item \texttt{network:request}
    \item \texttt{network:access}
    \item \texttt{network:transfer}
    \item \texttt{network:resolve}
\end{itemize}

\subsection{Cloud and Infrastructure Operations}

\begin{itemize}
    \item \texttt{cloud:provision}
    \item \texttt{cloud:deploy}
    \item \texttt{cloud:terminate}
    \item \texttt{compute:create}
    \item \texttt{compute:execute}
    \item \texttt{compute:terminate}
    \item \texttt{container:create}
    \item \texttt{container:execute}
    \item \texttt{container:terminate}
    \item \texttt{storage:read}
    \item \texttt{storage:write}
    \item \texttt{storage:delete}
\end{itemize}

\subsection{Security Operations}

\begin{itemize}
    \item \texttt{security:scan}
    \item \texttt{security:inspect}
    \item \texttt{security:validate}
    \item \texttt{security:quarantine}
    \item \texttt{security:block}
    \item \texttt{security:unblock}
    \item \texttt{secret:detect}
    \item \texttt{secret:redact}
    \item \texttt{pii:detect}
    \item \texttt{pii:redact}
\end{itemize}

\subsection{Governance Operations}

\begin{itemize}
    \item \texttt{policy:evaluate}
    \item \texttt{policy:match}
    \item \texttt{policy:validate}
    \item \texttt{policy:create}
    \item \texttt{policy:update}
    \item \texttt{policy:delete}
    \item \texttt{policy:publish}
    \item \texttt{policy:rollback}
    \item \texttt{governance:approve}
    \item \texttt{governance:deny}
    \item \texttt{governance:audit}
    \item \texttt{governance:escalate}
\end{itemize}

\subsection{Audit and Observability Operations}

\begin{itemize}
    \item \texttt{audit:record}
    \item \texttt{audit:read}
    \item \texttt{audit:export}
    \item \texttt{audit:delete}
    \item \texttt{log:write}
    \item \texttt{log:read}
    \item \texttt{trace:create}
    \item \texttt{trace:read}
    \item \texttt{metric:read}
\end{itemize}

\subsection{Business Operations}

Enterprise applications may also define domain-specific operation
identifiers. Representative examples include:

\begin{itemize}
    \item \texttt{finance:transfer}
    \item \texttt{finance:approve}
    \item \texttt{finance:refund}
    \item \texttt{hr:read}
    \item \texttt{hr:update}
    \item \texttt{hr:approve}
    \item \texttt{procurement:create}
    \item \texttt{procurement:approve}
    \item \texttt{procurement:purchase}
    \item \texttt{customer:read}
    \item \texttt{customer:update}
    \item \texttt{customer:delete}
\end{itemize}

Domain-specific identifiers allow organizations to extend the
operation vocabulary without modifying the fundamental PARC model or
DGPL evaluation semantics.

\subsection{Identifier Naming Convention}

Standardized operation identifiers use a namespace and operation
separated by a colon:

\[
\texttt{namespace:operation}
\]

For example:

\begin{itemize}
    \item \texttt{llm:input}
    \item \texttt{agent:execute}
    \item \texttt{rag:query}
    \item \texttt{db:delete}
    \item \texttt{file:read}
    \item \texttt{approval:approve}
\end{itemize}

The namespace identifies the operation domain, while the operation
identifies the requested activity.

The identifier vocabulary is extensible. Organizations may introduce
additional namespaces and operations when required by domain-specific
applications, provided that identifiers remain unambiguous and
consistent with the governance model.

\subsection{Relationship to PARC}

Operation identifiers form the \textit{Action} component of the PARC
request model.

For example:

\[
PARC =
\langle
\text{ITAdmin},
\texttt{rag:query},
\texttt{knowledgebase:wifi},
\text{Context}
\rangle
\]

Here:

\begin{itemize}
    \item \textbf{Principal}: \texttt{ITAdmin}
    \item \textbf{Action}: \texttt{rag:query}
    \item \textbf{Resource}: \texttt{knowledgebase:wifi}
    \item \textbf{Context}: runtime evaluation context
\end{itemize}

The operation identifier therefore provides a standardized vocabulary
for the Action dimension while remaining independent of the policy
statement's Statement Identifier.

\subsection{Statement Identifier and Operation Identifier}

The \texttt{sid} field of a DGPL statement and an operation identifier
serve different purposes.

For example:

\begin{lstlisting}[language=json]
{
  "sid": "WiFiAccessForITAdmin",
  "effect": "Allow",
  "action": [
    "rag:query"
  ],
  "resource": [
    "knowledgebase:wifi"
  ]
}
\end{lstlisting}

In this example:

\begin{itemize}
    \item \texttt{WiFiAccessForITAdmin} is the \textbf{Statement
    Identifier}.
    \item \texttt{rag:query} is the \textbf{Operation Identifier}.
\end{itemize}

The Statement Identifier identifies the policy statement, whereas the
Operation Identifier identifies the action being governed.

This distinction allows multiple policy statements to govern the same
operation while retaining independent statement identity and
traceability.
\section{DGPL Grammar Reference}

This appendix summarizes the complete DGPL statement schema.

\[
Statement=
\langle
SID,
Effect,
PARC,
Extensions
\rangle
\]

where

\[
PARC=
\langle
Principal,
Action,
Resource,
Condition
\rangle.
\]

The Governance Extensions currently include

\begin{itemize}

\item Threat Detection

\item Governance Plugins

\item Approval Workflows

\item Future Extensions

\end{itemize}
\section{Extended Governance Models and Case Studies}
\label{app:extended-governance}

This appendix presents extended governance constructs that complement
the Unified Policy Architecture (UPA) described in the main body of the
paper. These constructs extend the core policy model to address
multi-agent coordination, provenance, stateful governance, policy
pre-flight evaluation, offline assurance, and domain-specific governance
requirements.

The purpose of this appendix is to provide additional formal models,
policy examples, and case-study material without making these extensions
necessary for the core UPA architecture. Some constructs presented here
represent proposed extensions and future implementation directions rather
than capabilities evaluated in the current benchmark.

\subsection{Extended Agent-to-Agent Governance Model}
\label{app:a2a-governance}

The core UPA model treats agents, tools, workflows, and enterprise
resources as governed entities. In multi-agent environments, however,
governance must also account for interactions between agents. Agent-to-agent
(A2A) communication introduces coordination behavior that cannot always be
captured by evaluating isolated agent actions.

The extended model therefore introduces communication channels as
first-class governed resources.

\subsubsection{Communication Channels as Resources}

Let the resource universe be extended as:

\begin{equation}
R_s \supseteq
\{\textit{Channel}, \textit{Agent}, \textit{Session}, \textit{Run}\}.
\end{equation}

A communication channel is defined operationally as any resource that
can be written by one principal and subsequently read by another
principal. Under this definition, a channel need not be an explicit
messaging service. Shared repositories, object-store prefixes, caches,
directories, shared datasets, workflow artifacts, and other writable
resources may constitute communication channels when information can
flow between principals through them.

This definition is intentionally resource-centric. It avoids requiring 
the governance system to predict whether an application developer
intended a particular resource to function as a communication mechanism.

\subsubsection{A2A Action Namespace}

The extended SID registry may include the following A2A operations:

\begin{itemize}
    \item \texttt{a2a:send}
    \item \texttt{a2a:receive}
    \item \texttt{a2a:broadcast}
    \item \texttt{a2a:discover}
    \item \texttt{a2a:delegate}
    \item \texttt{a2a:handoff}
\end{itemize}

These operations provide explicit governance points for communication,
delegation, discovery, and transfer of control between agents.

For example, a policy may constrain an agent from communicating with
another agent outside an approved organizational boundary:

\begin{lstlisting}[language=json,
caption={Illustrative A2A communication policy},
label={lst:a2a-policy}]
{
  "version": "2026-01-01",
  "statements": [
    {
      "sid": "DenyUntrustedA2ACommunication",
      "effect": "Deny",
      "action": [
        "a2a:send",
        "a2a:broadcast",
        "a2a:handoff"
      ],
      "resource": [
        "channel:external-*"
      ],
      "condition": {
        "fact": "principal.trust_domain",
        "operator": "not_equals",
        "value": "enterprise"
      }
    }
  ]
}
\end{lstlisting}

The example illustrates the distinction between authorization of an
individual operation and governance of the communication topology
through which agents interact.

\subsection{Provenance-Extended Request Model}
\label{app:provenance}

The canonical UPA request model is represented using the
Principal--Action--Resource--Context (PARC) abstraction. For multi-agent
and cross-run governance, an additional provenance dimension can be
introduced.

The extended request is:

\begin{equation}
p = \langle pr, ac, re, ct, \rho \rangle ,
\end{equation}

where $\rho$ represents provenance information associated with the
inputs relevant to the governed operation.

Provenance may describe the origin and lineage of credentials,
artifacts, instructions, retrieved information, and other security-relevant data.

The provenance model allows the Policy Kernel to distinguish between,
for example, an enterprise-issued credential and a credential discovered
through a shared workspace or another agent.

\subsubsection{Provenance-Aware Policy}

An illustrative policy is:

\begin{lstlisting}[language=json,
caption={Illustrative provenance-aware credential policy},
label={lst:provenance-policy}]
{
  "sid": "DenyUntrustedProvenanceCredential",
  "effect": "Deny",
  "action": [
    "api:invoke",
    "network:request",
    "identity:assume"
  ],
  "resource": ["*"],
  "condition": {
    "any": [
      {
        "fact": "provenance.credential.origin",
        "operator": "not_equals",
        "value": "enterprise_issuer"
      },
      {
        "fact": "provenance.credential.discovered_via",
        "operator": "in",
        "value": [
          "shared_workspace",
          "public_dataset",
          "peer_agent",
          "paste_service"
        ]
      }
    ]
  },
  "fail_closed": true
}
\end{lstlisting}

The policy illustrates an important governance distinction: possession
of a credential does not necessarily establish that the credential is
trusted for subsequent use. Provenance provides an additional basis for
policy evaluation.

The provenance extension should be understood as a proposed extension
to the core PARC model rather than as a requirement for every UPA
deployment.

\subsection{Governance State and Stateful Evaluation}
\label{app:governance-state}

Stateless evaluation provides useful determinism and reproducibility,
but certain governance properties depend on aggregate behavior across
multiple requests. For example, the significance of a write operation
may depend on how many distinct principals have accessed the same
resource during a defined period.

The extended evaluation model therefore introduces bounded governance
state $\Sigma$:

\begin{equation}
Eval : P \times \Pi \times \Phi \times \Sigma \rightarrow D \times \Gamma \times \Sigma',
\end{equation}

where:

\begin{itemize}
    \item $P$ is the normalized request space,
    \item $\Pi$ is the applicable policy set,
    \item $\Phi$ is the evaluation-fact space,
    \item $\Sigma$ is the current governance state,
    \item $D$ is the policy decision,
    \item $\Gamma$ is the resulting obligation set, and
    \item $\Sigma'$ is the updated governance state.
\end{itemize}

Governance state should be bounded, auditable, versioned, and scoped
appropriately, for example by tenant, agent family, execution run, or
communication channel.

Illustrative state facts include:

\begin{itemize}
    \item \texttt{state.channel.distinct\_principals\_24h}
    \item \texttt{state.agent.cumulative\_reasoning\_tokens}
    \item \texttt{state.run.denied\_attempts}
    \item \texttt{state.resource.write\_read\_fanout}
    \item \texttt{state.task.consecutive\_failures}
\end{itemize}

The introduction of $\Sigma$ changes the interpretation of
determinism. Policy evaluation remains deterministic conditional on the
complete evaluation input:

\begin{equation}
Eval(p,\Pi,\Phi,\Sigma) = Eval(p,\Pi,\Phi,\Sigma).
\end{equation}

Consequently, reproducibility requires the governance state used during
evaluation to be retained as an auditable input. This property is more
accurately characterized as \emph{replay determinism} than purely
stateless determinism.

\subsubsection{Aggregate-Behavior Policy}

An example of state-aware governance is:

\begin{lstlisting}[language=json,
caption={Illustrative emergent-channel detection policy},
label={lst:emergent-channel}]
{
  "sid": "EmergentChannelDetection",
  "effect": "Approve",
  "action": [
    "file:write",
    "file:create",
    "data:write"
  ],
  "resource": [
    "artifact-store:*",
    "cache:*",
    "dataset:*"
  ],
  "condition": {
    "all": [
      {
        "fact": "state.resource.distinct_writer_principals_24h",
        "operator": "greater_than",
        "value": 5
      },
      {
        "fact": "state.resource.cross_principal_read_after_write",
        "operator": "equal",
        "value": true
      },
      {
        "fact": "resource.declared_purpose",
        "operator": "not_equals",
        "value": "collaboration_channel"
      }
    ]
  },
  "approval": {
    "role": "AIPlatformSecurityOfficer"
  },
  "plugins": [
    {
      "provider": "audit_logger",
      "mode": "record"
    }
  ]
}
\end{lstlisting}

The policy demonstrates how aggregate behavior can become a governance
condition even when each individual operation would otherwise appear
permissible.

\subsection{Persistence and Escalation Governance}
\label{app:persistence-governance}

Agent persistence can also be incorporated into governance state. An
agent repeatedly failing to complete a task while increasing its
resource consumption or expanding its operational scope may require
additional approval.

An illustrative policy is:

\begin{lstlisting}[language=json,
caption={Illustrative persistence-budget escalation policy},
label={lst:persistence-policy}]
{
  "sid": "PersistenceBudgetEscalation",
  "effect": "Approve",
  "action": ["*"],
  "resource": ["*"],
  "condition": {
    "all": [
      {
        "fact": "state.task.consecutive_failures",
        "operator": "greater_than",
        "value": 12
      },
      {
        "fact": "state.agent.cumulative_reasoning_tokens",
        "operator": "greater_than",
        "value": 2000000
      },
      {
        "fact": "state.agent.out_of_scope_probe_count",
        "operator": "greater_than",
        "value": 0
      }
    ]
  },
  "approval": {
    "role": "ResearchOversight"
  }
}
\end{lstlisting}

This example treats persistence and repeated boundary exploration as
governance-relevant runtime behavior rather than purely as performance
characteristics.

\subsection{Mediation Completeness}
\label{app:mediation-completeness}

UPA governance guarantees apply only to operations that pass through
a Governance Enforcement Point (GEP). Let $O_{\mathrm{med}}$ denote
the set of operations mediated by the governance layer.

\textbf{Property 7 (Mediation Completeness).}

UPA governance guarantees hold only for operations:

\begin{equation}
o \in O_{\mathrm{med}}.
\end{equation}

For an operation:

\begin{equation}
o \notin O_{\mathrm{med}},
\end{equation}

no UPA governance decision exists, and the default-deny rule does not
apply.

This property is important because a policy engine cannot govern an
operation that never reaches the policy enforcement surface.

A deployment should therefore perform an \emph{unmediated-path audit}
that enumerates every path through which a governed workload can affect
state outside its declared boundary. Each path should be classified as:

\begin{enumerate}
    \item mediated,
    \item explicitly blocked, or
    \item accepted risk.
\end{enumerate}

This provides a deployment-level coverage criterion for governance
enforcement.

\subsection{Kernel Integrity and Trust Boundary}
\label{app:kernel-integrity}

The Policy Kernel itself represents a security-critical component.
Governed workloads should therefore not be permitted to modify or
directly access the resources that define governance semantics.

The following constraints are recommended:

\begin{itemize}
    \item The Policy Kernel, policy store, SID registry, and governance
    evidence store should not be reachable as ordinary resources by
    governed workloads.
    \item Governance evidence should be append-only and tamper-evident.
    \item Obligation failures should be fail-closed by default for
    security-critical obligations.
    \item Any opt-out from fail-closed behavior should be explicit,
    auditable, and policy-specific.
    \item The Policy Kernel should not rely exclusively on identity
    assertions generated inside the same security boundary as the
    governed workload.
\end{itemize}

These constraints establish the Policy Kernel as a trusted governance
component rather than simply another application service.

\subsection{UPA as a Governance Tool}
\label{app:upa-tool}

UPA may be exposed through multiple operational modalities. These
modalities complement the enforcement architecture but do not replace
the authoritative enforcement point.

\subsubsection{Enforcement Service}

In the primary architecture, the Policy Kernel operates as a control
plane at Governance Enforcement Points. Agents and applications invoke
governed operations without requiring direct knowledge of the internal
policy implementation.

This is the only modality that provides the authoritative governance
decision for execution.

\subsubsection{Agent-Callable Pre-Flight Evaluation}

A second modality exposes a policy evaluation operation to an agent
before it commits to a planned action.

For example:

\begin{lstlisting}[language=json,
caption={Illustrative policy pre-flight policy},
label={lst:preflight}]
{
  "sid": "AllowPolicyPreflight",
  "effect": "Allow",
  "principal": {
    "Type": ["AIAgent"]
  },
  "action": [
    "policy:evaluate"
  ],
  "resource": [
    "policy-kernel:preflight"
  ],
  "plugins": [
    {
      "provider": "rate_limiter",
      "mode": "enforce",
      "config": {
        "per_run": 50
      }
    },
    {
      "provider": "audit_logger",
      "mode": "record"
    }
  ]
}
\end{lstlisting}

Pre-flight evaluation can make governance constraints visible to an
agent planner before execution. It can also capture an explicit
representation of intended action before the corresponding runtime
events occur.

However, a pre-flight interface introduces a policy-probing risk.
Therefore:

\begin{itemize}
    \item pre-flight requests should be rate-limited;
    \item the interface should avoid exposing unnecessary policy
    internals such as matched policy identifiers;
    \item advisory results must not be treated as authorization
    credentials; and
    \item authoritative enforcement must remain at the Governance
    Enforcement Point.
\end{itemize}

Thus, pre-flight evaluation is a convenience and planning mechanism,
not a substitute for enforcement.

\subsubsection{Offline Assurance and Red-Team Evaluation}

A third modality uses the Policy Kernel for offline analysis of
historical event streams.

Given a normalized event sequence:

\begin{equation}
E = \{e_1,e_2,\ldots,e_n\},
\end{equation}

the governance engine can evaluate the sequence against a candidate
policy set and identify:

\begin{itemize}
    \item operations that would have been denied;
    \item operations requiring approval;
    \item obligations that would have been generated;
    \item events with no applicable policy;
    \item potential unmediated paths; and
    \item policy coverage gaps.
\end{itemize}

This enables counterfactual governance analysis without modifying the
original execution.

Such replay can also support continuous red-team evaluation in which
agents search for execution paths that violate specified governance
invariants.

\subsection{Financial Crime and Dispute Management Harness}
\label{app:fincrime}

Financial crime and dispute-management workflows provide a useful
domain for demonstrating governance requirements that extend beyond
authorization.

In such workflows, automated decisions may influence which cases are
reviewed, which customers receive provisional credits, which
transactions are blocked, and which outcomes become available as
training labels.

\subsubsection{Selective-Label Governance}

An important distinction is that an unreviewed case should not
automatically be interpreted as a negative outcome.

A governance-aware system can distinguish between:

\begin{itemize}
    \item confirmed loss;
    \item analyst disposition;
    \item unresolved;
    \item unobserved because intervention occurred.
\end{itemize}

The fourth category explicitly records situations in which the
intervention itself prevented the underlying outcome from becoming
observable.

This distinction is relevant to model development because the
distribution of observed labels can depend on the operational policy
used to select cases for intervention.

\subsubsection{Reference Harness}

A governance-oriented dispute-management harness may incorporate:

\begin{itemize}
    \item multiple label bands;
    \item a randomized non-intervention slice;
    \item inverse-propensity weighting;
    \item temporally separated training, selection, and blind evaluation
    windows;
    \item role separation among proposer, implementer, executor, and
    analyst;
    \item explicit fairness and operational-capacity constraints;
    \item disturbance and drift scenarios;
    \item deterministic artifact generation; and
    \item an append-only execution ledger.
\end{itemize}

These controls transform governance requirements from procedural
guidance into enforceable runtime constraints.

\subsubsection{Role Governance}

An illustrative role mapping is shown below.

\begin{table}[ht]
\centering
\caption{Illustrative governance boundaries for a dispute-management harness.}
\label{tab:fincrime-roles}
\begin{tabular}{p{2.2cm}p{2.5cm}p{3cm}}
\hline
\textbf{Role} & \textbf{Principal} & \textbf{Representative Boundary} \\
\hline
Proposer &
AIAgent / ProposerRole &
May read summary information and create specifications; cannot access
raw data or blind evaluation windows. \\

Implementer &
AIAgent / ImplementerRole &
May fit models on designated replay data; cannot access objective
functions or blind data. \\

Executor &
AIAgent / ExecutorRole &
May evaluate models and append execution evidence; cannot modify
selection authority. \\

Analyst &
AIAgent / AnalystRole &
May inspect governance records and create directives; cannot access
restricted artifacts. \\

Validator &
Human / ModelRiskValidator &
May review artifacts and approve governed promotion decisions. \\
\hline
\end{tabular}
\end{table}

\subsubsection{Blind-Window Isolation}

The blind evaluation window can be treated as a first-class governed
resource.

\begin{lstlisting}[language=json,
caption={Illustrative blind-window isolation policy},
label={lst:blind-window}]
{
  "sid": "BlindWindowIsolation",
  "effect": "Deny",
  "principal": {
    "Type": ["AIAgent"]
  },
  "action": [
    "data:read",
    "model:fit",
    "model:evaluate",
    "rag:query"
  ],
  "resource": [
    "window:BLIND"
  ],
  "condition": {
    "any": [
      {
        "fact": "state.run.blind_reads",
        "operator": "greater_than",
        "value": 0
      },
      {
        "fact": "context.artifact_frozen",
        "operator": "equal",
        "value": false
      }
    ]
  },
  "fail_closed": true,
  "plugins": [
    {
      "provider": "audit_logger",
      "mode": "record"
    }
  ]
}
\end{lstlisting}

The example demonstrates how a procedural requirement can become a
declarative governance rule.

\subsection{Extended Financial Crime Policy Pack}
\label{app:fincrime-pack}

An illustrative Financial Crime / Dispute Management policy pack may
define domain-specific actions such as:

\begin{itemize}
    \item \texttt{dispute:file}
    \item \texttt{dispute:triage}
    \item \texttt{dispute:represent}
    \item \texttt{dispute:credit\_provisional}
    \item \texttt{dispute:deny}
    \item \texttt{transaction:block}
    \item \texttt{transaction:hold}
    \item \texttt{case:escalate}
    \item \texttt{label:emit}
    \item \texttt{population:audit\_sample}
\end{itemize}

The policy pack may associate autonomous decisions with governance
obligations including:

\begin{enumerate}
    \item \textbf{Evidence generation} --- record the normalized
    request, matched policy statements, relevant evaluation facts,
    model version, and artifact integrity information.
    
    \item \textbf{Label provenance} --- identify whether an outcome
    represents confirmed loss, analyst disposition, unresolved status,
    or an outcome that became unobservable because intervention
    occurred.
    
    \item \textbf{Exploration reserve} --- preserve a bounded,
    randomized non-intervention sample under explicit governance.
    
    \item \textbf{Maturity gate} --- prevent model retraining or
    promotion when the available outcome window has not matured
    sufficiently.
\end{enumerate}

These mechanisms illustrate how industry policy packs can encode
governance controls beyond conventional access authorization.

\subsection{Additional Extended Policies}
\label{app:extended-policies}

\subsubsection{Denying Leakage-Prone Features}

A policy may prohibit the use of features that would leak information
from restricted evaluation windows:

\begin{lstlisting}[language=json,
caption={Illustrative feature-leakage policy},
label={lst:feature-leakage}]
{
  "sid": "DenyLeakageFeatures",
  "effect": "Deny",
  "action": [
    "feature:use",
    "model:fit"
  ],
  "resource": [
    "feature:prior_confirmed_abuse",
    "feature:ring_linked",
    "feature:exculpatory_evidence"
  ]
}
\end{lstlisting}

\subsubsection{Privacy Protection Obligation}

A privacy policy may attach transformation and evidence obligations to
LLM inputs:

\begin{lstlisting}[language=json,
caption={Illustrative privacy protection policy},
label={lst:privacy-policy}]
{
  "sid": "FinCrimeInputProtection",
  "effect": "Allow",
  "action": [
    "llm:input"
  ],
  "resource": [
    "*"
  ],
  "plugins": [
    {
      "provider": "presidio",
      "mode": "redact"
    },
    {
      "provider": "pseudonymiser",
      "mode": "tokenise",
      "config": {
        "fields": [
          "customer_id",
          "merchant_id",
          "institution_id"
        ],
        "salt_scope": "run"
      }
    },
    {
      "provider": "audit_logger",
      "mode": "record"
    }
  ]
}
\end{lstlisting}

The example demonstrates that a permitted operation can still carry
mandatory governance obligations.

\subsubsection{Fairness and Capacity Constraints}

Governance conditions can also constrain promotion of detection
artifacts:

\begin{lstlisting}[language=json,
caption={Illustrative fairness and capacity policy},
label={lst:fairness-policy}]
{
  "sid": "FairnessConstraintOnDetectionArtifact",
  "effect": "Deny",
  "action": [
    "model:promote",
    "policy:publish"
  ],
  "resource": [
    "detection-artifact:*"
  ],
  "condition": {
    "any": [
      {
        "fact": "fairness.min_disparate_impact",
        "operator": "less_than",
        "value": 0.80
      },
      {
        "fact": "performance.max_segment_fp",
        "operator": "greater_than",
        "value": 0.06
      },
      {
        "fact": "operations.alert_rate",
        "operator": "greater_than",
        "value": 0.05
      }
    ]
  }
}
\end{lstlisting}

The numerical thresholds in this example are illustrative and should
not be interpreted as universal regulatory or operational thresholds.

\subsection{Human Approval Boundaries}
\label{app:approval-boundaries}

Certain operations may require explicit human approval rather than
fully autonomous execution.

For example:

\begin{lstlisting}[language=json,
caption={Illustrative human approval boundary},
label={lst:approval-boundary}]
{
  "sid": "DetectionArtifactPromotion",
  "effect": "Approve",
  "action": [
    "model:promote"
  ],
  "resource": [
    "detection-artifact:*"
  ],
  "approval": {
    "role": "ModelRiskValidator"
  },
  "plugins": [
    {
      "provider": "evidence_generator",
      "mode": "model_card"
    },
    {
      "provider": "integrity_verifier",
      "mode": "sha256"
    }
  ]
}
\end{lstlisting}

The \texttt{Approve} effect represents a governance requirement for
human authorization and is distinct from an ordinary \texttt{Allow}
decision.

\subsection{Benchmark Extensions}
\label{app:benchmark-extensions}

The existing EAGBench benchmark evaluates enterprise governance
scenarios across multiple domains. The extended governance model
suggests additional benchmark dimensions for evaluating properties that
cannot be captured by decision accuracy alone.

A future Financial Crime / Dispute Management benchmark extension may \\
include the following metrics.

\begin{table}[ht]
\centering
\caption{Proposed governance-specific benchmark metrics.}
\label{tab:extended-benchmark}
\begin{tabular}{p{3.5cm}p{4cm}}
\hline
\textbf{Metric} & \textbf{Evaluation Question} \\
\hline
Label-provenance completeness &
What fraction of automated decisions emit an appropriate
observability-band record? \\

Exploration-reserve compliance &
Is the mandated non-intervention sample preserved under operational
pressure? \\

Leakage-attempt interception &
Are restricted blind-window and excluded-feature accesses denied? \\

Constraint-breach containment &
Are fairness and operational-capacity violations detected before
artifact promotion? \\

Evidence reconstructability &
Can the complete governance decision chain be reconstructed from the
evidence log? \\

Approval-boundary integrity &
Did irreversible actions execute without the required approval? \\
\hline
\end{tabular}
\end{table}

For A2A governance, additional metrics may include:

\begin{itemize}
    \item mediation coverage;
    \item channel-formation detection latency;
    \item cross-run credential-reuse denial rate;
    \item provenance-policy interception rate; and
    \item time-to-containment.
\end{itemize}

These metrics are proposed extensions and require independent
implementation and experimental validation before being presented as
empirical benchmark results.

\subsection{Incident-Replay Evaluation as Future Work}
\label{app:incident-replay}

The extended UPA model also enables counterfactual replay of historical
agentic incidents.

Given a timestamped event sequence from an incident, each event can be
normalized into the UPA request model and evaluated against a candidate
policy set. The resulting analysis can identify:

\begin{enumerate}
    \item which events would have been denied;
    \item which events would have required human approval;
    \item which obligations would have been generated;
    \item which events lacked sufficient provenance;
    \item which operations occurred outside the mediated enforcement
    surface; and
    \item where governance state would have triggered escalation.
\end{enumerate}

Such an evaluation should be interpreted as a counterfactual analysis,
not as evidence that UPA would necessarily have prevented the original
incident. Application-layer governance cannot by itself guarantee
containment of host-level compromise, kernel compromise, or other
attacks that bypass the governance enforcement surface.

The defensible objective is instead to determine whether the governance
architecture can express, enforce, and evidence the relevant
coordination and execution constraints.

\subsection{Extended SID Namespace}
\label{app:extended-sid}

The following namespaces represent candidate extensions to the UPA SID
registry.

\begin{table}[ht]
\centering
\caption{Candidate extended SID namespaces.}
\label{tab:extended-sid}
\begin{tabular}{p{2.5cm}p{5cm}}
\hline
\textbf{Namespace} & \textbf{Representative Operations} \\
\hline
\texttt{a2a:*} &
\texttt{send}, \texttt{receive}, \texttt{broadcast},
\texttt{discover}, \texttt{delegate}, \texttt{handoff} \\

\texttt{dispute:*} &
\texttt{file}, \texttt{triage}, \texttt{represent},
\texttt{credit\_provisional}, \texttt{deny} \\

\texttt{transaction:*} &
\texttt{block}, \texttt{hold} \\

\texttt{case:*} &
\texttt{escalate} \\

\texttt{label:*} &
\texttt{emit} \\

\texttt{population:*} &
\texttt{audit\_sample} \\
\hline
\end{tabular}
\end{table}

These namespaces are proposed extensions and should be incorporated
into the normative SID registry only after their semantics and
interoperability requirements are established.

\subsection{Scope of the Extended Model}
\label{app:extended-scope}

The extensions in this appendix deliberately distinguish between the
core UPA architecture and proposed capabilities.

The core architecture establishes a policy-centric governance layer
for Enterprise AI execution. The extensions presented here investigate
how the same abstraction can be expanded to cover:

\begin{itemize}
    \item multi-agent communication;
    \item information provenance;
    \item aggregate runtime behavior;
    \item governance-state-dependent decisions;
    \item pre-flight policy evaluation;
    \item offline policy assurance;
    \item domain-specific governance;
    \item selective-label controls;
    \item human approval boundaries; and
    \item governance-specific benchmarking.
\end{itemize}

These extensions should therefore be viewed as an extensibility
demonstration of UPA rather than as claims that every capability is
already implemented or empirically validated.

In particular, UPA should not be interpreted as a complete substitute
for host security, identity infrastructure, network isolation, runtime
sandboxing, or other independent security controls. Governance
enforcement is strongest when combined with these layers and when the
mediation surface is explicitly audited.

\end{document}